\documentclass[12pt]{article}

\usepackage{geometry}
\usepackage[utf8]{inputenc}
\usepackage{authblk}

\usepackage{setspace}
\usepackage[superscript]{cite}
\usepackage{amssymb}
\usepackage{xcolor}
\usepackage{subcaption}
\usepackage{array}
\usepackage{pdflscape}
\usepackage{titlesec}
\usepackage{float}
\usepackage{multirow}
\usepackage{longtable}
\usepackage{afterpage}
\usepackage{amsmath}
\usepackage{graphicx}
\usepackage{tabularx}
\usepackage{hyperref}
\usepackage{tikz}
\hypersetup{
    colorlinks=true,
    linkcolor=blue,
    filecolor=magenta,      
    urlcolor=cyan,
    pdftitle={Overleaf Example},
    pdfpagemode=FullScreen,
    }

\newcommand{\indep}{\perp \!\!\! \perp}

\newcommand{\blue}{\textcolor{black}}
\title{\bf What is Fair? Defining Fairness in \\ Machine Learning for Health}
\author[1]{Jianhui Gao, MS}
\author[1]{Benson Chou, BA}
\author[2]{Zachary R. McCaw, PhD}
\author[3]{Hilary Thurston, MA}
\author[4]{Paul Varghese, MD}
\author[5]{Chuan Hong, PhD}
\author[1,6,7]{Jessica Gronsbell, PhD}

\affil[1]{Department of Statistical Sciences, University of Toronto, Toronto, ON, Canada}
\affil[2]{Department of Biostatistics, Harvard T.H.\ Chan School of Public Health, Boston, MA, USA}
\affil[3]{Department of Gender, Feminist \& Women's Studies, York University, Toronto, ON, Canada}
\affil[4]{Verily Life Sciences, Cambridge, MA, USA}
\affil[5]{Department of Biostatistics, Duke University Raleigh, NC, USA}
\affil[6]{Department of Family \& Community Medicine, University of Toronto, Toronto, ON, Canada}
\affil[7]{Department of Computer Science, University of Toronto, Toronto, ON, Canada}
\date{}

\begin{document}


%
%
%
%
%
\def\bzero{{\bf 0}}
\def\bone{{\bf 1}}
%
%
%
%
\def\ba{{\mbox{\boldmath$a$}}}
\def\bb{{\bf b}}
\def\bc{{\bf c}}
\def\bd{{\bf d}}
\def\be{{\bf e}}
\def\bdf{{\bf f}}
\def\bg{{\mbox{\boldmath$g$}}}
\def\bh{{\bf h}}
\def\bi{{\bf i}}
\def\bj{{\bf j}}
\def\bk{{\bf k}}
\def\bl{{\bf l}}
\def\bm{{\bf m}}
\def\bn{{\bf n}}
\def\bo{{\bf o}}
\def\bp{{\bf p}}
\def\bq{{\bf q}}
\def\br{{\bf r}}
\def\bs{{\bf s}}
\def\bt{{\bf t}}
\def\bu{{\bf u}}
\def\bv{{\bf v}}
\def\bw{{\bf w}}
\def\bx{{\bf x}}
\def\by{{\bf y}}
\def\bz{{\bf z}}
\def\bA{{\bf A}}
\def\bB{{\bf B}}
\def\bC{{\bf C}}
\def\bD{{\bf D}}
\def\bE{{\bf E}}
\def\bF{{\bf F}}
\def\bG{{\bf G}}
\def\bH{{\bf H}}
\def\bI{{\bf I}}
\def\bJ{{\bf J}}
\def\bK{{\bf K}}
\def\bL{{\bf L}}
\def\bM{{\bf M}}
\def\bN{{\bf N}}
\def\bO{{\bf O}}
\def\bP{{\bf P}}
\def\bQ{{\bf Q}}
\def\bR{{\bf R}}
\def\bS{{\bf S}}
\def\bT{{\bf T}}
\def\bU{{\bf U}}
\def\bV{{\bf V}}
\def\bW{{\bf W}}
\def\bX{{\bf X}}
\def\bY{{\bf Y}}
\def\bZ{{\bf Z}}
\def\smbZ{\scriptstyle{\bf Z}}
\def\smM{\scriptstyle{M}}
\def\smN{\scriptstyle{N}}
\def\smbT{\scriptstyle{\bf T}}
%
%
%
%
\def\thick#1{\hbox{\rlap{$#1$}\kern0.25pt\rlap{$#1$}\kern0.25pt$#1$}}
\def\balpha{\boldsymbol{\alpha}}
\def\bbeta{\boldsymbol{\beta}}
\def\bgamma{\boldsymbol{\gamma}}
\def\bdelta{\boldsymbol{\delta}}
\def\bepsilon{\boldsymbol{\epsilon}}
\def\bvarepsilon{\boldsymbol{\varepsilon}}
\def\bzeta{\boldsymbol{\zeta}}
\def\bdeta{\boldsymbol{\eta}}
\def\btheta{\boldsymbol{\theta}}
\def\biota{\boldsymbol{\iota}}
\def\bkappa{\boldsymbol{\kappa}}
\def\blambda{\boldsymbol{\lambda}}
\def\bmu{\boldsymbol{\mu}}
\def\bnu{\boldsymbol{\nu}}
\def\bxi{\boldsymbol{\xi}}
\def\bomicron{\boldsymbol{\omicron}}
\def\bpi{\boldsymbol{\pi}}
\def\brho{\boldsymbol{\rho}}
\def\bsigma{\boldsymbol{\sigma}}
\def\btau{\boldsymbol{\tau}}
\def\bupsilon{\boldsymbol{\upsilon}}
\def\bphi{\boldsymbol{\phi}}
\def\bchi{\boldsymbol{\chi}}
\def\bpsi{\boldsymbol{\psi}}
\def\bomega{\boldsymbol{\omega}}
\def\bAlpha{\boldsymbol{\Alpha}}
\def\bBeta{\boldsymbol{\Beta}}
\def\bGamma{\boldsymbol{\Gamma}}
\def\bDelta{\boldsymbol{\Delta}}
\def\bEpsilon{\boldsymbol{\Epsilon}}
\def\bZeta{\boldsymbol{\Zeta}}
\def\bEta{\boldsymbol{\Eta}}
\def\bTheta{\boldsymbol{\Theta}}
\def\bIota{\boldsymbol{\Iota}}
\def\bKappa{\boldsymbol{\Kappa}}
\def\bLambda{{\boldsymbol{\Lambda}}}
\def\bMu{\boldsymbol{\Mu}}
\def\bNu{\boldsymbol{\Nu}}
\def\bXi{\boldsymbol{\Xi}}
\def\bOmicron{\boldsymbol{\Omicron}}
\def\bPi{\boldsymbol{\Pi}}
\def\bRho{\boldsymbol{\Rho}}
\def\bSigma{\boldsymbol{\Sigma}}
\def\bTau{\boldsymbol{\Tau}}
\def\bUpsilon{\boldsymbol{\Upsilon}}
\def\bPhi{\boldsymbol{\Phi}}
\def\bChi{\boldsymbol{\Chi}}
\def\bPsi{\boldsymbol{\Psi}}
\def\bOmega{\boldsymbol{\Omega}}
%
%
%
\def\smalpha{{{\scriptstyle{\alpha}}}}
\def\smbeta{{{\scriptstyle{\beta}}}}
\def\smgamma{{{\scriptstyle{\gamma}}}}
\def\smdelta{{{\scriptstyle{\delta}}}}
\def\smepsilon{{{\scriptstyle{\epsilon}}}}
\def\smvarepsilon{{{\scriptstyle{\varepsilon}}}}
\def\smzeta{{{\scriptstyle{\zeta}}}}
\def\smdeta{{{\scriptstyle{\eta}}}}
\def\smtheta{{{\scriptstyle{\theta}}}}
\def\smiota{{{\scriptstyle{\iota}}}}
\def\smkappa{{{\scriptstyle{\kappa}}}}
\def\smlambda{{{\scriptstyle{\lambda}}}}
\def\smmu{{{\scriptstyle{\mu}}}}
\def\smnu{{{\scriptstyle{\nu}}}}
\def\smxi{{{\scriptstyle{\xi}}}}
\def\smomicron{{{\scriptstyle{\omicron}}}}
\def\smpi{{{\scriptstyle{\pi}}}}
\def\smrho{{{\scriptstyle{\rho}}}}
\def\smsigma{{{\scriptstyle{\sigma}}}}
\def\smtau{{{\scriptstyle{\tau}}}}
\def\smupsilon{{{\scriptstyle{\upsilon}}}}
\def\smphi{{{\scriptstyle{\phi}}}}
\def\smchi{{{\scriptstyle{\chi}}}}
\def\smpsi{{{\scriptstyle{\psi}}}}
\def\smomega{{{\scriptstyle{\omega}}}}
\def\smAlpha{{{\scriptstyle{\Alpha}}}}
\def\smBeta{{{\scriptstyle{\Beta}}}}
\def\smGamma{{{\scriptstyle{\Gamma}}}}
\def\smDelta{{{\scriptstyle{\Delta}}}}
\def\smEpsilon{{{\scriptstyle{\Epsilon}}}}
\def\smZeta{{{\scriptstyle{\Zeta}}}}
\def\smEta{{{\scriptstyle{\Eta}}}}
\def\smTheta{{{\scriptstyle{\Theta}}}}
\def\smIota{{{\scriptstyle{\Iota}}}}
\def\smKappa{{{\scriptstyle{\Kappa}}}}
\def\smLambda{{{\scriptstyle{\Lambda}}}}
\def\smMu{{{\scriptstyle{\Mu}}}}
\def\smNu{{{\scriptstyle{\Nu}}}}
\def\smXi{{{\scriptstyle{\Xi}}}}
\def\smOmicron{{{\scriptstyle{\Omicron}}}}
\def\smPi{{{\scriptstyle{\Pi}}}}
\def\smRho{{{\scriptstyle{\Rho}}}}
\def\smSigma{{{\scriptstyle{\Sigma}}}}
\def\smTau{{{\scriptstyle{\Tau}}}}
\def\smUpsilon{{{\scriptstyle{\Upsilon}}}}
\def\smPhi{{{\scriptstyle{\Phi}}}}
\def\smChi{{{\scriptstyle{\Chi}}}}
\def\smPsi{{{\scriptstyle{\Psi}}}}
\def\smOmega{{{\scriptstyle{\Omega}}}}
%
%

%
\def\smbalpha{\boldsymbol{{\scriptstyle{\alpha}}}}
\def\smbbeta{\boldsymbol{{\scriptstyle{\beta}}}}
\def\smbgamma{\boldsymbol{{\scriptstyle{\gamma}}}}
\def\smbdelta{\boldsymbol{{\scriptstyle{\delta}}}}
\def\smbepsilon{\boldsymbol{{\scriptstyle{\epsilon}}}}
\def\smbvarepsilon{\boldsymbol{{\scriptstyle{\varepsilon}}}}
\def\smbzeta{\boldsymbol{{\scriptstyle{\zeta}}}}
\def\smbdeta{\boldsymbol{{\scriptstyle{\eta}}}}
\def\smbtheta{\boldsymbol{{\scriptstyle{\theta}}}}
\def\smbiota{\boldsymbol{{\scriptstyle{\iota}}}}
\def\smbkappa{\boldsymbol{{\scriptstyle{\kappa}}}}
\def\smblambda{\boldsymbol{{\scriptstyle{\lambda}}}}
\def\smbmu{\boldsymbol{{\scriptstyle{\mu}}}}
\def\smbnu{\boldsymbol{{\scriptstyle{\nu}}}}
\def\smbxi{\boldsymbol{{\scriptstyle{\xi}}}}
\def\smbomicron{\boldsymbol{{\scriptstyle{\omicron}}}}
\def\smbpi{\boldsymbol{{\scriptstyle{\pi}}}}
\def\smbrho{\boldsymbol{{\scriptstyle{\rho}}}}
\def\smbsigma{\boldsymbol{{\scriptstyle{\sigma}}}}
\def\smbtau{\boldsymbol{{\scriptstyle{\tau}}}}
\def\smbupsilon{\boldsymbol{{\scriptstyle{\upsilon}}}}
\def\smbphi{\boldsymbol{{\scriptstyle{\phi}}}}
\def\smbchi{\boldsymbol{{\scriptstyle{\chi}}}}
\def\smbpsi{\boldsymbol{{\scriptstyle{\psi}}}}
\def\smbomega{\boldsymbol{{\scriptstyle{\omega}}}}
\def\smbAlpha{\boldsymbol{{\scriptstyle{\Alpha}}}}
\def\smbBeta{\boldsymbol{{\scriptstyle{\Beta}}}}
\def\smbGamma{\boldsymbol{{\scriptstyle{\Gamma}}}}
\def\smbDelta{\boldsymbol{{\scriptstyle{\Delta}}}}
\def\smbEpsilon{\boldsymbol{{\scriptstyle{\Epsilon}}}}
\def\smbZeta{\boldsymbol{{\scriptstyle{\Zeta}}}}
\def\smbEta{\boldsymbol{{\scriptstyle{\Eta}}}}
\def\smbTheta{\boldsymbol{{\scriptstyle{\Theta}}}}
\def\smbIota{\boldsymbol{{\scriptstyle{\Iota}}}}
\def\smbKappa{\boldsymbol{{\scriptstyle{\Kappa}}}}
\def\smbLambda{\boldsymbol{{\scriptstyle{\Lambda}}}}
\def\smbMu{\boldsymbol{{\scriptstyle{\Mu}}}}
\def\smbNu{\boldsymbol{{\scriptstyle{\Nu}}}}
\def\smbXi{\boldsymbol{{\scriptstyle{\Xi}}}}
\def\smbOmicron{\boldsymbol{{\scriptstyle{\Omicron}}}}
\def\smbPi{\boldsymbol{{\scriptstyle{\Pi}}}}
\def\smbRho{\boldsymbol{{\scriptstyle{\Rho}}}}
\def\smbSigma{\boldsymbol{{\scriptstyle{\Sigma}}}}
\def\smbTau{\boldsymbol{{\scriptstyle{\Tau}}}}
\def\smbUpsilon{\boldsymbol{{\scriptstyle{\Upsilon}}}}
\def\smbPhi{\boldsymbol{{\scriptstyle{\Phi}}}}
\def\smbChi{\boldsymbol{{\scriptstyle{\Chi}}}}
\def\smbPsi{\boldsymbol{{\scriptstyle{\Psi}}}}
\def\smbOmega{\boldsymbol{{\scriptstyle{\Omega}}}}
%
%
%
%
\def\ahat{{\widehat a}}
\def\bhat{{\widehat b}}
\def\chat{{\widehat c}}
\def\dhat{{\widehat d}}
\def\ehat{{\widehat e}}
\def\fhat{{\widehat f}}
\def\ghat{{\widehat g}}
\def\hhat{{\widehat h}}
\def\ihat{{\widehat i}}
\def\jhat{{\widehat j}}
\def\khat{{\widehat k}}
\def\lhat{{\widehat l}}
\def\mhat{{\widehat m}}
\def\nhat{{\widehat n}}
\def\ohat{{\widehat o}}
\def\phat{{\widehat p}}
\def\qhat{{\widehat q}}
\def\rhat{{\widehat r}}
\def\shat{{\widehat s}}
\def\that{{\widehat t}}
\def\uhat{{\widehat u}}
\def\vhat{{\widehat v}}
\def\what{{\widehat w}}
\def\xhat{{\widehat x}}
\def\yhat{{\widehat y}}
\def\zhat{{\widehat z}}
\def\Ahat{{\widehat A}}
\def\Bhat{{\widehat B}}
\def\Chat{{\widehat C}}
\def\Dhat{{\widehat D}}
\def\Ehat{{\widehat E}}
\def\Fhat{{\widehat F}}
\def\Ghat{{\widehat G}}
\def\Hhat{{\widehat H}}
\def\Ihat{{\widehat I}}
\def\Jhat{{\widehat J}}
\def\Khat{{\widehat K}}
\def\Lhat{{\widehat L}}
\def\Mhat{{\widehat M}}
\def\Nhat{{\widehat N}}
\def\Ohat{{\widehat O}}
\def\Phat{{\widehat P}}
\def\Qhat{{\widehat Q}}
\def\Rhat{{\widehat R}}
\def\Shat{{\widehat S}}
\def\That{{\widehat T}}
\def\Uhat{{\widehat U}}
\def\Vhat{{\widehat V}}
\def\What{{\widehat W}}
\def\Xhat{{\widehat X}}
\def\Yhat{{\widehat Y}}
\def\Zhat{{\widehat Z}}
%
%
%
\def\atilde{{\widetilde a}}
\def\btilde{{\widetilde b}}
\def\ctilde{{\widetilde c}}
\def\dtilde{{\widetilde d}}
\def\etilde{{\widetilde e}}
\def\ftilde{{\widetilde f}}
\def\gtilde{{\widetilde g}}
\def\htilde{{\widetilde h}}
\def\itilde{{\widetilde i}}
\def\jtilde{{\widetilde j}}
\def\ktilde{{\widetilde k}}
\def\ltilde{{\widetilde l}}
\def\mtilde{{\widetilde m}}
\def\ntilde{{\widetilde n}}
\def\otilde{{\widetilde o}}
\def\ptilde{{\widetilde p}}
\def\qtilde{{\widetilde q}}
\def\rtilde{{\widetilde r}}
\def\stilde{{\widetilde s}}
\def\ttilde{{\widetilde t}}
\def\utilde{{\widetilde u}}
\def\vtilde{{\widetilde v}}
\def\wtilde{{\widetilde w}}
\def\xtilde{{\widetilde x}}
\def\ytilde{{\widetilde y}}
\def\ztilde{{\widetilde z}}
\def\Atilde{{\widetilde A}}
\def\Btilde{{\widetilde B}}
\def\Ctilde{{\widetilde C}}
\def\Dtilde{{\widetilde D}}
\def\Etilde{{\widetilde E}}
\def\Ftilde{{\widetilde F}}
\def\Gtilde{{\widetilde G}}
\def\Htilde{{\widetilde H}}
\def\Itilde{{\widetilde I}}
\def\Jtilde{{\widetilde J}}
\def\Ktilde{{\widetilde K}}
\def\Ltilde{{\widetilde L}}
\def\Mtilde{{\widetilde M}}
\def\Ntilde{{\widetilde N}}
\def\Otilde{{\widetilde O}}
\def\Ptilde{{\widetilde P}}
\def\Qtilde{{\widetilde Q}}
\def\Rtilde{{\widetilde R}}
\def\Stilde{{\widetilde S}}
\def\Ttilde{{\widetilde T}}
\def\Utilde{{\widetilde U}}
\def\Vtilde{{\widetilde V}}
\def\Wtilde{{\widetilde W}}
\def\Xtilde{{\widetilde X}}
\def\Ytilde{{\widetilde Y}}
\def\Ztilde{{\widetilde Z}}
%
%
%
%
\def\bahat{{\widehat \ba}}
\def\bbhat{{\widehat \bb}}
\def\bchat{{\widehat \bc}}
\def\bdhat{{\widehat \bd}}
\def\behat{{\widehat \be}}
\def\bfhat{{\widehat \bf}}
\def\bghat{{\widehat \bg}}
\def\bhhat{{\widehat \bh}}
\def\bihat{{\widehat \bi}}
\def\bjhat{{\widehat \bj}}
\def\bkhat{{\widehat \bk}}
\def\blhat{{\widehat \bl}}
\def\bmhat{{\widehat \bm}}
\def\bnhat{{\widehat \bn}}
\def\bohat{{\widehat \bo}}
\def\bphat{{\widehat \bp}}
\def\bqhat{{\widehat \bq}}
\def\brhat{{\widehat \br}}
\def\bshat{{\widehat \bs}}
\def\bthat{{\widehat \bt}}
\def\buhat{{\widehat \bu}}
\def\bvhat{{\widehat \bv}}
\def\bwhat{{\widehat \bw}}
\def\bxhat{{\widehat \bx}}
\def\byhat{{\widehat \by}}
\def\bzhat{{\widehat \bz}}
\def\bAhat{{\widehat \bA}}
\def\bBhat{{\widehat \bB}}
\def\bChat{{\widehat \bC}}
\def\bDhat{{\widehat \bD}}
\def\bEhat{{\widehat \bE}}
\def\bFhat{{\widehat \bF}}
\def\bGhat{{\widehat \bG}}
\def\bHhat{{\widehat \bH}}
\def\bIhat{{\widehat \bI}}
\def\bJhat{{\widehat \bJ}}
\def\bKhat{{\widehat \bK}}
\def\bLhat{{\widehat \bL}}
\def\bMhat{{\widehat \bM}}
\def\bNhat{{\widehat \bN}}
\def\bOhat{{\widehat \bO}}
\def\bPhat{{\widehat \bP}}
\def\bQhat{{\widehat \bQ}}
\def\bRhat{{\widehat \bR}}
\def\bShat{{\widehat \bS}}
\def\bThat{{\widehat \bT}}
\def\bUhat{{\widehat \bU}}
\def\bVhat{{\widehat \bV}}
\def\bWhat{{\widehat \bW}}
\def\bXhat{{\widehat \bX}}
\def\bYhat{{\widehat \bY}}
\def\bZhat{{\widehat \bZ}}
%
%
%
%
%
\def\batilde{{\widetilde \ba}}
\def\bbtilde{{\widetilde \bb}}
\def\bctilde{{\widetilde \bc}}
\def\bdtilde{{\widetilde \bd}}
\def\betilde{{\widetilde \be}}
\def\bftilde{{\widetilde \bf}}
\def\bgtilde{{\widetilde \bg}}
\def\bhtilde{{\widetilde \bh}}
\def\bitilde{{\widetilde \bi}}
\def\bjtilde{{\widetilde \bj}}
\def\bktilde{{\widetilde \bk}}
\def\bltilde{{\widetilde \bl}}
\def\bmtilde{{\widetilde \bm}}
\def\bntilde{{\widetilde \bn}}
\def\botilde{{\widetilde \bo}}
\def\bptilde{{\widetilde \bp}}
\def\bqtilde{{\widetilde \bq}}
\def\brtilde{{\widetilde \br}}
\def\bstilde{{\widetilde \bs}}
\def\bttilde{{\widetilde \bt}}
\def\butilde{{\widetilde \bu}}
\def\bvtilde{{\widetilde \bv}}
\def\bwtilde{{\widetilde \bw}}
\def\bxtilde{{\widetilde \bx}}
\def\bytilde{{\widetilde \by}}
\def\bztilde{{\widetilde \bz}}
\def\bAtilde{{\widetilde \bA}}
\def\bBtilde{{\widetilde \bB}}
\def\bCtilde{{\widetilde \bC}}
\def\bDtilde{{\widetilde \bD}}
\def\bEtilde{{\widetilde \bE}}
\def\bFtilde{{\widetilde \bF}}
\def\bGtilde{{\widetilde \bG}}
\def\bHtilde{{\widetilde \bH}}
\def\bItilde{{\widetilde \bI}}
\def\bJtilde{{\widetilde \bJ}}
\def\bKtilde{{\widetilde \bK}}
\def\bLtilde{{\widetilde \bL}}
\def\bMtilde{{\widetilde \bM}}
\def\bNtilde{{\widetilde \bN}}
\def\bOtilde{{\widetilde \bO}}
\def\bPtilde{{\widetilde \bP}}
\def\bQtilde{{\widetilde \bQ}}
\def\bRtilde{{\widetilde \bR}}
\def\bStilde{{\widetilde \bS}}
\def\bTtilde{{\widetilde \bT}}
\def\bUtilde{{\widetilde \bU}}
\def\bVtilde{{\widetilde \bV}}
\def\bWtilde{{\widetilde \bW}}
\def\bXtilde{{\widetilde \bX}}
\def\bYtilde{{\widetilde \bY}}
\def\bZtilde{{\widetilde \bZ}}
%
%
%
%
%
%
\def\alphahat{{\widehat\alpha}}
\def\betahat{{\widehat\beta}}
\def\gammahat{{\widehat\gamma}}
\def\deltahat{{\widehat\delta}}
\def\epsilonhat{{\widehat\epsilon}}
\def\varepsilonhat{{\widehat\varepsilon}}
\def\zetahat{{\widehat\zeta}}
\def\etahat{{\widehat\eta}}
\def\thetahat{{\widehat\theta}}
\def\iotahat{{\widehat\iota}}
\def\kappahat{{\widehat\kappa}}
\def\lambdahat{{\widehat\lambda}}
\def\muhat{{\widehat\mu}}
\def\nuhat{{\widehat\nu}}
\def\xihat{{\widehat\xi}}
\def\omicronhat{{\widehat\omicron}}
\def\pihat{{\widehat\pi}}
\def\rhohat{{\widehat\rho}}
\def\sigmahat{{\widehat\sigma}}
\def\tauhat{{\widehat\tau}}
\def\upsilonhat{{\widehat\upsilon}}
\def\phihat{{\widehat\phi}}
\def\chihat{{\widehat\chi}}
\def\psihat{{\widehat\psi}}
\def\omegahat{{\widehat\omega}}
\def\Alphahat{{\widehat\Alpha}}
\def\Betahat{{\widehat\Beta}}
\def\Gammahat{{\widehat\Gamma}}
\def\Deltahat{{\widehat\Delta}}
\def\Epsilonhat{{\widehat\Epsilon}}
\def\Zetahat{{\widehat\Zeta}}
\def\Etahat{{\widehat\Eta}}
\def\Thetahat{{\widehat\Theta}}
\def\Iotahat{{\widehat\Iota}}
\def\Kappahat{{\widehat\Kappa}}
\def\Lambdahat{{\widehat\Lambda}}
\def\Muhat{{\widehat\Mu}}
\def\Nuhat{{\widehat\Nu}}
\def\Xihat{{\widehat\Xi}}
\def\Omicronhat{{\widehat\Omicron}}
\def\Pihat{{\widehat\Pi}}
\def\Rhohat{{\widehat\Rho}}
\def\Sigmahat{{\widehat\Sigma}}
\def\Tauhat{{\widehat\Tau}}
\def\Upsilonhat{{\widehat\Upsilon}}
\def\Phihat{{\widehat\Phi}}
\def\Chihat{{\widehat\Chi}}
\def\Psihat{{\widehat\Psi}}
\def\Omegahat{{\widehat\Omega}}
%
%
%
%
%
\def\alphatilde{{\widetilde\alpha}}
\def\betatilde{{\widetilde\beta}}
\def\gammatilde{{\widetilde\gamma}}
\def\deltatilde{{\widetilde\delta}}
\def\epsilontilde{{\widetilde\epsilon}}
\def\varepsilontilde{{\widetilde\varepsilon}}
\def\zetatilde{{\widetilde\zeta}}
\def\etatilde{{\widetilde\eta}}
\def\thetatilde{{\widetilde\theta}}
\def\iotatilde{{\widetilde\iota}}
\def\kappatilde{{\widetilde\kappa}}
\def\lambdatilde{{\widetilde\lambda}}
\def\mutilde{{\widetilde\mu}}
\def\nutilde{{\widetilde\nu}}
\def\xitilde{{\widetilde\xi}}
\def\omicrontilde{{\widetilde\omicron}}
\def\pitilde{{\widetilde\pi}}
\def\rhotilde{{\widetilde\rho}}
\def\sigmatilde{{\widetilde\sigma}}
\def\tautilde{{\widetilde\tau}}
\def\upsilontilde{{\widetilde\upsilon}}
\def\phitilde{{\widetilde\phi}}
\def\chitilde{{\widetilde\chi}}
\def\psitilde{{\widetilde\psi}}
\def\omegatilde{{\widetilde\omega}}
\def\Alphatilde{{\widetilde\Alpha}}
\def\Betatilde{{\widetilde\Beta}}
\def\Gammatilde{{\widetilde\Gamma}}
\def\Deltatilde{{\widetilde\Delta}}
\def\Epsilontilde{{\widetilde\Epsilon}}
\def\Zetatilde{{\widetilde\Zeta}}
\def\Etatilde{{\widetilde\Eta}}
\def\Thetatilde{{\widetilde\Theta}}
\def\Iotatilde{{\widetilde\Iota}}
\def\Kappatilde{{\widetilde\Kappa}}
\def\Lambdatilde{{\widetilde\Lambda}}
\def\Mutilde{{\widetilde\Mu}}
\def\Nutilde{{\widetilde\Nu}}
\def\Xitilde{{\widetilde\Xi}}
\def\Omicrontilde{{\widetilde\Omicron}}
\def\Pitilde{{\widetilde\Pi}}
\def\Rhotilde{{\widetilde\Rho}}
\def\Sigmatilde{{\widetilde\Sigma}}
\def\Tautilde{{\widetilde\Tau}}
\def\Upsilontilde{{\widetilde\Upsilon}}
\def\Phitilde{{\widetilde\Phi}}
\def\Chitilde{{\widetilde\Chi}}
\def\Psitilde{{\widetilde\Psi}}
\def\Omegatilde{{\widetilde\Omega}}
%
%
%
%
%
%
\def\balphahat{{\widehat\balpha}}
\def\bbetahat{{\widehat\bbeta}}
\def\bgammahat{{\widehat\bgamma}}
\def\bdeltahat{{\widehat\bdelta}}
\def\bepsilonhat{{\widehat\bepsilon}}
\def\bzetahat{{\widehat\bzeta}}
\def\bdetahat{{\widehat\bdeta}}
\def\bthetahat{{\widehat\btheta}}
\def\biotahat{{\widehat\biota}}
\def\bkappahat{{\widehat\bkappa}}
\def\blambdahat{{\widehat\blambda}}
\def\bmuhat{{\widehat\bmu}}
\def\bnuhat{{\widehat\bnu}}
\def\bxihat{{\widehat\bxi}}
\def\bomicronhat{{\widehat\bomicron}}
\def\bpihat{{\widehat\bpi}}
\def\brhohat{{\widehat\brho}}
\def\bsigmahat{{\widehat\bsigma}}
\def\btauhat{{\widehat\btau}}
\def\bupsilonhat{{\widehat\bupsilon}}
\def\bphihat{{\widehat\bphi}}
\def\bchihat{{\widehat\bchi}}
\def\bpsihat{{\widehat\bpsi}}
\def\bomegahat{{\widehat\bomega}}
\def\bAlphahat{{\widehat\bAlpha}}
\def\bBetahat{{\widehat\bBeta}}
\def\bGammahat{{\widehat\bGamma}}
\def\bDeltahat{{\widehat\bDelta}}
\def\bEpsilonhat{{\widehat\bEpsilon}}
\def\bZetahat{{\widehat\bZeta}}
\def\bEtahat{{\widehat\bEta}}
\def\bThetahat{{\widehat\bTheta}}
\def\bIotahat{{\widehat\bIota}}
\def\bKappahat{{\widehat\bKappa}}
\def\bLambdahat{{\widehat\bLambda}}
\def\bMuhat{{\widehat\bMu}}
\def\bNuhat{{\widehat\bNu}}
\def\bXihat{{\widehat\bXi}}
\def\bOmicronhat{{\widehat\bOmicron}}
\def\bPihat{{\widehat\bPi}}
\def\bRhohat{{\widehat\bRho}}
\def\bSigmahat{{\widehat\bSigma}}
\def\bTauhat{{\widehat\bTau}}
\def\bUpsilonhat{{\widehat\bUpsilon}}
\def\bPhihat{{\widehat\bPhi}}
\def\bChihat{{\widehat\bChi}}
\def\bPsihat{{\widehat\bPsi}}
\def\bOmegahat{{\widehat\bOmega}}%
\def\balphahattrans{{\balphahat^{_{\transpose}}}}
\def\bbetahattrans{{\bbetahat^{_{\transpose}}}}
\def\bgammahattrans{{\bgammahat^{_{\transpose}}}}
\def\bdeltahattrans{{\bdeltahat^{_{\transpose}}}}
\def\bepsilonhattrans{{\bepsilonhat^{_{\transpose}}}}
\def\bzetahattrans{{\bzetahat^{_{\transpose}}}}
\def\bdetahattrans{{\bdetahat^{_{\transpose}}}}
\def\bthetahattrans{{\bthetahat^{_{\transpose}}}}
\def\biotahattrans{{\biotahat^{_{\transpose}}}}
\def\bkappahattrans{{\bkappahat^{_{\transpose}}}}
\def\blambdahattrans{{\blambdahat^{_{\transpose}}}}
\def\bmuhattrans{{\bmuhat^{_{\transpose}}}}
\def\bnuhattrans{{\bnuhat^{_{\transpose}}}}
\def\bxihattrans{{\bxihat^{_{\transpose}}}}
\def\bomicronhattrans{{\bomicronhat^{_{\transpose}}}}
\def\bpihattrans{{\bpihat^{_{\transpose}}}}
\def\brhohattrans{{\brhohat^{_{\transpose}}}}
\def\bsigmahattrans{{\bsigmahat^{_{\transpose}}}}
\def\btauhattrans{{\btauhat^{_{\transpose}}}}
\def\bupsilonhattrans{{\bupsilonhat^{_{\transpose}}}}
\def\bphihattrans{{\bphihat^{_{\transpose}}}}
\def\bchihattrans{{\bchihat^{_{\transpose}}}}
\def\bpsihattrans{{\bpsihat^{_{\transpose}}}}
\def\bomegahattrans{{\bomegahat^{_{\transpose}}}}
\def\bAlphahattrans{{\bAlphahat^{_{\transpose}}}}
\def\bBetahattrans{{\bBetahat^{_{\transpose}}}}
\def\bGammahattrans{{\bGammahat^{_{\transpose}}}}
\def\bDeltahattrans{{\bDeltahat^{_{\transpose}}}}
\def\bEpsilonhattrans{{\bEpsilonhat^{_{\transpose}}}}
\def\bZetahattrans{{\bZetahat^{_{\transpose}}}}
\def\bEtahattrans{{\bEtahat^{_{\transpose}}}}
\def\bThetahattrans{{\bThetahat^{_{\transpose}}}}
\def\bIotahattrans{{\bIotahat^{_{\transpose}}}}
\def\bKappahattrans{{\bKappahat^{_{\transpose}}}}
\def\bLambdahattrans{{\bLambdahat^{_{\transpose}}}}
\def\bMuhattrans{{\bMuhat^{_{\transpose}}}}
\def\bNuhattrans{{\bNuhat^{_{\transpose}}}}
\def\bXihattrans{{\bXihat^{_{\transpose}}}}
\def\bOmicronhattrans{{\bOmicronhat^{_{\transpose}}}}
\def\bPihattrans{{\bPihat^{_{\transpose}}}}
\def\bRhohattrans{{\bRhohat^{_{\transpose}}}}
\def\bSigmahattrans{{\bSigmahat^{_{\transpose}}}}
\def\bTauhattrans{{\bTauhat^{_{\transpose}}}}
\def\bUpsilonhattrans{{\bUpsilonhat^{_{\transpose}}}}
\def\bPhihattrans{{\bPhihat^{_{\transpose}}}}
\def\bChihattrans{{\bChihat^{_{\transpose}}}}
\def\bPsihattrans{{\bPsihat^{_{\transpose}}}}
\def\bOmegahattrans{{\bOmegahat^{_{\transpose}}}}%
%
\def\smbalpha{\widehat{\smbalpha}}
\def\smbbetahat{\widehat{\smbbeta}}
\def\smbgammahat{\widehat{\smbgamma}}
\def\smbdeltahat{\widehat{\smbdelta}}
\def\smbepsilonhat{\widehat{\smbepsilon}}
\def\smbvarepsilonhat{\widehat{\smbvarepsilon}}
\def\smbzetahat{\widehat{\smbzeta}}
\def\smbdetahat{\widehat{\smbeta}}
\def\smbthetahat{\widehat{\smbtheta}}
\def\smbiotahat{\widehat{\smbiota}}
\def\smbkappahat{\widehat{\smbkappa}}
\def\smblambdahat{\widehat{\smblambda}}
\def\smbmuhat{\widehat{\smbmu}}
\def\smbnuhat{\widehat{\smbnu}}
\def\smbxihat{\widehat{\smbxi}}
\def\smbomicronhat{\widehat{\smbomicron}}
\def\smbpihat{\widehat{\smbpi}}
\def\smbrhohat{\widehat{\smbrho}}
\def\smbsigmahat{\widehat{\smbsigma}}
\def\smbtauhat{\widehat{\smbtau}}
\def\smbupsilonhat{\widehat{\smbupsilon}}
\def\smbphihat{\widehat{\smbphi}}
\def\smbchihat{\widehat{\smbchi}}
\def\smbpsihat{\widehat{\smbpsi}}
\def\smbomegahat{\widehat{\smbomega}}
\def\smbAlphahat{\widehat{\smbAlpha}}
\def\smbBetahat{\widehat{\smbBeta}}
\def\smbGammahat{\widehat{\smbGamma}}
\def\smbDeltahat{\widehat{\smbDelta}}
\def\smbEpsilonhat{\widehat{\smbEpsilon}}
\def\smbZetahat{\widehat{\smbZeta}}
\def\smbEtahat{\widehat{\smbEta}}
\def\smbThetahat{\widehat{\smbTheta}}
\def\smbIotahat{\widehat{\smbIota}}
\def\smbKappahat{\widehat{\smbKappa}}
\def\smbLambdahat{\widehat{\smbLambda}}
\def\smbMuhat{\widehat{\smbMu}}
\def\smbNuhat{\widehat{\smbNu}}
\def\smbXihat{\widehat{\smbXi}}
\def\smbOmicronhat{\widehat{\smbOmicron}}
\def\smbPihat{\widehat{\smbPi}}
\def\smbRhohat{\widehat{\smbRho}}
\def\smbSigmahat{\widehat{\smbSigma}}
\def\smbTauhat{\widehat{\smbTau}}
\def\smbUpsilonhat{\widehat{\smbUpsilon}}
\def\smbPhihat{\widehat{\smbPhi}}
\def\smbChihat{\widehat{\smbChi}}
\def\smbPsihat{\widehat{\smbPsi}}
\def\smbOmegahat{\widehat{\smbOmega}}
%
%
%
%
%
\def\balphatilde{{\widetilde\balpha}}
\def\bbetatilde{{\widetilde\bbeta}}
\def\bgammatilde{{\widetilde\bgamma}}
\def\bdeltatilde{{\widetilde\bdelta}}
\def\bepsilontilde{{\widetilde\bepsilon}}
\def\bzetatilde{{\widetilde\bzeta}}
\def\bdetatilde{{\widetilde\bdeta}}
\def\bthetatilde{{\widetilde\btheta}}
\def\biotatilde{{\widetilde\biota}}
\def\bkappatilde{{\widetilde\bkappa}}
\def\blambdatilde{{\widetilde\blambda}}
\def\bmutilde{{\widetilde\bmu}}
\def\bnutilde{{\widetilde\bnu}}
\def\bxitilde{{\widetilde\bxi}}
\def\bomicrontilde{{\widetilde\bomicron}}
\def\bpitilde{{\widetilde\bpi}}
\def\brhotilde{{\widetilde\brho}}
\def\bsigmatilde{{\widetilde\bsigma}}
\def\btautilde{{\widetilde\btau}}
\def\bupsilontilde{{\widetilde\bupsilon}}
\def\bphitilde{{\widetilde\bphi}}
\def\bchitilde{{\widetilde\bchi}}
\def\bpsitilde{{\widetilde\bpsi}}
\def\bomegatilde{{\widetilde\bomega}}
\def\bAlphatilde{{\widetilde\bAlpha}}
\def\bBetatilde{{\widetilde\bBeta}}
\def\bGammatilde{{\widetilde\bGamma}}
\def\bDeltatilde{{\widetilde\bDelta}}
\def\bEpsilontilde{{\widetilde\bEpsilon}}
\def\bZetatilde{{\widetilde\bZeta}}
\def\bEtatilde{{\widetilde\bEta}}
\def\bThetatilde{{\widetilde\bTheta}}
\def\bIotatilde{{\widetilde\bIota}}
\def\bKappatilde{{\widetilde\bKappa}}
\def\bLambdatilde{{\widetilde\bLambda}}
\def\bMutilde{{\widetilde\bMu}}
\def\bNutilde{{\widetilde\bNu}}
\def\bXitilde{{\widetilde\bXi}}
\def\bOmicrontilde{{\widetilde\bOmicron}}
\def\bPitilde{{\widetilde\bPi}}
\def\bRhotilde{{\widetilde\bRho}}
\def\bSigmatilde{{\widetilde\bSigma}}
\def\bTautilde{{\widetilde\bTau}}
\def\bUpsilontilde{{\widetilde\bUpsilon}}
\def\bPhitilde{{\widetilde\bPhi}}
\def\bChitilde{{\widetilde\bChi}}
\def\bPsitilde{{\widetilde\bPsi}}
\def\bOmegatilde{{\widetilde\bOmega}}
%
%
%
%
%
\def\abar{\bar{ a}}
\def\bbar{\bar{ b}}
\def\cbar{\bar{ c}}
\def\dbar{\bar{ d}}
\def\ebar{\bar{ e}}
\def\fbar{\bar{ f}}
\def\gbar{\bar{ g}}
\def\hbar{\bar{ h}}
\def\ibar{\bar{ i}}
\def\jbar{\bar{ j}}
\def\kbar{\bar{ k}}
\def\lbar{\bar{ l}}
\def\mbar{\bar{ m}}
\def\nbar{\bar{ n}}
\def\obar{\bar{ o}}
\def\pbar{\bar{ p}}
\def\qbar{\bar{ q}}
\def\rbar{\bar{ r}}
\def\sbar{\bar{ s}}
\def\tbar{\bar{ t}}
\def\ubar{\bar{ u}}
\def\vbar{\bar{ v}}
\def\wbar{\bar{ w}}
\def\xbar{\bar{ x}}
\def\ybar{\bar{ y}}
\def\zbar{\bar{ z}}
\def\Abar{\bar{ A}}
\def\Bbar{\bar{ B}}
\def\Cbar{\bar{ C}}
\def\Dbar{\bar{ D}}
\def\Ebar{\bar{ E}}
\def\Fbar{\bar{ F}}
\def\Gbar{\bar{ G}}
\def\Hbar{\bar{ H}}
\def\Ibar{\bar{ I}}
\def\Jbar{\bar{ J}}
\def\Kbar{\bar{ K}}
\def\Lbar{\bar{ L}}
\def\Mbar{\bar{ M}}
\def\Nbar{\bar{ N}}
\def\Obar{\bar{ O}}
\def\Pbar{\bar{ P}}
\def\Qbar{\bar{ Q}}
\def\Rbar{\bar{ R}}
\def\Sbar{\bar{ S}}
\def\Tbar{\bar{ T}}
\def\Ubar{\bar{ U}}
\def\Vbar{\bar{ V}}
\def\Wbar{\bar{ W}}
\def\Xbar{\bar{ X}}
\def\Ybar{\bar{ Y}}
\def\Zbar{\bar{ Z}}
%
%
%
%
%
\def\babar{\overline{ \ba}}
\def\bbbar{\overline{ \bb}}
\def\bcbar{\overline{ \bc}}
\def\bdbar{\overline{ \bd}}
\def\bebar{\overline{ \be}}
\def\bfbar{\overline{ \bf}}
\def\bgbar{\overline{ \bg}}
\def\bhbar{\overline{ \bh}}
\def\bibar{\overline{ \bi}}
\def\bjbar{\overline{ \bj}}
\def\bkbar{\overline{ \bk}}
\def\blbar{\overline{ \bl}}
\def\bmbar{\overline{ \bm}}
\def\bnbar{\overline{ \bn}}
\def\bobar{\overline{ \bo}}
\def\bpbar{\overline{ \bp}}
\def\bqbar{\overline{ \bq}}
\def\brbar{\overline{ \br}}
\def\bsbar{\overline{ \bs}}
\def\btbar{\overline{ \bt}}
\def\bubar{\overline{ \bu}}
\def\bvbar{\overline{ \bv}}
\def\bwbar{\overline{ \bw}}
\def\bxbar{\overline{ \bx}}
\def\bybar{\overline{ \by}}
\def\bzbar{\overline{ \bz}}
\def\bAbar{\overline{ \bA}}
\def\bBbar{\overline{ \bB}}
\def\bCbar{\overline{ \bC}}
\def\bDbar{\overline{ \bD}}
\def\bEbar{\overline{ \bE}}
\def\bFbar{\overline{ \bF}}
\def\bGbar{\overline{ \bG}}
\def\bHbar{\overline{ \bH}}
\def\bIbar{\overline{ \bI}}
\def\bJbar{\overline{ \bJ}}
\def\bKbar{\overline{ \bK}}
\def\bLbar{\overline{ \bL}}
\def\bMbar{\overline{ \bM}}
\def\bNbar{\overline{ \bN}}
\def\bObar{\overline{ \bO}}
\def\bPbar{\overline{ \bP}}
\def\bQbar{\overline{ \bQ}}
\def\bRbar{\overline{ \bR}}
\def\bSbar{\overline{ \bS}}
\def\bTbar{\overline{ \bT}}
\def\bUbar{\overline{ \bU}}
\def\bVbar{\overline{ \bV}}
\def\bWbar{\overline{ \bW}}
\def\bXbar{\overline{ \bX}}
\def\bYbar{\overline{ \bY}}
\def\bZbar{\overline{ \bZ}}
%
%

%
%
%
\def\asc{{\cal a}}
\def\bsc{{\cal b}}
\def\csc{{\cal c}}
\def\dsc{{\cal d}}
\def\esc{{\cal e}}
\def\dsc{{\cal f}}
\def\gsc{{\cal g}}
\def\hsc{{\cal h}}
\def\isc{{\cal i}}
\def\jsc{{\cal j}}
\def\ksc{{\cal k}}
\def\lsc{{\cal l}}
\def\msc{{\cal m}}
\def\nsc{{\cal n}}
\def\osc{{\cal o}}
\def\psc{{\cal p}}
\def\qsc{{\cal q}}
\def\rsc{{\cal r}}
\def\ssc{{\cal s}}
\def\tsc{{\cal t}}
\def\usc{{\cal u}}
\def\vsc{{\cal v}}
\def\wsc{{\cal w}}
\def\xsc{{\cal x}}
\def\ysc{{\cal y}}
\def\zsc{{\cal z}}
\def\Asc{{\cal A}}
\def\Bsc{{\cal B}}
\def\Csc{{\cal C}}
\def\Dsc{{\cal D}}
\def\Esc{{\cal E}}
\def\Fsc{{\cal F}}
\def\Gsc{{\cal G}}
\def\Hsc{{\cal H}}
\def\Isc{{\cal I}}
\def\Jsc{{\cal J}}
\def\Ksc{{\cal K}}
\def\Lsc{{\cal L}}
\def\Msc{{\cal M}}
\def\Nsc{{\cal N}}
\def\Osc{{\cal O}}
\def\Psc{{\cal P}}
\def\Qsc{{\cal Q}}
\def\Rsc{{\cal R}}
\def\Ssc{{\cal S}}
\def\Tsc{{\cal T}}
\def\Usc{{\cal U}}
\def\Vsc{{\cal V}}
\def\Wsc{{\cal W}}
\def\Xsc{{\cal X}}
\def\Ysc{{\cal Y}}
\def\Zsc{{\cal Z}}
\def\Aschat{\widehat{{\cal A}}}
\def\Bschat{\widehat{{\cal B}}}
\def\Cschat{\widehat{{\cal C}}}
\def\Dschat{\widehat{{\cal D}}}
\def\Eschat{\widehat{{\cal E}}}
\def\Fschat{\widehat{{\cal F}}}
\def\Gschat{\widehat{{\cal G}}}
\def\Hschat{\widehat{{\cal H}}}
\def\Ischat{\widehat{{\cal I}}}
\def\Jschat{\widehat{{\cal J}}}
\def\Kschat{\widehat{{\cal K}}}
\def\Lschat{\widehat{{\cal L}}}
\def\Mschat{\widehat{{\cal M}}}
\def\Nschat{\widehat{{\cal N}}}
\def\Oschat{\widehat{{\cal O}}}
\def\Pschat{\widehat{{\cal P}}}
\def\Qschat{\widehat{{\cal Q}}}
\def\Rschat{\widehat{{\cal R}}}
\def\Sschat{\widehat{{\cal S}}}
\def\Tschat{\widehat{{\cal T}}}
\def\Uschat{\widehat{{\cal U}}}
\def\Vschat{\widehat{{\cal V}}}
\def\Wschat{\widehat{{\cal W}}}
\def\Xschat{\widehat{{\cal X}}}
\def\Yschat{\widehat{{\cal Y}}}
\def\Zschat{\widehat{{\cal Z}}}
\def\Asctilde{\widetilde{{\cal A}}}
\def\Bsctilde{\widetilde{{\cal B}}}
\def\Csctilde{\widetilde{{\cal C}}}
\def\Dsctilde{\widetilde{{\cal D}}}
\def\Esctilde{\widetilde{{\cal E}}}
\def\Fsctilde{\widetilde{{\cal F}}}
\def\Gsctilde{\widetilde{{\cal G}}}
\def\Hsctilde{\widetilde{{\cal H}}}
\def\Isctilde{\widetilde{{\cal I}}}
\def\Jsctilde{\widetilde{{\cal J}}}
\def\Ksctilde{\widetilde{{\cal K}}}
\def\Lsctilde{\widetilde{{\cal L}}}
\def\Msctilde{\widetilde{{\cal M}}}
\def\Nsctilde{\widetilde{{\cal N}}}
\def\Osctilde{\widetilde{{\cal O}}}
\def\Psctilde{\widetilde{{\cal P}}}
\def\Qsctilde{\widetilde{{\cal Q}}}
\def\Rsctilde{\widetilde{{\cal R}}}
\def\Ssctilde{\widetilde{{\cal S}}}
\def\Tsctilde{\widetilde{{\cal T}}}
\def\Usctilde{\widetilde{{\cal U}}}
\def\Vsctilde{\widetilde{{\cal V}}}
\def\Wsctilde{\widetilde{{\cal W}}}
\def\Xsctilde{\widetilde{{\cal X}}}
\def\Ysctilde{\widetilde{{\cal Y}}}
\def\Zsctilde{\widetilde{{\cal Z}}}
\def\bAsc{\mathbf{\cal A}}
\def\bBsc{\mathbf{\cal B}}
\def\bCsc{\mathbf{\cal C}}
\def\bDsc{\mathbf{\cal D}}
\def\bEsc{\mathbf{\cal E}}
\def\bFsc{\mathbf{\cal F}}
\def\bGsc{\mathbf{\cal G}}
\def\bHsc{\mathbf{\cal H}}
\def\bIsc{\mathbf{\cal I}}
\def\bJsc{\mathbf{\cal J}}
\def\bKsc{\mathbf{\cal K}}
\def\bLsc{\mathbf{\cal L}}
\def\bMsc{\mathbf{\cal M}}
\def\bNsc{\mathbf{\cal N}}
\def\bOsc{\mathbf{\cal O}}
\def\bPsc{\mathbf{\cal P}}
\def\bQsc{\mathbf{\cal Q}}
\def\bRsc{\mathbf{\cal R}}
\def\bSsc{\mathbf{\cal S}}
\def\bTsc{\mathbf{\cal T}}
\def\bUsc{\mathbf{\cal U}}
\def\bVsc{\mathbf{\cal V}}
\def\bWsc{\mathbf{\cal W}}
\def\bXsc{\mathbf{\cal X}}
\def\bYsc{\mathbf{\cal Y}}
\def\bZsc{\mathbf{\cal Z}}
\def\bAschat{\widehat{\mathbf{\cal A}}}
\def\bBschat{\widehat{\mathbf{\cal B}}}
\def\bCschat{\widehat{\mathbf{\cal C}}}
\def\bDschat{\widehat{\mathbf{\cal D}}}
\def\bEschat{\widehat{\mathbf{\cal E}}}
\def\bFschat{\widehat{\mathbf{\cal F}}}
\def\bGschat{\widehat{\mathbf{\cal G}}}
\def\bHschat{\widehat{\mathbf{\cal H}}}
\def\bIschat{\widehat{\mathbf{\cal I}}}
\def\bJschat{\widehat{\mathbf{\cal J}}}
\def\bKschat{\widehat{\mathbf{\cal K}}}
\def\bLschat{\widehat{\mathbf{\cal L}}}
\def\bMschat{\widehat{\mathbf{\cal M}}}
\def\bNschat{\widehat{\mathbf{\cal N}}}
\def\bOschat{\widehat{\mathbf{\cal O}}}
\def\bPschat{\widehat{\mathbf{\cal P}}}
\def\bQschat{\widehat{\mathbf{\cal Q}}}
\def\bRschat{\widehat{\mathbf{\cal R}}}
\def\bSschat{\widehat{\mathbf{\cal S}}}
\def\bTschat{\widehat{\mathbf{\cal T}}}
\def\bUschat{\widehat{\mathbf{\cal U}}}
\def\bVschat{\widehat{\mathbf{\cal V}}}
\def\bWschat{\widehat{\mathbf{\cal W}}}
\def\bXschat{\widehat{\mathbf{\cal X}}}
\def\bYschat{\widehat{\mathbf{\cal Y}}}
\def\bZschat{\widehat{\mathbf{\cal Z}}}
\def\afrak{\mathfrak{a}}
\def\bfrak{\mathfrak{b}}
\def\cfrak{\mathfrak{c}}
\def\dfrak{\mathfrak{d}}
\def\efrak{\mathfrak{e}}
\def\ffrak{\mathfrak{f}}
\def\gfrak{\mathfrak{g}}
\def\hfrak{\mathfrak{h}}
\def\ifrak{\mathfrak{i}}
\def\jfrak{\mathfrak{j}}
\def\kfrak{\mathfrak{k}}
\def\lfrak{\mathfrak{l}}
\def\mfrak{\mathfrak{m}}
\def\nfrak{\mathfrak{n}}
\def\ofrak{\mathfrak{o}}
\def\pfrak{\mathfrak{p}}
\def\qfrak{\mathfrak{q}}
\def\rfrak{\mathfrak{r}}
\def\sfrak{\mathfrak{s}}
\def\tfrak{\mathfrak{t}}
\def\ufrak{\mathfrak{u}}
\def\vfrak{\mathfrak{v}}
\def\wfrak{\mathfrak{w}}
\def\xfrak{\mathfrak{x}}
\def\yfrak{\mathfrak{y}}
\def\zfrak{\mathfrak{z}}
\def\Afrak{\mathfrak{ A}}
\def\Bfrak{\mathfrak{ B}}
\def\Cfrak{\mathfrak{ C}}
\def\Dfrak{\mathfrak{ D}}
\def\Efrak{\mathfrak{ E}}
\def\Ffrak{\mathfrak{ F}}
\def\Gfrak{\mathfrak{ G}}
\def\Hfrak{\mathfrak{ H}}
\def\Ifrak{\mathfrak{ I}}
\def\Jfrak{\mathfrak{ J}}
\def\Kfrak{\mathfrak{ K}}
\def\Lfrak{\mathfrak{ L}}
\def\Mfrak{\mathfrak{ M}}
\def\Nfrak{\mathfrak{ N}}
\def\Ofrak{\mathfrak{ O}}
\def\Pfrak{\mathfrak{ P}}
\def\Qfrak{\mathfrak{ Q}}
\def\Rfrak{\mathfrak{ R}}
\def\Sfrak{\mathfrak{ S}}
\def\Tfrak{\mathfrak{ T}}
\def\Ufrak{\mathfrak{ U}}
\def\Vfrak{\mathfrak{ V}}
\def\Wfrak{\mathfrak{ W}}
\def\Xfrak{\mathfrak{ X}}
\def\Yfrak{\mathfrak{ Y}}
\def\Zfrak{\mathfrak{ Z}}

\def\bAfrak{\mathbf{\mathfrak{A}}}
\def\bBfrak{\mathbf{\mathfrak{B}}}
\def\bCfrak{\mathbf{\mathfrak{C}}}
\def\bDfrak{\mathbf{\mathfrak{D}}}
\def\bEfrak{\mathbf{\mathfrak{E}}}
\def\bFfrak{\mathbf{\mathfrak{F}}}
\def\bGfrak{\mathbf{\mathfrak{G}}}
\def\bHfrak{\mathbf{\mathfrak{H}}}
\def\bIfrak{\mathbf{\mathfrak{I}}}
\def\bJfrak{\mathbf{\mathfrak{J}}}
\def\bKfrak{\mathbf{\mathfrak{K}}}
\def\bLfrak{\mathbf{\mathfrak{L}}}
\def\bMfrak{\mathbf{\mathfrak{M}}}
\def\bNfrak{\mathbf{\mathfrak{N}}}
\def\bOfrak{\mathbf{\mathfrak{O}}}
\def\bPfrak{\mathbf{\mathfrak{P}}}
\def\bQfrak{\mathbf{\mathfrak{Q}}}
\def\bRfrak{\mathbf{\mathfrak{R}}}
\def\bSfrak{\mathbf{\mathfrak{S}}}
\def\bTfrak{\mathbf{\mathfrak{T}}}
\def\bUfrak{\mathbf{\mathfrak{U}}}
\def\bVfrak{\mathbf{\mathfrak{V}}}
\def\bWfrak{\mathbf{\mathfrak{W}}}
\def\bXfrak{\mathbf{\mathfrak{X}}}
\def\bYfrak{\mathbf{\mathfrak{Y}}}
\def\bZfrak{\mathbf{\mathfrak{Z}}}

\def\bAfrakhat{\mathbf{\widehat{\mathfrak{A}}}}
\def\bBfrakhat{\mathbf{\widehat{\mathfrak{B}}}}
\def\bCfrakhat{\mathbf{\widehat{\mathfrak{C}}}}
\def\bDfrakhat{\mathbf{\widehat{\mathfrak{D}}}}
\def\bEfrakhat{\mathbf{\widehat{\mathfrak{E}}}}
\def\bFfrakhat{\mathbf{\widehat{\mathfrak{F}}}}
\def\bGfrakhat{\mathbf{\widehat{\mathfrak{G}}}}
\def\bHfrakhat{\mathbf{\widehat{\mathfrak{H}}}}
\def\bIfrakhat{\mathbf{\widehat{\mathfrak{I}}}}
\def\bJfrakhat{\mathbf{\widehat{\mathfrak{J}}}}
\def\bKfrakhat{\mathbf{\widehat{\mathfrak{K}}}}
\def\bLfrakhat{\mathbf{\widehat{\mathfrak{L}}}}
\def\bMfrakhat{\mathbf{\widehat{\mathfrak{M}}}}
\def\bNfrakhat{\mathbf{\widehat{\mathfrak{N}}}}
\def\bOfrakhat{\mathbf{\widehat{\mathfrak{O}}}}
\def\bPfrakhat{\mathbf{\widehat{\mathfrak{P}}}}
\def\bQfrakhat{\mathbf{\widehat{\mathfrak{Q}}}}
\def\bRfrakhat{\mathbf{\widehat{\mathfrak{R}}}}
\def\bSfrakhat{\mathbf{\widehat{\mathfrak{S}}}}
\def\bTfrakhat{\mathbf{\widehat{\mathfrak{T}}}}
\def\bUfrakhat{\mathbf{\widehat{\mathfrak{U}}}}
\def\bVfrakhat{\mathbf{\widehat{\mathfrak{V}}}}
\def\bWfrakhat{\mathbf{\widehat{\mathfrak{W}}}}
\def\bXfrakhat{\mathbf{\widehat{\mathfrak{X}}}}
\def\bYfrakhat{\mathbf{\widehat{\mathfrak{Y}}}}
\def\bZfrakhat{\mathbf{\widehat{\mathfrak{Z}}}}
%
%
%
%
\def\etal{{\em et al.}}
%
%
%
%
%
\def\cumsum{\mbox{cumsum}}
\def\real{{\mathbb R}}
\def\intinfinf{\int_{-\infty}^{\infty}}
\def\intzinf{\int_{0}^{\infty}}
\def\intzt{\int_0^t}
\def\transpose{{\sf \scriptscriptstyle{T}}}
\def\smhalf{{\textstyle{1\over2}}}
\def\third{{\textstyle{1\over3}}}
\def\twothirds{{\textstyle{2\over3}}}
\def\bell{\bmath{\ell}}
\def\half{\frac{1}{2}}
\def\ninv{n^{-1}}
\def\nhalf{n^{\half}}
\def\mhalf{m^{\half}}
\def\nnhalf{n^{-\half}}
\def\mnhalf{m^{-\half}}
\def\MN{\mbox{MN}}
\def\N{\mbox{N}}
\def\E{\mbox{E}}
\def\pr{P}
\def\var{\mbox{var}}
\def\limn{\lim_{n\to \infty} }
\def\intt{\int_{\tau_a}^{\tau_b}}
\def\sumin{\sum_{i=1}^n}
\def\sumjn{\sum_{j=1}^n}
\def\SUMin{{\displaystyle \sum_{i=1}^n}}
\def\SUMjn{{\displaystyle \sum_{j=1}^n}}
\def\myendthm{\begin{flushright} $\diamond $ \end{flushright}}
\def\convd{\overset{\Dsc}{\longrightarrow}}
\def\convp{\overset{\Psc}{\longrightarrow}}
\def\convas{\overset{a.s.}{\longrightarrow}}
\def\hn{\mbox{H}_0}
\def\ha{\mbox{H}_1}

%
%
%
%
%
\def\trans{^{\transpose}}
\def\inv{^{-1}}
\def\twobyone#1#2{\left[
\begin{array}
{c}
#1\\
#2\\
\end{array}
\right]}
%
%
%
%
%
\def\argmindum{\mathop{\mbox{argmin}}}
\def\argmin#1{\argmindum_{#1}}
\def\argmaxdum{\mathop{\mbox{argmax}}}
\def\argmax#1{\argmaxdum_{#1}}
\def\blockdiag{\mbox{blockdiag}}
\def\corr{\mbox{corr}}
\def\cov{\mbox{cov}}
\def\diag{\mbox{diag}}
\def\dffit{df_{{\rm fit}}}
\def\dfres{df_{{\rm res}}}
\def\dfyhat{df_{\yhat}}
\def\diag{\mbox{diag}}
\def\diagonal{\mbox{diagonal}}
\def\logit{\mbox{logit}}
\def\stdev{\mbox{st.\,dev.}}
\def\stdevhat{{\widehat{\mbox{st.dev}}}}
\def\tr{\mbox{tr}}
\def\trigamma{\mbox{trigamma}}
\def\var{\mbox{var}}
\def\vecof{\mbox{vec}}
\def\AIC{\mbox{AIC}}
\def\AMISE{\mbox{AMISE}}
\def\Corr{\mbox{Corr}}
\def\Cov{\mbox{Cov}}
\def\CV{\mbox{CV}}
\def\GCV{\mbox{GCV}}
\def\LR{\mbox{LR}}
\def\MISE{\mbox{MISE}}
\def\MSSE{\mbox{MSSE}}
\def\ML{\mbox{ML}}
\def\REML{\mbox{REML}}
\def\RMSE{{\rm RMSE}}
\def\RSS{\mbox{RSS}}
\def\Var{\mbox{Var}}
%
%
%
%
\def\bib{\vskip12pt\par\noindent\hangindent=1 true cm\hangafter=1}
\def\jump{\vskip3mm\noindent}
\def\mybox#1{\vskip1mm \begin{center}
        \hspace{.0\textwidth}\vbox{\hrule\hbox{\vrule\kern6pt
\parbox{.9\textwidth}{\kern6pt#1\vskip6pt}\kern6pt\vrule}\hrule}
        \end{center} \vskip-5mm}
\def\lboxit#1{\vbox{\hrule\hbox{\vrule\kern6pt
      \vbox{\kern6pt#1\vskip6pt}\kern6pt\vrule}\hrule}}
\def\boxit#1{\begin{center}\fbox{#1}\end{center}}
\def\thickboxit#1{\vbox{{\hrule height 1mm}\hbox{{\vrule width 1mm}\kern6pt
          \vbox{\kern6pt#1\kern6pt}\kern6pt{\vrule width 1mm}}
               {\hrule height 1mm}}}
\def\instep{\vskip12pt\par\hangindent=30 true mm\hangafter=1}
\def\uWand{\underline{Wand}}
\def\remtask#1#2{\mmnote{\thickboxit
                 {\bf #1\ \theremtask}}\refstepcounter{remtask}}
%
%
%

%
%
\def\aism{{\it Ann. Inst. Statist. Math.}\ }
\def\ajs{{\it Austral. J. Statist.}\ }
\def\ANNSTAT{{\it The Annals of Statistics}\ }
\def\annmath{{\it Ann. Math. Statist.}\ }
\def\applstat{{\it Appl. Statist.}\ }
\def\BIOMETRICS{{\it Biometrics}\ }
\def\cjs{{\it Canad. J. Statist.}\ }
\def\csda{{\it Comp. Statist. Data Anal.}\ }
\def\cstm{{\it Comm. Statist. Theory Meth.}\ }
\def\ieeetit{{\it IEEE Trans. Inf. Theory}\ }
\def\isr{{\it Internat. Statist. Rev.}\ }
\def\JASA{{\it Journal of the American Statistical Association}\ }
\def\JCGS{{\it Journal of Computational and Graphical Statistics}\ }
\def\jscs{{\it J. Statist. Comput. Simulation}\ }
\def\jma{{\it J. Multivariate Anal.}\ }
\def\jns{{\it J. Nonparametric Statist.}\ }
\def\JRSSA{{\it Journal of the Royal Statistics Society, Series A}\ }
\def\JRSSB{{\it Journal of the Royal Statistics Society, Series B}\ }
\def\JRSSC{{\it Journal of the Royal Statistics Society, Series C}\ }
\def\jspi{{\it J. Statist. Planning Inference}\ }
\def\ptrf{{\it Probab. Theory Rel. Fields}\ }
\def\sankhyaa{{\it Sankhy$\bar{{\it a}}$} Ser. A\ }
\def\sjs{{\it Scand. J. Statist.}\ }
\def\spl{{\it Statist. Probab. Lett.}\ }
\def\statsci{{\it Statist. Sci.}\ }
\def\techno{{\it Technometrics}\ }
\def\tpa{{\it Theory Probab. Appl.}\ }
\def\zw{{\it Z. Wahr. ver. Geb.}\ }
%
%
%
%
\def\Brent{{\bf BRENT:}\ }
\def\David{{\bf DAVID:}\ }
\def\Erin{{\bf ERIN:}}
\def\Gerda{{\bf GERDA:}\ }
\def\Joel{{\bf JOEL:}\ }
\def\Marc{{\bf MARC:}\ }
\def\Matt{{\bf MATT:}\ }
\def\Tianxi{{\bf TIANXI:}\ }
%
%
%
%
\def\bZE{\bZ_{\scriptscriptstyle E}}
\def\bZT{\bZ_{\scriptscriptstyle T}}
\def\bbE{\bb_{\scriptscriptstyle E}}
\def\bbT{\bb_{\scriptscriptstyle T}}
\def\bbhatT{\bbhat_{\scriptscriptstyle T}}
\def\fX{f_{\scriptscriptstyle X}}
\def\sigeps{\sigma_{\varepsilon}}
\def\bVtheta{\bV_{\smbtheta}}
\def\bVthetainv{\bVtheta^{-1}}
\def\bKsc{\boldsymbol{\Ksc}}
\def\bxbar{\bar{\bx}}
\def\bPL{b^{\scriptscriptstyle{\rm PL}}}
\def\bVA{b^{\scriptscriptstyle{\rm VA}}}
\def\zPL{z^{\scriptscriptstyle{\rm PL}}}
\def\zVA{z^{\scriptscriptstyle{\rm VA}}}
\def\bYmis{\bY_{\scriptscriptstyle{\rm mis}}}
\def\bYmishat{{\widehat{\bYmis}}}
\def\bYmisone{\bY_{\scriptscriptstyle{\rm mis,1}}}
\def\bYmistwo{\bY_{\scriptscriptstyle{\rm mis,2}}}
\def\bYobs{\bY_{\scriptscriptstyle{\rm obs}}}
\def\bdobs{\bd_{\scriptscriptstyle{\rm obs}}}
\def\bdmis{\bd_{\scriptscriptstyle{\rm mis}}}
%
%
%
%
\def\bfDelta{{\mbox{\boldmath$\Delta$}}}
\def\bfkappa{{\mbox{\boldmath$\kappa$}}}
\def\bfgamma{{\mbox{\boldmath$\gamma$}}}
\def\bftheta{{\mbox{\boldmath$\theta$}}}
\def\bfmu{{\mbox{\boldmath$\mu$}}}
\def\bfdelta{{\mbox{\boldmath$\delta$}}}
\def\bfeps{{\mbox{\boldmath$\varepsilon$}}}
\def\bfnu{{\mbox{\boldmath$\nu$}}}
\def\bfzeta{{\mbox{\boldmath$\zeta$}}}
\def\bfchi{{\mbox{\boldmath$\chi$}}}
\def\bbX{\mathbb{X}}
\def\bbV{\mathbb{V}} 
\def\bbA{\mathbb{A}}
\def\bbB{\mathbb{B}}
\def\bbK{\mathbb{K}}
\def\bbP{\mathbb{P}}
\def\bbD{\mathbb{D}}

\def\Abb{\mathbb{A}}
\def\Bbb{\mathbb{B}}
\def\Cbb{\mathbb{C}}
\def\Dbb{\mathbb{D}}
\def\Ebb{\mathbb{E}}
\def\Fbb{\mathbb{F}}
\def\Gbb{\mathbb{G}}
\def\Hbb{\mathbb{H}}
\def\Ibb{\mathbb{I}}
\def\Jbb{\mathbb{J}}
\def\Kbb{\mathbb{K}}
\def\Lbb{\mathbb{L}}
\def\Mbb{\mathbb{M}}
\def\Nbb{\mathbb{N}}
\def\Mbb{\mathbb{M}}
\def\Nbb{\mathbb{N}}
\def\Obb{\mathbb{O}}
\def\Pbb{\mathbb{P}}
\def\Qbb{\mathbb{Q}}
\def\Rbb{\mathbb{R}}
\def\Sbb{\mathbb{S}}
\def\Tbb{\mathbb{T}}
\def\Ubb{\mathbb{U}}
\def\Vbb{\mathbb{V}}
\def\Wbb{\mathbb{W}}
\def\Xbb{\mathbb{X}}
\def\Ybb{\mathbb{Y}}
\def\Zbb{\mathbb{Z}}

\def\Abbtilde{\widetilde{\mathbb{A}}}
\def\Bbbtilde{\widetilde{\mathbb{B}}}
\def\Cbbtilde{\widetilde{\mathbb{C}}}
\def\Dbbtilde{\widetilde{\mathbb{D}}}
\def\Ebbtilde{\widetilde{\mathbb{E}}}
\def\Fbbtilde{\widetilde{\mathbb{F}}}
\def\Gbbtilde{\widetilde{\mathbb{G}}}
\def\Hbbtilde{\widetilde{\mathbb{H}}}
\def\Ibbtilde{\widetilde{\mathbb{I}}}
\def\Jbbtilde{\widetilde{\mathbb{J}}}
\def\Kbbtilde{\widetilde{\mathbb{K}}}
\def\Lbbtilde{\widetilde{\mathbb{L}}}
\def\Mbbtilde{\widetilde{\mathbb{M}}}
\def\Nbbtilde{\widetilde{\mathbb{N}}}
\def\Mbbtilde{\widetilde{\mathbb{M}}}
\def\Nbbtilde{\widetilde{\mathbb{N}}}
\def\Obbtilde{\widetilde{\mathbb{O}}}
\def\Pbbtilde{\widetilde{\mathbb{P}}}
\def\Qbbtilde{\widetilde{\mathbb{Q}}}
\def\Rbbtilde{\widetilde{\mathbb{R}}}
\def\Sbbtilde{\widetilde{\mathbb{S}}}
\def\Tbbtilde{\widetilde{\mathbb{T}}}
\def\Ubbtilde{\widetilde{\mathbb{U}}}
\def\Vbbtilde{\widetilde{\mathbb{V}}}
\def\Wbbtilde{\widetilde{\mathbb{W}}}
\def\Xbbtilde{\widetilde{\mathbb{X}}}
\def\Ybbtilde{\widetilde{\mathbb{Y}}}
\def\Zbbtilde{\widetilde{\mathbb{Z}}}

%
%
%
%
\def\miss{\mbox{{\tiny miss}}}
\def\obs{\scriptsize{\mbox{obs}}}

%
%
%
%
\def\bmath#1{\mbox{\boldmath$#1$}}
\def\fat#1{\hbox{\rlap{$#1$}\kern0.25pt\rlap{$#1$}\kern0.25pt$#1$}}
\def\wh{\widehat}
\def\flambda{\fat{\lambda}}
\def\beps{\bmath{\varepsilon}}
\def\bSlambda{\bS_{\lambda}}
\def\ErrorSS{\mbox{RSS}}
\def\bsqbar{\bar{{b^2}}}
\def\bcubar{\bar{{b^3}}}
\def\plargest{p_{\rm \,largest}}
\def\summheading#1{\subsection*{#1}\hskip3mm}
\def\summbreak{\vskip3mm\par}
\def\df{df}
\def\adf{adf}
\def\dffit{df_{{\rm fit}}}
\def\dfres{df_{{\rm res}}}
\def\dfyhat{df_{\yhat}}
\def\sigb{\sigma_b}
\def\sigu{\sigma_u}
\def\sigepshat{{\widehat\sigma}_{\varepsilon}}
\def\siguhat{{\widehat\sigma}_u}
\def\sigepshat{{\widehat\sigma}_{\varepsilon}}
\def\sigbhat{{\widehat\sigma}_b}
\def\sighat{{\widehat\sigma}}
\def\sigsqb{\sigma^2_b}
\def\sigsqeps{\sigma^2_{\varepsilon}}
\def\sigsqepszerohat{{\widehat\sigma}^2_{\varepsilon,0}}
\def\sigsqepshat{{\widehat\sigma}^2_{\varepsilon}}
\def\sigsqbhat{{\widehat\sigma}^2_b}
\def\dfnumer{{\rm df(II}|{\rm I)}}
\def\mhatlam{{\widehat m}_{\lambda}}
\def\calD{\Dsc}
\def\Aeps{A_{\epsilon}}
\def\Beps{B_{\epsilon}}
\def\Ab{A_b}
\def\Bb{B_b}
\def\bXtmain{\tilde{\bX}_r}
\def\main{\mbox{\tt main}}
\def\argminbetab{\argmin{\bbeta,\bb}}
\def\calB{\Bsc}
\def\respvar{\mbox{\tt log(amt)}}

\def\Abb{\mathbb{A}}
\def\Zbb{\mathbb{Z}}
\def\Wbb{\mathbb{W}}
\def\Wbbhat{\widehat{\mathbb{W}}}
\def\Kbbtilde{\widetilde{\mathbb{K}}}
\def\Pbbtilde{\widetilde{\mathbb{P}}}
\def\Dbbtilde{\widetilde{\mathbb{D}}}
\def\Bbbtilde{\widetilde{\mathbb{B}}}

\def\Abbhat{\widehat{\mathbb{A}}}

\def\ellhat{\widehat{\ell}}
\def\pn{\phantom{-}}
\def\pp{\phantom{1}}

\def\PP{\stackrel{P}{\rightarrow}}
\def\DD{\Rightarrow}
%
%

{
\let\newpage\relax
\maketitle
}
\hrule
\bigskip
\textbf{Correspondence to:}\\
Jessica Gronsbell \\
Postal address: 700 University Ave, Toronto, ON, Canada, M5G 1Z5\\
Email: \url{j.gronsbell@utoronto.ca}.\\
Telephone number: 416-978-3452\\
\hrule
\bigskip
\textbf{Keywords:} Fairness; Machine Learning; Predictive Models; \blue{Review} \\

\newpage

\section*{Abstract}
{\blue{Ensuring that machine learning (ML) models are safe, effective, and equitable across all patients is critical for clinical decision-making and for preventing the amplification of existing health disparities. In this work, we examine how fairness is conceptualized in ML for health, including why ML models may lead to unfair decisions and how fairness has been measured in diverse real-world applications. We review commonly used fairness notions within group, individual, and causal-based frameworks.  We also discuss the outlook for future research and highlight opportunities and challenges in operationalizing fairness in health-focused applications.}}  

\newpage

\section{Introduction}
There are myriad potential applications of machine learning (ML) for health\footnote{\blue{``ML for health” is a moniker used within the computer science literature to refer to the application of ML to healthcare and biomedical domains to improve patient outcomes, assist in clinical decision-making, optimize healthcare processes, and advance scientific research \cite{ghassemi_review_2020,adibi_recent_2025}. }}, including automated disease detection, computer-aided diagnosis, and personalized treatment planning \cite{rajpurkar2022ai}.  However, there is substantial evidence that, without appropriate forethought and planning, ML models can introduce or exacerbate health inequities by making less accurate decisions for certain groups or individuals \cite{grote_enabling_2022}. Within medical imaging, state-of-the-art ML models used for disease diagnosis, risk prediction, and triage management are known to underperform within minority groups defined by protected attributes, including sex, race, and ethnicity \cite{rajpurkar_chexnet_2017,ting_development_2017, liu_comparison_2019,lee_machine_2020, seyyed-kalantari_chexclusion_2020,larrazabal_gender_2020, seyyed-kalantari_underdiagnosis_2021,puyol-anton_fairness_2022, bressem_deep_2021, roest_comparative_2022, lin_evaluate_2023}.  For example, deep learning models used to detect 14 common diseases from chest X-rays were found to under-diagnose under-served subgroups \blue{occupying intersections of oppression}, such as Hispanic female patients, potentially resulting in treatment delays if deployed in practice \cite{seyyed-kalantari_underdiagnosis_2021}.  {\blue{Similarly, ML models trained on electronic health records (EHR), administrative claims, and genomic data have shown biased performance, often making less accurate predictions for certain subgroups. \cite{li_evaluating_2023, banda_characterizing_2023}. A landmark study revealed that a widely used commercial risk prediction tool for identifying patients with complex health needs exhibited significantly lower accuracy for Black patients compared to White patients. As a result, Black patients with similar levels of illness were less likely to be recommended for essential care services \cite{obermeyer_dissecting_2019}. In genomics, polygenic risk scores frequently perform less accurately for individuals of non-European ancestry due to their historical underrepresentation in genomic datasets, which can contribute to unequal access to preventive care \cite{fritzsche_ethical_2023, martin_clinical_2019, sirugo_missing_2019}.}}\\

As applications of ML in health become commonplace, it is crucial to recognize, account for, and mitigate such disparities in model performance to support health equity.  Broadly, an ML model is said to be {\it{fair}} if it does not discriminate against an individual or group \cite{barocas_fairness_2023}.  Concepts of fairness have been extensively studied across various disciplines, including social choice theory, game theory, economics, \blue{philosophy} and law \cite{young_equity_1995,roemer_equality_1998,roemer_theories_1998,rawls_theory_1971, rawls_justice_2001, hutchinson201950,  dwork_fairness_2012}. Building on these principles, the subfield of fairness within ML provides a framework for evaluating and mitigating bias throughout the model development process.   \\

Although fairness has been an extremely active area of research over the past decade, implementing fairness within ML for health is relatively nascent \cite{wang_bias_2022, feng_fair_2022,  liu_translational_2023}. A systematic review of articles utilizing ML for EHR-based phenotyping found that only 5\% of studies assessed fairness \cite{yang_machine_2023}. Similarly, a review of EHR-based prediction models found that most studies investigate the overall performance of ML models, but do not interrogate potential biases \cite{chen2024unmasking}. Beyond EHR applications, several scoping reviews of clinical ML models {\blue{developed with diverse data sources}} found that the adoption of fairness remains inconsistent, partly due to a knowledge gap between ML and clinical researchers \cite{canali_challenges_2022, yfantidou_beyond_2023, liu_translational_2023, plecko_causal_2024, anderson2024algorithmic}. Moreover, over two dozen definitions of fairness have been proposed, most of which originate outside health-focused literature, making it a particularly challenging domain to navigate.\\

To review this fundamental topic, we examine key notions of fairness and their use in ML \blue{for} health\footnote{This \blue{review} focuses on quantitative measures of fairness. Alternative approaches are discussed in the outlook section.}.  We first introduce biases that can emerge throughout the model development process using examples from the literature to illustrate why ML models can be unfair.  We then review what it means for a model to be fair, beginning with an overview of the most common fairness framework, group fairness, and moving to the emerging frameworks of individual and causal fairness \cite{awasthi2020beyond}.  Our discussion includes the mathematical formulation of various fairness criteria {\blue{as well as numerous}} real-world examples.  \blue{We conclude by highlighting the limitations of current approaches and outlining opportunities for future research.}  Our work highlights that, in light of the large number of fairness notions,  \blue{operationalizing} fairness in health-focused applications remains an open problem.  We therefore aim to provide a concise overview and organization of fairness definitions used within ML in health that can be used as a foundational resource for researchers navigating and contributing to the rich and nuanced fairness literature (see Supplementary Table S1 for a detailed comparison with existing references).

\section{Why ML models are unfair}
Even with a well-defined and well-intentioned research question, ML models can be unfair due to {\it{biases}} in the data, in the model, and/or in the deployment of the model \cite{rajkomar_ensuring_2018, mehrabi_survey_2021, fazelpour_algorithmic_2021, mhasawade2021machine, wang_bias_2022, xu_algorithmic_2022, drukker_toward_2023, ferrara_fairness_2024}.  Figure \ref{fig:bias-source} provides a visualization of sources of bias that can arise throughout model development and Table \ref{tab:bias-sources} details common biases \blue{using examples from the literature}.  

\begin{figure}[h!]
    \centering
    \includegraphics[scale = 0.35]{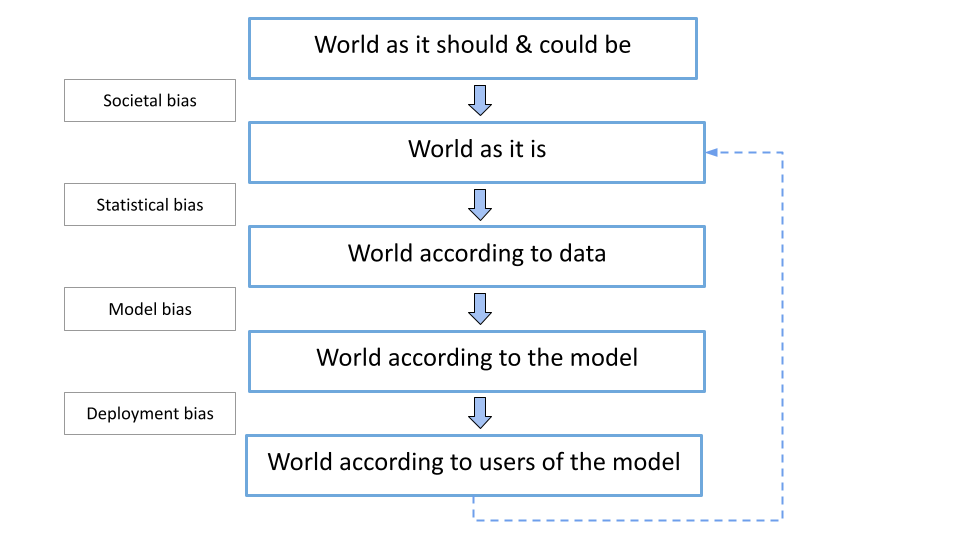} 
    \caption{\textbf{Sources of bias.} Bias can arise at each stage of the model development process, from data collection to deployment.  Each stage influences the next, with bias being potentially perpetuated and compounded throughout the development process. After deployment, an unfair model can also introduce or reinforce societal bias. This figure is adapted from \blue{existing literature}\cite{mitchell_algorithmic_2021,mbakwe2023fairness}. }
    \label{fig:bias-source}
\end{figure}

Briefly expanding on Table \ref{tab:bias-sources}, bias in the data arises from (i) societal and/or (ii) statistical bias \cite{mitchell_algorithmic_2021}.  With respect to the former, the data used in health applications most often measures and categorizes people and therefore encodes societal structures, injustices, and stereotypes, such as gender, racial, and age bias \cite{iliadis2016critical, suresh2021framework}.  For example,  underlying social inequalities in healthcare access can limit the amount of data available in EHRs of certain subgroups, such as Black patients \cite{getzen_mining_2023}.  In terms of statistical bias, the data can also fail to represent the population of interest due \blue{to} the sampling method (selection bias), time of collection (temporal bias), or data quality issues such as  missing, mismeasured, or insufficient data (missing data, measurement, and minority bias).  In the fairness literature, data with any of these undesirable properties are informally referred to as {\it{biased}} \cite{chouldechova_fair_2017, barocas_fairness_2023}.  Observational clinical data (e.g., claims data, medical images) inherently contains various biases and researchers must carefully consider sources of bias in any application \cite{genevieve_structural_2020, mccoy_ensuring_2020}.\footnote{\blue{While this section focuses on common sources of bias across health-focused applications, we acknowledge that different data present distinct fairness challenges. For more detailed discussion of fairness challenges in specific data, such as EHR\cite{holmesWhyElectronicHealth2021}, medical imaging\cite{drukker_toward_2023,xuAddressingFairnessIssues2024,yangLimitsFairMedical2024} and wearable data\cite{canali_challenges_2022}, we refer readers to relevant work in these domains.}}\\

Additionally, choices made during model training and evaluation can further amplify biases in the data or incorporate new bias, including selecting an inappropriate label (label bias), model (algorithmic bias), and/or evaluation metric or data set (evaluation bias).  Table \ref{tab:bias-sources} details several health examples subject to model bias.  Lastly, during deployment, bias in the data and/or model can be reinforced or introduced when users selectively disregard (dismissal bias) or overly trust (automation bias) a model’s outputs, or if models are inaccessible to certain subpopulations (privilege bias).  For instance, ML-assisted diagnostic imaging requires that hospitals have the necessary equipment, yet fewer than 20\% of rural emergency departments in Canada have in-house computed tomography (CT) scanners to utilize these advances \cite{davidson_landscape_2024}.  A review of biases outside of health-focused applications can be found in \blue{several recent works}\cite{suresh2021framework, mehrabi_survey_2021}.\\

Dr.\ Leo Celi, a health AI expert, emphasizes that ``data bias is the \blue{W}aterloo of health AI."  Bias must be a primary consideration throughout the ML pipeline - from problem formulation to model deployment \cite{tedx_talks_data_2023}.  Building on existing AI reporting guidelines\cite{wolff2019probast, liu2020reporting}, \blue{a bias evaluation checklist was introduced} to enable practitioners to systematically and holistically address bias in clinical predictive models \cite{wang_bias_2022}. Bias mitigation strategies and common pitfalls in applying ML in health are further discussed in \blue{the clinical literature}\cite{gichoya2023ai, panch_artificial_2019}.  The fairness criteria we introduce in the subsequent sections play an important role in this process. The criteria are typically integrated directly into model training or used as evaluative metrics to identify disparities after the model has been trained \cite{hardt_equality_2016, menon_cost_2018, pessach_review_2022, brahmbhatt2023towards, mbakwe2023fairness, goethals2024beyond}.  While this paper focuses on how to define fairness, strategies for mitigating bias are an equally important area of research and have been reviewed \blue{previously} \cite{hort2023bias}.

\begin{table}[htbp!]
\centering
\scriptsize
\begin{subtable}{\textwidth}
\centering
\caption{\scriptsize Bias in the data}
\begin{tabularx}{\textwidth}{|>{\hsize=0.4\hsize}X|>{\hsize=1\hsize}X|>{\hsize=1.6\hsize}X|} 

\hline
\textbf{Type of Bias} & \textbf{Definition} & \textbf{Example} \\ \hline
Societal (or historical) Bias & The data reflects long-standing societal disparities encoded within the data over time. & Clinical word embeddings trained on large corpora of text, such as clinical notes from healthcare systems, reflect biases about ethnic minorities \cite{zhang2020hurtful}. \\ \hline
Selection Bias & The data is not representative of the population of interest.  & \blue{Data collected from wearable devices does not reflect the general population, as usage rates are higher among younger individuals and those with higher socioeconomic status\cite{panSociodemographicsDigitalHealth2024}.}. \\ \hline
Measurement Bias & The data contains variables that are collected or measured inaccurately. & {\blue{When documented in EHRs, gender identity is often recorded without adequate provider training, which can lead to misrepresentation of an individual's identity \cite{kronk_transgender_2022}.}} \\ \hline
Temporal Bias & The data captures a specific time period that may not reflect current or future conditions. & {\blue{Administrative health data showed that pediatric mental health visits were lower than expected among individuals with lower socioeconomic status during the first year of the COVID-19 pandemic \cite{toulany_sociodemographic_2023}.}} \\ \hline
Minority Bias & The data lacks adequate representation from the minority group for the model to accurately learn {\blue{about them}}.  & Most of the data for genetic studies is from European ancestry populations  \cite{martin_clinical_2019}. \\ \hline
Missing Data Bias & The data has variables that are incompletely measured. & Patients from low-income backgrounds have higher rates of missing medical measurements in their medical records, which can be partially attributed to inequities in access to healthcare \cite{Chapman_Exploring_2022}. \\ \hline
\end{tabularx}
\end{subtable}

\begin{subtable}{\textwidth}
\centering
\caption{\scriptsize Bias in the model}
\begin{tabularx}{\textwidth}{|>{\hsize=0.4\hsize}X|>{\hsize=1\hsize}X|>{\hsize=1.6\hsize}X|} 

\hline
\textbf{Type of Bias} & \textbf{Definition} & \textbf{Example} \\ \hline
Label Bias & An imperfect proxy is selected to train a model instead of the outcome of interest. & Healthcare cost was used as a proxy for healthcare need in a commercial algorithm used to identify patients for high-risk care management programs.  The algorithm significantly under-identified Black patients for care services as less money is spent on Black patients relative to similarly healthy white patients \cite{obermeyer_dissecting_2019}. \\ \hline
Algorithmic Bias & Properties of a model and/or its training algorithm create or amplify bias in the data. & \blue{In clinical prediction tasks, differentially private models ensure privacy through the addition of calibrated noise.  The noise can reduce the model’s ability to learn from the tails of the data distribution, leading to accuracy loss that disproportionately affects minority groups} \cite{suriyakumar2021chasing}.   \\ \hline
Evaluation Bias & An inappropriate choice of benchmark data or metrics is used for evaluation. &  Existing public skin disease AI benchmarks do not have images of biopsy-proven malignancy, the gold standard for disease annotation, on dark skin \cite{daneshjou_lack_2021, daneshjou_disparities_2022}. \\ \hline
\end{tabularx}
\end{subtable}

\begin{subtable}{\textwidth}
\centering
\caption{\scriptsize Bias in deployment}
\begin{tabularx}{\textwidth}{|>{\hsize=0.4\hsize}X|>{\hsize=1\hsize}X|>{\hsize=1.6\hsize}X|} 
\hline
\textbf{Type of Bias} & \textbf{Definition} & \textbf{Example} \\ \hline
Automation Bias & Model users overly trust model outputs, sometimes even against their own knowledge. & A computer-aided diagnosis system that scans mammograms and marks suspicious areas of potential cancer features had lower sensitivity for women aged 40--49 compared to older age groups. A clinician using this system can disproportionately miss cancers for this age group \cite{lehman_diagnostic_2015}. \\ \hline
Dismissal Bias & Model users ignore model recommendations, often due to frequent false alerts. & If the same computer-aided diagnosis system also has a lower positive predictive value for women aged 40--49, then clinicians may disregard its recommendations, as they are more likely to be false positives \cite{ancker2017effects}.  \\ 
\hline
Privilege Bias & Models are not available or cannot be deployed in all settings, such as parts of the world where medical technology is unavailable.  & Less than 20\% of rural emergency departments in Canada have in-house access to computed tomography (CT) scanners required for ML-assisted diagnostic imaging \cite{davidson_landscape_2024}. \\
\hline
\end{tabularx}
\end{subtable}
\caption{\textbf{Common biases that arise in machine learning (ML) applications in health and examples from the literature.}}
\label{tab:bias-sources}
\end{table}

\section{What it means for an ML model to be fair}
Existing definitions of fairness primarily fall into three categories: \textit{group fairness}, \textit{individual fairness}, and \textit{causal fairness}. Group fairness criteria are commonly used in health and deem a model as fair if its predictions are similarly accurate or calibrated across a predefined set of groups. These groups are most often defined by a protected attribute(s)\footnote{Sensitive attribute and protected attribute are used interchangeably \cite{caton_fairness_2024}.  Fairness is also evaluated across groups defined by social determinants of health \blue{(e.g., income, education, job insecurity)} \cite{mccradden2020ethical}. } such as age or race. \blue{Other commonly used} protected attributes used in the health-focused literature \blue{include disability, marital status, national origin, sex, and socioeconomic status. We summarize these attributes in Table S2, along with representative examples from the literature.} In contrast, individual fairness is a less commonly used framework and requires that the model provide similar predictions to similar individuals based on user-defined similarity metrics \cite{dwork_fairness_2012, anderson2024algorithmic}. Lastly, causal fairness criteria utilize causal estimands to quantify unfairness and link observed disparities in model performance to their underlying cause \cite{plecko_causal_2024, makhlouf_when_2024}.  Causality-based fairness notions are particularly attractive for health-focused applications as they enable practitioners to interrogate biases \cite{zhang_causal_2024}. \\ 

Group fairness criteria are referred to as ``oblivious'' as they equate fairness with parities based solely on the distribution of the data (i.e., the predictions, outcome, and protected attribute), while individual and causal fairness criteria are ``nonoblivious'' as they require additional context in the form of user-defined similarity metrics and causal models, respectively.  Importantly, many fairness criteria are {\it{incompatible}} in that they cannot be simultaneously satisfied. A taxonomy of fairness, including different notions of group, individual, and causal fairness and their incompatibilities \blue{which} are introduced in subsequent sections, is presented in Figure \ref{fig:taxonomy}. \\

\begin{figure}[htbp!]
    \centering 
\includegraphics[width = \textwidth]{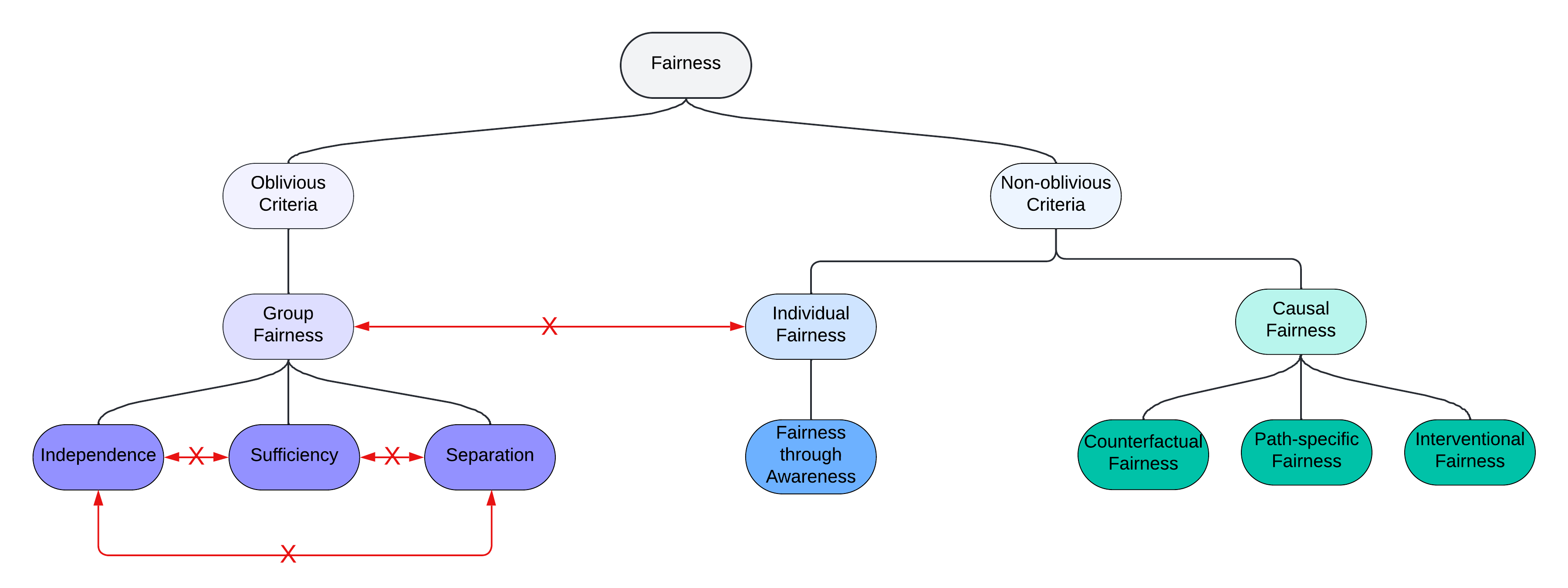}
    \caption{\textbf{Taxonomy of fairness.} Fairness criteria primarily fall into three categories: group, individual, and causal fairness. Group fairness criteria are oblivious in the sense that they can be entirely inferred from the distribution of the data. Individual and causal fairness criteria are non-oblivious criteria as they require specification of similarity metrics and a causal model, respectively. Many fairness criteria cannot be simultaneously satisfied and incompatibilities between different notions of fairness are depicted with a red X.  This taxonomy is not exhaustive and includes the notions of fairness and incompatibilities introduced in this \blue{review}.}
    \label{fig:taxonomy}
\end{figure}

\subsection{Notation}
Throughout, we denote the outcome of interest as $Y$, the features used for model training as $\bX$\footnote{Note that $\bX$ does not have to be a vector; it could be an image, text, tensor, or any other type of data.}, and the variable for the protected attribute as $A$.  $\bX$ may or may not contain $A$ and we discuss this issue when we introduce individual fairness.  We let $S = \hat{f}(\bX)$ be the output from the ML model, where $\hat{f}$ is learned from a set of training data.  For example, in classification, $Y$ is a binary label for membership in the positive or negative class and $S$ is the predicted probability of being in the positive class, referred to as the score.  \blue{In this work, we primarily focus on binary classification settings, where the final decision $D$ is classified as positive if the score $S$ exceeds a predetermined threshold and negative otherwise.  Binary classification tasks are the most widely used and well-studied in the fairness literature \cite{mitchell_algorithmic_2021, mehrabi_survey_2021, castelnovo_clarification_2022}.}

\subsection{Group fairness}
Group fairness criteria require ML models to perform similarly across groups of interest {\blue{defined by $A$}} and are the most popular fairness framework in health-focused applications \cite{awasthi2020beyond}.  The criteria primarily fall into three categories: \textit{independence}, \textit{separation}, and \textit{sufficiency} \cite{berk_fairness_2018, barocas_fairness_2023}.  \blue{In the subsequent sections, we provide textual descriptions of the definitions of common metrics as well as examples of their usage within health-focused applications. Table \ref{tab:MetricsSummary} summarizes mathematical definitions together with an interpretation in the context of a real-world example.  We provide a brief discussion of approaches} for continuous outcomes, continuous protected attributes, categorical protected attributes that define more than two groups, and multiple protected attributes (i.e., subgroup fairness) in Supplementary Section 2.

\subsubsection{Independence}
{\color{black}
\textit{Definitions}\\
Under independence, an ML model is said to be fair if its decisions do not depend on the protected attribute (i.e., $D \indep A$).  \textit{Statistical (or demographic) parity} is a common measure of independence that requires that the model classify individuals into the positive class at the same rate in each group \cite{dwork_fairness_2012, corbett-davies_algorithmic_2017}.  Conditional statistical parity relaxes this concept by requiring the rate of positive classifications to be the same within more granular groups defined by the protected attribute and other relevant factors.  \\

\textit{Usage within health-focused applications}\\
Independence-based fairness metrics, such as statistical parity, are infrequently used in health-focused applications as the prevalence of clinical outcomes often differs across groups defined by protected attributes (e.g., multiple sclerosis is more common in females than males). Enforcing independence may also prevent a model from learning a genuine association between the protected attribute and the outcome, potentially leading to an overall reduction in performance \cite{pfohl_creating_2019, lei2024inductive}.  However, independence-based metrics may still be informative when the goal is to assess whether a model disproportionately assigns high-risk predictions to specific groups. For example, statistical parity was used to evaluate a model predicting heart failure length of stay and in-hospital mortality based on clinical and social determinants of health data \cite{li2022improving}. This metric assessed whether patients from different ethnoracial groups were equally likely to be classified as high risk for the outcomes,  revealing disparities in prediction rates, particularly in models that excluded social determinants of health, and demonstrating that incorporating such data can improve fairness without compromising on overall accuracy.\\

\subsubsection{Separation}
\textit{Definitions}\\
Separation requires that the model's decisions do not depend on the protected attribute within the positive and negative classes (i.e., $D \indep A \mid Y)$. This implies that, among individuals in the positive (or negative) class, the rate of making a positive (or negative) decision is consistent across groups.  Common separation-based metrics therefore aim to equalize error rates across the groups, including the false negative rate (FNR, known as {\it{equal opportunity}}), false positive rate (FPR, known as {\it{predictive equality}}), or both (known as {\it{equalized odds}}).  Additional separation-based metrics are detailed in Table \ref{tab:MetricsSummary}, including \textit{balance for the positive class} and \textit{balance for the negative class} \cite{kleinberg2017inherent}. \\

\textit{Usage within health-focused applications}\\
Separation-based metrics have been widely used in health-focused applications, although the specific choice of metric depends on the context. When false negatives have the most severe consequences, equal opportunity may be preferred.  For instance, this metric was used in a study of state-of-the-art computer vision models used to detect common diseases from chest X-rays, where a false negative corresponded to incorrectly identifying a patient as not having ``no finding'' on their X-ray \cite{seyyed-kalantari_underdiagnosis_2021}. The models had higher FNRs in several under-served subpopulations occupying intersections of oppression, such as Hispanic female patients, potentially resulting in delayed access to care.  There are also situations where balancing both the FNR and FPR is more appropriate and equalized odds should be prioritized \cite{hardt_equality_2016}. \blue{For instance,  Yang et al\cite{yang2023algorithmic} employed equalized odds to assess the fairness of ML-based rapid COVID-19 screening tools used in emergency departments as disparities in FNRs can lead to inadequate monitoring in certain groups, while imbalanced FPRs may result in disproportionate unnecessary testing.} 

\subsubsection{Sufficiency}
\textit{Definitions}\\
Sufficiency aims to equalize error rates among individuals with similar decisions \cite{castelnovo_clarification_2022}. Formally, sufficiency requires that the label does not depend on the protected attribute given the model's decision ( i.e., $Y \indep A \mid D)$.  The decision is therefore ``sufficient'' for predicting the outcome in the sense that it subsumes the protected attributes \cite{barocas_fairness_2023}.  Common sufficiency-based metrics focus on equalizing the positive predictive value (PPV, known as \textit{predictive parity}), both the PPV and negative predictive value (NPV, known as \textit{conditional use accuracy equality}), and calibration (known as \textit{well-calibration}\footnote{In statistical literature, this concept is also referred to as \textit{strong calibration}\cite{van_calster_calibration_2019}.})\cite{chouldechova_fair_2017, berk_fairness_2018, kleinberg2017inherent, guo_calibration_2017}. \\

\textit{Usage within health-focused applications}\\
Sufficiency-based metrics have been used to evaluate models for disease screening, management, and triage. For example, Raza et al.\ evaluate a model for predicting 30-day hospital readmission using predictive parity \cite{raza2024exploring}. In this setting, disparities in the PPV across age groups can cause the model to disproportionately flag elderly patients as being at high risk of readmission, even when their baseline health status is similar to that of younger patients. This type of bias may lead to unnecessary interventions and undue stress for elderly patients. \\

Calibration-based metrics are frequently applied in health-focused examples. Well-calibration was evaluated in our introductory example of evaluating the fairness of a commercial risk prediction algorithm used to enroll patients in a high-risk care management program.  In this example, healthcare costs were used as a proxy for health needs in training the model. As a result, at equivalent model scores, Black patients were in significantly poorer health than white patients as less money had historically been spent on their healthcare, meaning that they had to be sicker to qualify for the program \cite{obermeyer_dissecting_2019}.}

\begin{table}[htbp!]
\begin{subtable}{\linewidth}
\centering
\tiny 
\renewcommand{\arraystretch}{1.5} 
\begin{tabular}{|p{0.08\textwidth}|p{0.45\textwidth}|p{0.4\textwidth}|>{\centering\arraybackslash}p{0.04\textwidth}|}
\hline 
\multirow{2}{0.05\textwidth}{\textbf{Metric}} & \multirow{2}{0.5\textwidth}{\textbf{Definition}} & \multirow{2}{0.4\textwidth}{\textbf{Interpretation}} & \multirow{2}{0.04\textwidth}{\centering \textbf{Relax-}} \\ 
& & & \textbf{ation}\\ \hline
 \multicolumn{4}{|l|}{\textit{Independence-based criteria}  } \\ \hline
 Statistical Parity & $P(D = 1 \mid A = a_0) = P(D = 1 \mid A = a_1) \quad \forall  a_0, a_1 \in \mathcal{A}$ & In predicting hospitalization and emergency department (ED) visits in heart failure, the probability of a predicted hospitalization or an ED visit is the same for males and females \cite{davoudi_fairness_2024}. &  \\
\hline 
 Conditional Statistical Parity & $ P(D = 1 \mid \bZ,  A=a_0) = P(D = 1 \mid \bZ, A = a_1) \quad \forall a_0, a_1 \in \mathcal{A}$  & In predicting hospitalization and ED visits in heart failure, the probability of a predicted hospitalization or an ED visit is the same for males and females after adjusting for age and pre-existing conditions \cite{davoudi_fairness_2024}. & \checkmark \\
 \hline 
 \multicolumn{4}{|l|}{\textit{Separation-based criteria} } \\ \hline
 Equalized Odds & $P(D = 1 \mid Y =0, A=a_0) = P(D = 1 \mid Y =0, A=a_1) \quad \text{and}$ \newline $P(D = 0 \mid Y =1, A=a_0) = P(D = 0 \mid Y =1, A=a_1)  \quad \forall  a_0, a_1 \in \mathcal{A}$ & In predicting asthma exacerbation in children, the rates of false positives (children without exacerbation incorrectly identified as having exacerbation) and false negatives (children with exacerbation incorrectly identified as not having exacerbation) are the same for children from low and high socio-economic classes \cite{juhn2022assessing}. 
 & \\
 \hline
 Predictive Equality & $P(D = 1 \mid Y = 0, A = a_0) = P(D = 1 \mid Y = 0, A= a_1) \quad \forall  a_0, a_1 \in \mathcal{A}$ & \blue{In a model predicting cardiovascular disease (CVD), the rate of false positives (people without CVD incorrectly identified as having CVD) is the same for males and females \cite{Vargae002395}.} & \checkmark \\
 \hline 
 Equal Opportunity &  $P(D = 0 \mid Y =1, A=a_0) = P(D = 0 \mid Y =1, A=a_1) \quad \forall  a_0, a_1 \in \mathcal{A}$ &  \blue{In a model predicting suspicious findings from chest X-rays, the rate of false negatives (patients with true findings incorrectly classified as normal) is the same for males and females \cite{ueda_artificial_2021}.} & \checkmark \\
\hline 
 Balance for Positive Class & $\mathbb{E}(S \mid Y=1, A=a_0) = \mathbb{E}(S \mid Y=1, A=a_1)  \quad \forall  a_0, a_1 \in \mathcal{A}$ &In a model used to screen for lung cancer from chest X-rays, the average score of people with lung cancer is equal for males and females \cite{ueda_artificial_2021}. & \checkmark\\
\hline 
 Balance for Negative Class &  $\mathbb{E}(S \mid Y=0, A=a_0) = \mathbb{E}(S \mid Y=0, A= a_1) \quad \forall  a_0, a_1 \in \mathcal{A}$ & In a model used to screen for lung cancer from chest X-rays, the average score among males and females without lung cancer are equal \cite{ueda_artificial_2021}.& \checkmark\\
 \hline
  \multicolumn{4}{|l|}{\textit{Sufficiency-based criteria}  }  \\ \hline
Conditional Use Accuracy Equality & $P(Y = 1 \mid D =1, A=a_0) = P(Y = 1 \mid D =1, A=a_1)$ and \newline $P(Y=0 \mid D=0, A=a_0) = P(Y=0\mid D=0, A = a_1) \quad \forall  a_0, a_1 \in \mathcal{A}$  &  In predicting hospital readmission, the probability of being readmitted given the model makes that decision and the probability of not being readmitted given the model makes that decision is the same for Black and white patients \cite{raza2023auditing}. & \\ 
\hline
 Predictive Parity & $ P( Y=1 \mid  D=1, A=a_0) = P(Y=1 \mid D=1, A = a_1) \quad \forall  a_0, a_1 \in \mathcal{A}$ & In predicting hospital readmission, the probability of being readmitted given the model makes that decision is the same for Black and white patients \cite{raza2023auditing}. & \checkmark \\ \hline
Well Calibration & $P(Y=1\mid S=s, A = a) = s \quad \forall  a \in \mathcal{A}, s \in [0,1]$ & In a model used to predict in-hospital mortality, the predicted event rates match the observed event rates at all values of the score for males and females \cite{roosli_2022_peeking}. & \\
\hline
Test Fairness & $P(Y = 1 \mid S=s, A = a_0) = P(Y = 1 \mid S=s, A= a_1) \quad \forall  a_0, a_1 \in \mathcal{A}, s \in [0,1]$ & In a model used to predict in-hospital mortality, the predicted event rates are the same for males and females at all values of the score \cite{roosli_2022_peeking}. & \checkmark\\
\hline
\multicolumn{4}{|l|}{\textit{Other criteria}} \\ \hline 
Brier Score Parity & $\mathbb{E}[(Y-S)^2 \mid A =a_0] = \mathbb{E}[(Y-S)^2 \mid A =a_1],  \quad \forall  a_0, a_1 \in \mathcal{A}$ & In a model used to enroll patients into high-risk care management programs, the mean squared error between the score and the label is the same for Black and White patients \cite{obermeyer_dissecting_2019}. & \\
\hline
Overall Accuracy Equality & $P(D=Y\mid A=a_0) = P(D=Y \mid A=a_1)  \quad \forall a_0, a_1 \in \Asc$  & In a model used to  \blue{classify sex based on gait data collected from wearable sensors, the probability of correct classification (i.e., correctly identifying an individual’s true sex) is the same for younger and older individuals\cite{ahadWearableSensorBasedGait2020}.} & \\
\hline 
Treatment Equality & $\frac{\#\{D = 0, Y = 1, A = a_0\}}{\#\{D = 1, Y = 0, A = a_0\}} = \frac{\#\{D = 0, Y = 1, A = a_1\}}{\#\{D = 1, Y = 0, A = a_1\}} \quad \forall a_0, a_1 \in \Asc$  & In a model used to screen for lung cancer from chest X-rays, the ratio of the false negatives \blue{(patients with lung cancer incorrectly classified as not having lung cancer) and  false positives (patients without lung cancer incorrectly classified as having lung cancer)} is the same for both males and females \cite{ueda_artificial_2021}. &\\
\hline
\end{tabular}
\end{subtable}
\caption{\footnotesize\textbf{Common group fairness criteria.} Mathematical definitions of group fairness criteria, their interpretation in the context of an example from the literature, and an indication of whether the criterion is a relaxation of independence ($D \indep A$), separation ($D \indep A \mid Y$), or sufficiency ($Y \indep A \mid D$). Symbols: $\# =$ number of, $P =$ probability, $E =$ expected value. Notations: $Y$: outcome, $\bX$: features used for model training, $A$: protected attribute that takes value in the set $\mathcal{A}$, $S$: model score, $D$: model classification based on thresholding $S$, $\bZ$: additional set of features.}
\label{tab:MetricsSummary}
\end{table}

\subsubsection{Incompatibilities}
Independence, sufficiency, and separation provide different perspectives on what it means for a model to be fair.  Except under highly restrictive conditions, it is not possible for an algorithm to fulfill all criteria simultaneously \cite{hardt_equality_2016,chouldechova_fair_2017, corbett-davies_measure_2023}.  It is therefore critical for researchers to choose which group fairness considerations are most relevant to their context.  More specifically, the following pairs of criteria are {\it{incompatible}} in the sense that they cannot generally be simultaneously satisfied: independence and sufficiency, independence and separation, and separation and sufficiency. A basic requirement for any of these pairs to hold is that the outcome and the protected attribute are marginally independent (i.e., $Y \indep A$).  In classification problems, this means the probability of being in the positive class is the same across groups. \blue{This condition is violated in many clinical contexts, such as when a disease or outcome is more common among certain subpopulations.  Supplementary Section 3 provides additional mathematical details related to the incompatibilities and also introduces \textit{approximate} (or $\epsilon$)\textit{ fairness}, which allows for small deviations in group fairness metrics in order to address these incompatibilities\cite{denis_fairness_2023}.} \blue{Under approximate fairness, it becomes possible to satisfy multiple fairness criteria across the three categories\cite{bell2023possibility}. }

\subsection{Individual fairness}
\subsubsection{\blue{Definitions}}
In contrast to group fairness that targets the average performance of a model across groups, individual fairness ensures that ``like cases are treated alike'', an idea grounded in Aristotle's consistency principles \cite{dwork_fairness_2012, fleisher_whats_2021}. Here we introduce a foundational concept in individual fairness known as {\it{fairness through awareness}} (FTA). As one of the earliest formalizations of individual fairness, FTA introduced the idea of using similarity metrics to quantify the distance between individuals and ensure that similar individuals receive similar predictions.  More formally, fairness is achieved if \blue{for any two individuals with features $\bX_i$ and $\bX_j$,
$$d_S(S_{i}, S_{j}) <d_{\bX}(\bX_{i}, \bX_{j})$$
where  $d_S$ and $d_{\bX}$ denote distance metrics defined on the predicted probability and feature spaces, respectively\cite{dwork_fairness_2012}.}\footnote{\blue{Suppose $A$ is the protected attribute of interest. A simple choice of $d_{\bX}$ is to define the distance between two individuals as 0 if all features other than $A$ are identical and 1 otherwise. With $D_S$ defined to take the value 0 if the model produces the same classification and 1 otherwise, FTA is closely related to a property known as {\it{causal discrimination}}. Causal discrimination is a causality-based fairness metric that examines whether changing a protected attribute causes a change in the model's output, holding all else equal \cite{galhotra_fairness_2017}.}} \blue{A counter approach to FTA, though not an individual fairness criterion, is fairness through unawareness (FTU). FTU is intended to be a catch-all solution to prevent bias by not explicitly including protected attributes into modeling.}

\subsubsection{\blue{Usage within health-focused applications}}
The implementation of FTA-based approaches critically depends on the metrics used to define similarity between individuals and their predictions. For predictions, \blue{the most commonly used metric is the absolute difference between predicted probabilities \cite{dwork_fairness_2012, zemel_learning_2013, keya2021equitable, rahman2022fair , anderson2024algorithmic}. }However, the metric used to measure the similarity between individuals has been a focus of ongoing research as it determines how individuals are comparable, which relies on an ``awareness'' of the context. \blue{For example, Zemel et al. proposed to assess similarity between individuals by identifying the $k$ closest individuals in the feature space based on a chosen distance metric, such as the Euclidean distance \cite{zemel_learning_2013}.} In the original FTA paper, the authors suggest leaving the choice of metric to domain experts. As there is currently no agreed-upon standard, particularly within health settings, various strategies have been proposed to learn the metric from available data \cite{kim2018fairness, ilvento2019metric, mukherjee2020two}.\\ 

\blue{A scoping review\cite{anderson2024algorithmic} on individual fairness in healthcare provides a comprehensive list of individual fairness methods and corresponding software. The review highlights that individual fairness is just emerging within health-focused applications, likely due to the relative nascency of individual fairness methods developed within the ML literature and the \blue{difficulty} in appropriately quantifying similarity across individuals. Nonetheless, a small number of studies have developed models aimed at achieving individual fairness in the context of survival analysis.  Motivated by an AI system used to perform needs-based prioritization of the Medicaid waitlists, Keya et al\cite{keya2021equitable} introduced an individual fairness constraint for the Cox proportional hazards model, using Euclidean distance in the feature space to define similarity across individuals and penalizing the absolute difference in predicted hazard scores. Subsequently, Rahman and Purushotham \cite{rahman2022fair} extended this approach to general survival models (e.g., non-hazard-based) by reformulating the fairness constraint based on the absolute difference in predicted survival probabilities.  The proposed method significantly reduced disparities in predicted survival probabilities across individuals while maintaining competitive overall predictive accuracy with standard survival models in analyses of three diverse real-world healthcare datasets.} \\

\blue{It is important to note that individual fairness is motivated by the inherent weakness in group fairness criteria that only consider average model performance within groups.  That is, there are situations in which group fairness can be satisfied, but individuals within a group can be discriminated against\cite{dwork_fairness_2012}. The incompatibilities among individual and group fairness have been examined from a theoretical perspective in \blue{several recent studies} \cite{friedler_impossibility_2016, binns_apparent_2020,  castelnovo_clarification_2022, xu_compatibility_2024}. However, under certain conditions, such as when the Wasserstein distance between the distributions of protected attributes is small, individual fairness can imply statistical parity \cite{dwork_fairness_2012, zhou_group_2022}.} \\

\blue{We close our discussion of individual fairness by briefly commenting on FTU as it is a counter-approach to FTA.  In particular, FTU warrants careful consideration in health-focused applications as it removes protected attributes from a model.}  First, protected attributes may serve as critical predictors for the outcome of interest, such as age in sepsis or cardiovascular disease prediction and race in cancer screening models \cite{damen2016prediction, kim_sex_2018, chicco2020survival, cirillo2020sex}. Excluding these predictors can diminish overall predictive accuracy and impact all individuals adversely or even lead to bias against the majority group.  Second, protected attributes are often highly correlated with non-protected attributes. Simply removing protected attributes from the model does not prevent the model from inferring them from other attributes. This is evident in a recent study\cite{gichoya_ai_2022} that found ML models can infer race from medical images. Finally, the quality predictors included in the model may vary across the protected groups (e.g., rates of missing data, measurement error). \blue{For instance, family history is an important predictor for cancer risk prediction, but has been shown to be less reliably documented for Black participants in self-reported family history data and therefore less useful for models developed with these data \cite{kupfer_racial_2006, chavez-yenter_association_2022}. To address racial disparities in data quality, Zink et al\cite{zink_race_2024} showed that including race as a predictor significantly improved model performance compared to race-blind algorithms for colorectal cancer.  FTU is therefore not always ethical, achievable, or desirable \cite{banja2023fairness}.} \\

\subsection{Causal fairness}
We next provide a brief overview of causal fairness, an increasingly popular topic in ML that warrants a dedicated review within health applications \cite{loftus2018causal, fawkes2022selection, makhlouf2022survey, carey2022causal, nilforoshan_causal_2022, anderson2024algorithmic, zhang_causal_2024}. As the name suggests, causal fairness focuses on understanding the causal relationships between protected attributes and a model’s decisions \cite{plecko_causal_2024}.  \blue{Although causality-based approaches are especially valuable in health applications for mitigating confounding effects and disentangling mechanisms of bias, the use of causal fairness in health is in its early stages.  As such, we use toy examples to introduce most of the concepts and reference existing real-world examples whenever possible.}\\

\blue{We focus our discussion on several common fairness notions based on counterfactuals, many of which build upon the previously introduced notions of group and individual fairness.  Additional definitions that we do not cover in detail and that require background knowledge in causal inference, such as path-specific fairness, direct and indirect discrimination, principal fairness, and interventional fairness, are detailed extensively in the ML literature \cite{chiappa_path-specific_2019,nabi_fair_2018,wu_pc-fairness_2019,zhang_causal_2017, zhang_mitigating_2018, ogburn_comment_2019,  wang_blessings_2019, salimi2019interventional, zuo_interventional_2024, zhang_causal_2024}.  In the context of defining what it means to be fair, the counterfactual, or more simply the ``what-if'' statement, is most often the unobserved model's decision that would have happened if the protected attribute had been different \cite{kusner_counterfactual_2017, fawkes2022selection, bynum2024new}.  One of the earliest fairness criteria based on counterfactual reasoning is (individual) {\it{counterfactual fairness}}, which considers a model fair if, for a given individual, the predicted outcome remains the same when their protected attribute is counterfactually altered while holding all other variables constant \cite{kusner_counterfactual_2017}.  Counterfactual fairness may also be viewed as an individual fairness criterion as it is grounded in achieving similar treatment for similar individuals\cite{loftus2018causal}.}     \\

\blue{To further explain the idea behind counterfactual fairness, we use a simplified example adapted from the original work\cite{kusner_counterfactual_2017} on the topic and provide the technical definition in Table \ref{tab:causal}.  Figure \ref{fig:dag} presents the directed acyclic graph (DAG) for this example.  For illustration, suppose that $Y$ is an indicator of 30-day hospital readmission, $X$ a variable for emergency healthcare service utilization in the past 6 months that is used to predict $Y$, $A$ an indicator of membership in a socioeconomic group, and $U$ a latent variable measuring frailty that is not observed.  In this example, frailty causes people to be more likely to be readmitted and to visit the emergency department.  Further, individuals of a particular socioeconomic group are more likely to use the emergency department, not because they are more frail (i.e., $A$ does not impact $U$), but potentially due to structural barriers such as lack of primary care access or insurance coverage. These individuals, however, are no more likely to be readmitted than anyone else with the same level of frailty.  In this scenario, if a model uses $X$ to predict $Y$, it may assign higher predicted probabilities of readmission to individuals from that socioeconomic group, even though their true risk of readmission given $U$ is no higher. In terms of counterfactual fairness, this prediction is considered unfair; changing $A$ while holding $U$ constant would change $X$ and consequently the prediction. In other words, the model’s prediction is sensitive to $A$ in a way that is not justified by underlying differences in health.} \\

\begin{figure}[htbp!]
    \centering
\begin{tikzpicture}[node distance=2cm]

\tikzstyle{block} = [circle, draw, fill=gray!30, text centered, inner sep=0.5cm, node distance = 4cm]
\tikzstyle{whiteblock} = [circle, draw, fill=white, text centered, inner sep=0.5cm, node distance = 4cm]
\tikzstyle{arrow} = [thick,->,>=stealth]

\node (A) [block] {A};
\node (X) [block, below of=A] {X};
\node (Y) [block, right of=X] {Y};
\node (U) [whiteblock, above of=Y] {U};

\draw [arrow] (A) -- (X);
\draw [arrow] (U) -- (X);
\draw [arrow] (U) -- (Y);

\end{tikzpicture}
\caption{\textbf{Directed acyclic graph (DAG) for the counterfactual fairness example}. The DAG illustrates the relationships among the outcome, $Y$, the feature $X$, the protected attribute $A$, and unobserved factors $U$.  Gray nodes denote variables that are observed and white nodes denote variables that are unobserved. }
\label{fig:dag}
\end{figure}

\blue{While an oversimplified view of the many factors that impact hospital readmission, this example illustrates both the value of causal thinking in understanding the contributors to unfairness as well as the potential difficulties.  Namely, causal fairness criteria typically rely on a well-defined causal model, which can be difficult to specify in more complex situations, and whose assumptions cannot generally be verified using observational data.  \cite{kusner_counterfactual_2017, salimi2019interventional, fawkes2022selection}. Moreover, issues of identifiability can also arise (i.e., situations where causality-based fairness notions cannot be measured uniquely from the data) and we refer readers to the work of Makhlouf et al.\ for further discussion of this topic \cite{makhlouf2022survey}.  That said, numerous other metrics based on counterfactuals have been proposed to quantify fairness.  For example, \textit{counterfactual parity} measures a population-level, rather than an individual-level, causal effect of the protected attribute on the model's decision \cite{coston_counter_20}. In our example, counterfactual parity holds if the overall proportion of predicted readmissions stays the same if everyone's socioeconomic status were counterfactually altered. That is, the model's prediction does not causally depend on socioeconomic status at the population level.} \\ 

\blue{Counterfactual extensions of a number of group fairness metrics have also been proposed. For example, Pfohl et al.\ introduced an individual-level extension of equalized odds in order to evaluate prediction models for prolonged inpatient length of stay and mortality across groups determined by gender, race, and age  \cite{pfohl_counterfactual_2019}. The criterion, \textit{individual equalized counterfactual odds}, is satisfied if the observed and counterfactual predictions align for an individual when their protected attribute is changed (i) given all other variables are held constant, and (ii) conditioned on the observed outcome matching the counterfactual outcome.  The condition in (ii) distinguishes individual equalized counterfactual odds from counterfactual fairness. The purpose of adding this condition is to ensure that the fairness comparison is made only in cases where the individual’s outcome would remain the same, even if their protected attribute were different. This avoids penalizing the model for differences in prediction that may be justified by genuine differences in observed and counterfactual outcomes.  In our running example, individualized counterfactual equalized odds means that for any patient, the model’s prediction of 30-day readmission should stay the same if we counterfactually assign them to a different socioeconomic group, provided their readmission status itself would also remain unchanged and holding other variables (i.e., frailty) constant.  This notion of fairness helps identify whether the model’s use of socioeconomic group reflects unjustified bias, as opposed to reflecting legitimate differences in outcomes due to other factors.  Additional extensions of group fairness criteria, including \textit{counterfactual equalized odds} and \textit{counterfactual predictive parity}, are formally defined in Table \ref{tab:causal}.}

\begin{table}[h!]
\centering
\blue{
\begin{tabular}{|p{3cm}|l|}
\hline
\textbf{Metric} & \textbf{Definition} \\
\hline
(Individual) Counterfactual Fairness\cite{kusner_counterfactual_2017} & 
\parbox{11.3cm}{$P(D_{A \leftarrow a}(U) = d \mid \bX = \bx, A = a) ={}$\\
$P(D_{A \leftarrow a'}(U) = d \mid \bX = \bx, A = a) \quad \forall a, a' \in \mathcal{A}$} \\
\hline
Counterfactual Parity\cite{coston_counter_20} & 
$P(D  = 1\mid A = a) = P(D_{A \leftarrow a'} = 1 \mid A = a) \quad  \forall a, a' \in \mathcal{A}$ \\
\hline
Individualized Equalized Counterfactual Odds\cite{pfohl_counterfactual_2019} & 
\parbox{11.3cm}{$P(D_{A \leftarrow a}(U) = d \mid \bX = \bx, A = a, Y_{A \leftarrow a} = y) ={}$\\
$P(D_{A \leftarrow a'}(U) = d \mid \bX = \bx, A = a, Y_{A \leftarrow a'} = y) \quad \forall a, a' \in \mathcal{A}$} \\
\hline
Counterfactual Equalized Odds\cite{coston_counter_20} & \parbox{11.3cm}{\blue{$P(D = d\mid Y^0 = y, A) = P(D = d\mid Y^0 = y)$ for $d \in \{0, 1\}$, and $y \in \{0,1\}$} } \\
\hline
Counterfactual Predictive Parity\cite{coston_counter_20} & \parbox{11.3cm}{$P(D = d\mid Y^0 = y, A) = P(D = d\mid Y^0 = y)$ for $d \in \{0, 1\}$ and $y \in \{0,1\}$} \\
\hline
\end{tabular}
}
\caption{\blue{\textbf{Common Notions of Causal Fairness}. Symbols: $P =$ probability. Notations: $Y$: outcome, $D$: model classification, $\bX$: features used for model training,  $A$: protected attribute that takes value in a set $\mathcal{A}$, $D_{A \leftarrow a}$: classification when $A = a$, $D_{A \leftarrow a'}$: classification when the protected attribute is counterfactually altered to $A = a'$, $D_{A \leftarrow a}(U)$: classification when $A = a$ holding $U$ constant, $D_{A \leftarrow a'}(U)$: classification when the protected attribute is counterfactually altered to $A = a'$ holding $U$ constant,  $Y_{A \leftarrow a}$: outcome when $A = a$,  $Y_{A \leftarrow a'}$: outcome when the protected attribute is counterfactually altered to $A = a'$, $Y^0$: potential outcome with $D = 0$.}}
\label{tab:causal}
\end{table}

\section{Outlook}
Our work explores why models are unfair and various ways fairness has been defined in ML for health.  \blue{Despite substantial progress in recognizing the importance of bias and in applying group, individual, and causal-based criteria in health applications, there is a lack of consensus on how to appropriately quantify fairness \cite{zong_medfair_2023, gichoya_equity_2021}.} \blue{Each framework comes with unique benefits and challenges.}  Group fairness criteria are relatively easy to implement and interpret, but limited by potential individual fairness violations and by incompatibilities that exist among various criteria.  Likewise, individual fairness criteria prioritize similar outcomes for similar individuals to enhance equity in decision-making, but can mask disparities at the group level and leave certain populations underserved. \blue{Additionally, individual fairness requires researchers to define appropriate similarity metrics tailored to their specific context as no agreed-upon standards currently exist.} Causal fairness, which is gaining traction in health, enables researchers to investigate causes of unfairness to inform fair decision-making.  However, specifying reliable causal models can be challenging in health contexts. \blue{Compared to group and individual fairness, causal fairness remains underexplored in ML for health\cite{pfohl_counterfactual_2019, zhang_causal_2024}. This gap presents important opportunities for future work, specifically in elucidating the feasibility and utility of causal-based criteria in diverse health-focused applications.}\\ 

\blue{In practice, balancing different fairness frameworks requires weighing the benefits of group-level equity, individual-level considerations, and causal understanding \cite{schrouff_algorithmic_2022}. For instance, in large-scale lung cancer screening, equal opportunity may be prioritized as missing a diagnosis can lead to delayed treatment and poor clinical outcomes for certain subgroups\cite{rajkomar_ensuring_2018}. Conversely, in predicting organ compatibility after transplant, individual fairness may take precedence to ensure that patients with similar clinical profiles and predicted outcomes are treated similarly irrespective of protected attributes given the scarcity of donor organs\cite{Farnadi_St-Arnaud_Babaki_Carvalho_2021}. Causal fairness may be relevant in either setting if the goal is to identify and interrogate sources of bias\cite{plecko_causal_2024}.} \blue{We therefore recommend that researchers stay informed about the full range of available fairness methods and collaborate within interdisciplinary teams (e.g., data scientists, clinicians, ethicists) to make context-appropriate decisions.  Significant effort has been made to provide practical recommendations for incorporating fairness into ML, including the development of a bias evaluation checklist, a framework for integrating health equity into model development, and recommendations for identifying ethical concerns \cite{wang_bias_2022, kim_development_2024, cagliero_framework_2023}.  Importantly, these resources emphasize that a quantitative understanding of bias is just one component of operationalizing fairness in ML for health.  Establishing common principles and standards that prioritize fairness across the entire model development pipeline is essential, as post hoc fairness assessments alone are insufficient  \cite{mccradden_whats_2023}.}  \\

\blue{While we focus on how to define fairness, strategies for mitigating bias within ML applications are an equally active and important area of research.  We refer interested readers to a recent systematic review that presents various bias mitigation strategies and their use in practice \cite{hort2023bias}.  Importantly, a necessary step in any bias mitigation strategy is to select the fairness definition that the mitigation strategy will attempt to enforce.  Group fairness metrics are the most common targets, though alternative strategies have been proposed \cite{goethals2024beyond}.  Broadly, bias mitigation can be performed by de-biasing the training data (pre-processing), during model training (in-processing), and after model training (post-processing).  Pre-processing techniques include resampling\cite{romano_achieving_20} or reweighting samples\cite{calders_building_09} to correct imbalances among groups.  One approach to in-processing is to include additional regularization terms to penalize a model's deviation from a pre-specified fairness metric\cite{kamiran_discrimination_10}. In post-processing, for example, a technique to achieve equalized odds is to set distinct decision thresholds for different groups \cite{hardt_equality_2016}. However, a practical consideration in any bias mitigation strategy is the well-known trade-off between achieving fairness and high overall accuracy and calibration \cite{menon_cost_2018, brahmbhatt2023towards, mbakwe2023fairness}. Strategies for bias mitigation throughout the ML pipeline are further detailed in Gichoya et al.\cite{gichoya2023ai}. \\}

\blue{Lastly, health data present inherent and unique challenges, making it difficult or sometimes impossible to measure fairness accurately \cite{wick2019unlocking, ji2020can, choi2020fair, zhang2021fairness}.  Most existing fairness definitions were developed outside the healthcare context and may not adequately capture the forms of bias that arise in health-related settings \cite{mccoy_ensuring_2020}. For instance, most fairness approaches focus on a single protected attribute, which limits their applicability in complex, real-world scenarios that involve imperfect models, multiple intersecting attributes, and numerous practical considerations across the ML pipeline, such as data limitations, resource constraints, and policy or operational impacts \cite{lalor_should_2024}. Moreover, protected attributes like religion, gender identity, and sexual orientation are often misreported, incompletely recorded, or entirely absent due to sociocultural issues and inconsistent data collection practices \cite{kronk_transgender_2022, ruberg2020data}. For example, socioeconomic status is frequently missing in observational health data, though proxy variables such as insurance type may be available \cite{monuteaux2024evaluation}. To address such limitations, the concept of proxy fairness has been introduced to enable fairness assessments using variables that approximate the protected attributes of interest \cite{guptaProxyFairness2018}. When neither protected attributes nor reliable proxies are available, fairness metrics that account for missing data become essential.  While by no means a remedy for inadequate data, there has been progress in fairness approaches that operate with limited data on protected attributes and that may prove useful for health applications \cite{zhang_fairlisa_2023, roy_fairness_2024}. To date, however, many protected attributes have been omitted from fairness considerations, which underscores the necessity of future methodological work to acknowledge ``a multiplicity of considerations, from privacy preservation, context sensitivity and process fairness, to an awareness of sociotechnical impact and the increasingly important role of inclusive and participatory research processes'' \cite{tomasev2021fairness}. We anticipate future work in these aforementioned directions,  particularly in strategies focused on data collection and promoting data equity in health contexts to establish a strong foundation for fair decision-making.}

\section*{Acknowledgements}
J. Gronsbell is grateful for support of an NSERC Discovery Grant (RGPIN-2021-03734) and a University of Toronto Data Science Institute Seed Funding for Methodologists Grant.

\section*{Author contributions}
JeG conceived and supervised the study.  JeG and JiG drafted the manuscript. PV provided clinical oversight, ZM and CH provided guidance on ML, and HT provided guidance on ethics. All authors provided valuable feedback on the manuscript. 


\clearpage

\begin{singlespace}
\bibliographystyle{wileyNJD-AMA}
\bibliography{references}
\end{singlespace}

\end{document}


\maketitle

\section{FAIRNESS DEFINITION INCOMPATIBILITIES PROOFS}

In this section, we go through the mathematical proofs to understand why mathematically or ethically various fairness metrics cannot be satisfied simultaneously. These proofs are derived by \citet{garg2020fairness} with more detail provided here.

Using conditional probability, we have $P(A \cap B) = P(A \mid B)P(B) = P(B \mid A)P(A)$. Then, we can rewrite the two following formulas:
\begin{equation}
\label{eqn:1}
\begin{split}
P(Y \mid D \cap G) \times P(D \mid G) & = \frac{P(Y \cap D \cap G)}{P(D \cap G)} \times \frac{P(D \cap G)}{P(G)} \\
& = \frac{P(Y \cap D \cap G)}{P(G)} \\
& = P(Y \cap D \mid G) 
\end{split}
\end{equation}

\begin{equation}
\label{eqn:2}
\begin{split}
P(D \mid Y \cap G) \times P(Y \mid G) & = \frac{P(D \cap Y \cap G)}{P(Y \cap G)} \times \frac{P(Y \cap G)}{P(G)} \\
& = \frac{P(D \cap Y \cap G)}{P(G)} \\
& = P(D \cap Y \mid G) 
\end{split}
\end{equation}

Using equations \ref{eqn:1} and \ref{eqn:2}, we have:
\begin{equation}
\label{eqn:3}
P(Y \cap D \mid G) = \underbrace{P(Y \mid D \cap G)}_{\text{Predictive Parity}} \times \underbrace{P(D \mid G)}_{\text{Statistical Parity}} = \underbrace{P(D \mid Y \cap G)}_{\text{Equalized Odds}} \times \underbrace{P(Y \mid G)}_{\text{Base rate}}
\end{equation}

Using equation \ref{eqn:3}, we will demonstrate the incompatibilities with metrics assuming we have unequal base rates.

\subsection{Statistical Parity, Equalized Odds and Predictive Parity}

First, assuming unequal base rates we have:
\begin{align*}
P(Y=1, D=1 \mid G) &= P(Y=1 \mid D=1, G) \times P(1 \mid G) \\
\Rightarrow \frac{P(Y=1, D=1, G)}{P(G)} &= P(Y=1 \mid D=1, G) \times P(D=1,G) \\
\Rightarrow \frac{P(Y=1, D=1, G)}{P(Y=1, G)} \frac{P(Y=1, G)}{P(G)} &= P(Y=1 \mid D=1, G) \times P(D=1,G) \\
\Rightarrow P(D=1 \mid Y=1, G) \times P(Y=1 \mid G) &= P(Y=1 \mid D=1, G) \times P(D=1,G) \\
\Rightarrow P(Y=1 \mid D=1, G) &= \frac{P(D=1 \mid Y=1, G) \times P(Y=1 \mid G)}{P(D=1 \mid G)}
\end{align*}

Let our predictor satisfy equalized odds and statistical parity. Then, by definition, $TPR_0 = TPR_1 = TPR$, $FPR_0 = FPR_1 = FPR$ and $P(D=1 \mid G=0) = P(D=1 \mid G=1) = P(D=1)$ respectively. Assuming these two metrics are satisfied, we investigate PPV for predictive parity by taking the difference of the PPV of the two groups.
\begin{align*}
P(Y=1 \mid D=1, G=0) - P(Y=1 \mid D=1, G=1) &= \frac{TPR_0P(Y=1 \mid G=0) - TPR_1 P(Y=1 \mid G=1)}{P(D=1)} \\
                                            &= \frac{TPR[P(Y=1 \mid G=0) - P(Y=1 \mid G=1)]}{P(D=1)} \\
\end{align*}

The second step is due to equalized odds resulting in $TPR_0 = TPR_1 = TPR$. To have predictive parity hold, we need $P(Y=1 \ \mid D=1, G=0) = P(Y=1 \mid D=1, G=1)$, i.e. $PPV_0 = PPV_1 = PPV$. From our equation above, this requires either $TPR=0$ which would render our predictor useless if it was not predicting any positive classes correctly or we need equal base rates. With unequal base rates, the three metrics cannot be satisfied. 

\subsection{Statistical Parity, Predictive Parity}
Let statistical parity hold, so we have $P(D=1 \mid G=0) = P(D=1 \mid G=1) = P(D=1)$. Again, we take the difference in PPV between the groups to investigate predictive parity. This gives us:
\begin{equation*}
P(Y=1 \mid D=1, G=0) - P(Y=1 \mid D=1, G=1) = \frac{TPR_0 P(Y=1 \mid G=0) - TPR_1P(Y=1 \mid G=1)}{P(D=1)}
\end{equation*}

Assuming predictive parity holds, we would have:
\begin{align*}
0 &= TPR_0P(Y=1 \mid G=0) - TPR_1 P(Y=1 \mid G=1) \\
\Rightarrow TPR_0 P(Y=1 \mid G=0) &= TPR_1 P(Y=1 \mid G=1) \\
\Rightarrow \frac{TPR_0}{TPR_1} &= \frac{P(Y=1 \mid G=1)}{P(Y=1 \mid G=0)} \\
&= \frac{\text{Base rate of Group 1}}{\text{Base rate of Group 0}}
\end{align*}

Thus, both metrics can hold with equal base rates. While they can also be satisfied with unequal base rates, this would result in the $TPR$ of one group being higher than the other which makes our model less effective.

\subsection{Equalized Odds and Predictive Parity}
For equalized odds, we have $TPR_0 = TPR_1 = TPR$, $FPR_0 = FPR_1 = FPR$. Predictive parity is satisfied if $PPV_0 = PPV_1 = PPV$. Then, we have
\begin{align*}
P(D=1 \mid G) &= \sum_Y P(D=1 \mid Y, G)P(Y, G)  \\
              &= P(D=1 \mid Y=1, G)P(Y=1, G)P(Y=1 \mid G) + P(D=1 \mid Y=0, G)P(Y=0 \mid G) \\
              &= TPR(Y=1 \mid G) + FPR P(Y=0 \mid G)
\end{align*}

Using equation \ref{eqn:3}, we have:
\begin{align*}
P(D \mid Y, G) \times P(Y \mid G) & = P(Y \mid D,G) \times P(D\mid G) \\
\Rightarrow P(D=1 \mid Y=1, G=0) P(Y=1 \mid G=0) & = P(Y=1 \mid D=1, G=0)P(D=1 \mid G=0) \\
\Rightarrow P(D=1 \mid Y=1, G=0) P(Y=1 \mid G=0) & = P(Y=1 \mid D=1, G=0)[TPR_0P(Y=1 \mid G=0) \\
& \quad + FPR_0 P(Y=0 \mid G=0)] \\
\Rightarrow TPR_0 P(Y=1 \mid G=0) & = P(Y=1 \mid D=1, G=0)[TPR_0P(Y=1 \mid G=0) \\ 
& \quad + FPR_0(1- P(Y=1 \mid G=0))] \\
\Rightarrow TPR_0 P(Y=1 \mid G=0) & = P(Y=1 \mid D=1, G=0)[TPR_0P(Y=1 \mid G=0) \\ 
& \quad + FPR_0 - FPR_0P(Y=1 \mid G=0))] \\
\Rightarrow P(Y=1 \mid G=0)[TPR_0 - PPV_0TPR_0 + PPV_0FPR_0] & = PPV_0 FPR_0 \\
\Rightarrow P(Y=1 \mid G=0)[TPR_0(1-PPV_0) + PPV_0FPR_0] & = PPV_0 FPR_0 \\
\Rightarrow P(Y=1 \mid G=0) & = \frac{PPV_0 FPR_0}{TPR_0(1-PPV_0) + PPV_0FPR_0}\\
\end{align*}

Similarly, $P(Y=1 \mid G=1) = \frac{PPV_1 FPR_1}{TPR_1(1-PPV_1) + PPV_1FPR_1}$. Thus, for both to hold, we would need $P(Y=1 \mid G=0) = P(Y=1 \mid G=1)$ which would mean equal base rates. If we related equalized odds to equal opportunity, then we would only require $TPR_0 = TPR_1 = TPR$. Then, the two formulas derived are possible meaning both metrics can be satisfied with unequal base rates; however, this would require $FPR_0 \neq FPR_1$. The associated implications with this inequality should be considered carefully as this could be biased.

\subsection{Equalized Odds and Statistical Parity}
For equalized odds, we have $TPR_0 = TPR_1 = TPR$, $FPR_0 = FPR_1 = FPR$. This gives us:
\begin{align*}
P(D=1 \mid = G) &= TPR[P(Y=1 \mid G)] + FPR[P(Y=0 \mid G)]\\
\Rightarrow P(D=1 \mid G=0) - P(D=1 \mid G=1) &= TPR[P(Y=1 \mid G=0) - P(Y=1 \mid G=1)] +  \\
& \quad FPR[P(Y=0 \mid G=0) - P(Y=0 \mid G=1)] \\
&= TPR[P(Y=1 \mid G=0) - P(Y=1 \mid G=1)] +  \\
& \quad FPR[(1-P(Y=1 \mid G=0)) - (1-P(Y=1 \mid G=1))] \\
&= TPR[P(Y=1 \mid G=0) - P(Y=1 \mid G=1)] +  \\
& \quad FPR[(P(Y=1 \mid G=1) - P(Y=1 \mid G=0)] \\
&= (TPR-FPR)[[P(Y=1 \mid G=0) - P(Y=1 \mid G=1)]
\end{align*}

Now, for statistical parity to hold, we need $P(D=1 \mid G=0) = P(D=1 \mid G=1) = P(D=1)$. So, we need either $TPR=FPR$ or $P(Y=1 \mid G=0) = P(Y=1 \mid G=1)$. 

The first is not ideal for a predictor as we would want more correct predictions. The latter contradicts our initial assumption of unequal base rates. 

\subsection{Predictive Parity and Calibration}
Consider a threshold $c \in [0,1]$ such that $\forall s > c$ we would have $D=1$ and $D=0$ otherwise. Then, we can rewrite predictive parity as:
\begin{equation*}
P(Y=1 \mid D=1, G) = P(Y=1 \mid S > c, G)
\end{equation*}

Now, we note that 
\begin{align*}
P(Y, S > c, G) &= \int_{c}^{1} P(Y, S \mid G)\, ds\ \\
               &= \int_{c}^{1} P(Y \mid S, G) P(S \mid G) \, ds\
\end{align*}

Then, we have:
\begin{equation*}
\underbrace{P(Y \mid S > c, G)}_{\text{Predictive Parity}} = \frac{\int_{c}^{1} \overbrace{P(Y \mid S, G)}^{\text{Calibration Term}}P(S \mid G) ds}{\int_{c}^{1} P(S \mid G) ds}
\end{equation*}

Now, when $P(Y \mid S, G=1) = P(Y \mid S, G=0)$, the probability distribution of the score, $P(S \mid G)$ can cause predictive parity to be unsatisfied. Thus, while it is possible to satisfy both of these metrics, it is also not always guaranteed and dependent on the threshold, $c$, applied.

\subsection{Statistical Parity and Calibration}
Statistical parity is satisfied when $P(D=1 \mid G=0) = P(D=0 \mid G=1) = P(D=1)$. Calibration as shown above is $P(Y \mid S, G=0) = P(Y \mid S, G=1)$. Hence, we want to be able to have these two metrics in terms of the other. 
\begin{equation}
\label{eqn:calib1}
P(Y=1 \mid D=1, G) = P(Y=1 \mid S > c, G) \\
\end{equation}
and 
\begin{equation}
\label{eqn:calib2}
P(Y=1 \mid D=1, G) = \frac{P(Y=1, D=1, G)}{P(D=1, G)}
\end{equation}

Then, using equations \ref{eqn:calib1} and \ref{eqn:calib2} and setting them equal to each other, we get:
\begin{align*}
P(Y=1 \mid S > c, G) &= \frac{P(Y=1, D=1, G)}{P(D=1, G)} \\
\Rightarrow P(Y=1 \mid S > c, G) &= \frac{P(Y=1, D=1, G)}{P(D=1)} \\
\Rightarrow \frac{\int_{c}^{1} \overbrace{P(Y \mid S, G)}^{\text{Calibration Term}}P(S \mid G) ds}{\int_{c}^{1} P(S \mid G) ds} &= \frac{P(Y=1, D=1, G)}{\underbrace{P(D=1)}_{\text{Statistical Parity}}}
\end{align*}

Similarly to the previous case, when $P(Y \mid S, G=1) = P(Y \mid S, G=0)$, the probability distribution of the score, $P(S \mid G)$ can cause statistical parity to be unsatisfied. Thus, both metrics could be satisfied but this will depend on the threshold $c$ and the probability distribution of the score. 

\section*{MISPECIFIDED MODEL SIMULATIONS}

\subsection*{Simulation Settings}

\begin{table}[H]
\centering
\begin{tabular}{|p{0.02\textwidth}|p{0.3\textwidth}|p{0.10\textwidth}|p{0.45\textwidth}|p{0.11\textwidth}|}
\hline 
 & \begin{center}
\textbf{Purpose of Simulation}
\end{center}
 & \textbf{Base Rate} & \textbf{Score Generating Function} & \begin{center}
\textbf{Threshold}
\end{center}
 \\
\hline 
  1 & To demonstrate that with equal base rates, it is possible to have these four metrics hold simultaneously.  & (0.5, 0.5) & $S \mid G, Y=1 \sim 0.15N(0.2, 0.85^{2}) + 0.85N(0.85, 1.5^2) \newline S \mid G, Y=0 \sim 0.25N(1, 0.5^{2}) + 0.75(0.95, 0.5^2)$ & 0.5 \\
\hline 
  2 & To demonstrate that by changing the base rates, we have incompatibilities within our metrics. We show that we have equal TPR and FPR so EO holds but SP and PP fail.  & (0.7, 0.5) & $S \mid G, Y=1 \sim 0.15N(0.2, 0.85^{2}) + 0.85N(0.85, 1.5^2) \newline S \mid G, Y=0 \sim 0.25N(1, 0.5^{2}) + 0.75(0.95, 0.5^2)$  & 0.5 \\
\hline 
  3 & To demonstrate that with unequal base rates, EO and SP can both hold; however, it is not desirable as it results in TPR=FPR for both groups. Furthermore, while EO and SP both hold, PP fails to hold in this setting. & (0.7, 0.5) & $S \mid G, Y \sim 0.25N(1, 0.5^{2}) + 0.75(0.95, 0.5^2)$  & 0.5 \\
\hline 
  4 & To demonstrate how if the distributions of generating $S \mid Y=1$ differs amongst groups, then predictive parity can still hold true but SP and EO will fail. This shows the converse of Setting 3 in terms of which metrics hold. & (0.7, 0.5) & 
 $ S \mid G=1, Y=1 \sim 0.15N(0.01, 0.95^{2}) + 0.85N(0.01, 1.5^2)\newline 
   S \mid G=0, Y=1 \sim 0.15N(0.95, 0.01^{2}) + 0.85N(1.5, 0.5^2) \newline
   S \mid G, Y=0 \sim 0.25N(1, 0.5^{2}) + 0.75(0.95, 0.5^2)$ & 0.5 \\
 \hline
  5 & To demonstrate if we generate our distributions of $S \mid Y$ entirely differently for both groups, we can have a scenario where both SP and PP hold. However, not only does EO fail, but one group will have a TPR higher than the other group (similarly lower FPR) which is not ideal. & (0.7, 0.5) &  
  $ S \mid G=1, Y=1 \sim 0.15N(0.7, 0.95^{2}) + 0.85N(0.8, 1.5^2)\newline 
    S \mid G=1, Y=0 \sim 0.25N(0.5, 0.5^2) + 0.75N(0.8, 0.25^2) \newline
    S \mid G=0, Y=1 \sim 0.15N(0.5, 0.01^{2}) + 0.85N(0.95, 0.5^2) \newline
    S \mid G, Y=0 \sim 0.25N(0.1 0.75^{2}) + 0.75(0.3, 0.85^2)$ & 0.5 \\
 \hline
\end{tabular}
\caption{Descriptions of Misspecified Model Simulation Settings. Each setting is created to illustrate an incompatibility between fairness metrics.}
\footnotesize{$p_0$: Prevalence of Y for $G=0$; $p_1$: Prevalence of Y for $G=1$; $\mu_{YG}$: The mean for the distribution of $S \mid Y, G$; $t$: Threshold for converting probability score to binary decision}
\label{tab:4}
\end{table}

\subsection*{Results}

\begin{figure}[H]
    \centering
    \includegraphics[scale=0.22]{images/mis_sim_comparison.png}
    \caption{Comparison of Fairness Metrics from Misspecified Model Simulation. The metrics for both groups are shown in an individual plot for each setting. The bars compare whether the fairness metrics are equal between both sensitive groups.}
    \footnotesize{BS: Brier's Score; FPR: False Positive Rate; MR: Misclassification Rate; NPV: Negative Predictive Value; PPV: Positive Predictive Value; PP: Probability of Positive Prediction; TPR: True Positive Rate}
    \label{figure:misModel}
\end{figure}

\section{DATA PREPROCESSING}

The data is extracted from the MIMIC-III database which consists of discharge summaries. The data is filtered to only include adult patients by removing newborn and NICU patients. Furthermore, patients are identified as frequent flyers or not-frequent flyers based on whether they had more than 3 ICU visits in a 365-day consecutive period. These two categories were used to obtain random samples of discharge summaries from each to account for different types of patients. The first 415 ICU frequent flyer discharge summaries and 313 consequent visits of the same patients were randomly sampled. From the non-frequent flyers, 882 random summaries were extracted. 

From each discharge summary, we want to extract the important medical terms; this is done using Apache clinical Text Analysis and Knowledge Extraction System (cTAKES). First, the processing steps were completed using a package `medspaCy`. This is a package to perform natural language processing (NLP) and ttxt processing tasks specifically in the clinical domain \cite{medspacy}. Processing steps include tokenizing the sentences and phrases to individual linguistic units, normalizing them to their base form (i.e. plural words) and tagging them for their parts-of-speech (PoS). A named entity recognition algorithm is then used in `medspaCy` to extract concepts from the clinical text \cite{medspacy}. These medical concepts are then filtered using a package `QuickULMS` using concept mapping to only include those that a medical concept unique identifier (CUI) within the Unified Medical Language System (UMLS) \citep{soldaini2016quickumls}. Lastly, in order to use these CUIs as input in our model, they need to be transformed into vectors known as word embeddings. This is done using a package `cui2vec` which creates word embeddings for each CUI in each discharge summary. Word embeddings translate these concepts into multidimensional vectors that represent semantic similarity using closeness in a vector space. With `cui2vec`, this is translated using a large collection of medical data \cite{beam2019clinical}.

\section{CAUSAL DEFINITIONS OF FAIRNESS}

\subsection*{Notation}
\begin{figure}[hbt!]
    \centering
    \includegraphics[scale=0.28]{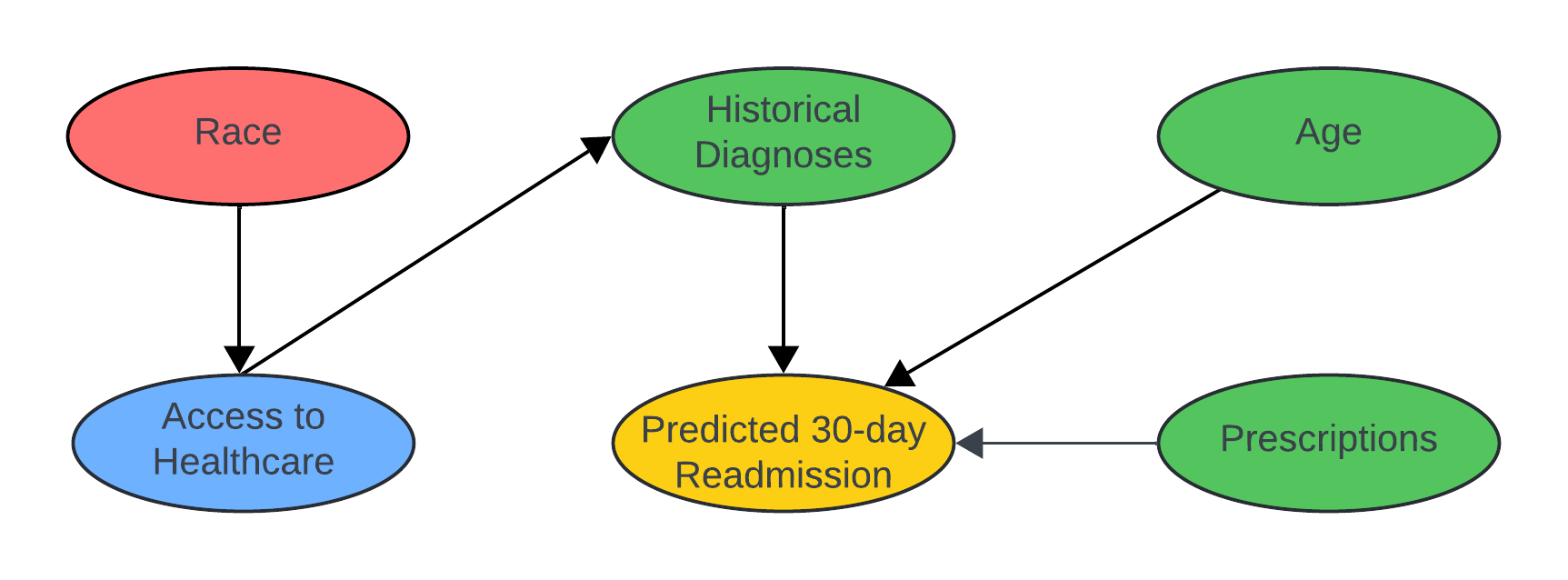}
    \caption{A causal DAG illustrating a hypothetical process for classifying a disease in patients. The DAG shows the different paths from the patient data to classification of disease through sensitive and non-sensitive attributes.}
    \label{figure:dag}
\end{figure}

For notions of fairness using causal reasoning, directed, acyclic graphs (DAGs) are used. Let $G'=(V,E)$ be a graph with a set of nodes, $V$, and a set of directed edges, $E$. Let the nodes represent $\bf{X}$, the attributes of a patient, and the edges represent the relationship between the given attributes. These edges can be used to understand the causal impact an attribute has on the decision being made \citep{castelnovo2022clarification}. While we are not able to test causal definitions of fairness with an existing classifier, causal graphs can be used to initially build fair classifiers and ML algorithms \citep{verma2018fairness}. These causal graphs require thorough domain knowledge to understand the relationships and implications between attributes and the decision. They can then be used to further explore how social goals are being attained using a mathematical framework. For example, medical practitioners or policymakers can use the DAG to determine if a causal pathway from a sensitive attribute is admissible or problematic. We demonstrate the notions of fairness through causal definitions with a simple graph using the phenotyping example. Our graph includes our protected attribute ($G$), age, number of pre-existing conditions,  number of doctor visits in the last year and the predicted outcome of having a kidney disease or not ($D$). In causal graphs, there are proxy and resolving attributes. A proxy attribute can be used to derive another attribute while a resolving attribute is influenced by the protected attribute in a non-discriminatory way \citep{verma2018fairness, kilbertus2017avoiding}. In our scenario, a proxy attribute for race could be the number of doctor visits in the last year; access to medical care can vary based on race due to barriers including socioeconomic status, insurance status, trust in the healthcare system, etc. Using the graph in Figure \ref{figure:dag}, we define several notions of fairness in the context of our phenotyping scenario.

\subsection*{Counterfactual Fairness}
In terms of Figure \ref{figure:dag}, counterfactual fairness would mean that our classification of the disease does not depend on any attribute that is dependent on the protected attribute $G$ \citep{verma2018fairness, mitchell2021algorithmic}. In our graph, we have that the classification of a disease is impacted by the number of doctor visits which is a descendent of race; hence, this would not be counterfactually fair. Intuitively, this would mean that if we had a White patient with $\bf{X}=\bf{x_0}$ and only counterfactually altered their race, we would expect them to have the same probability of being classified as having the disease \citep{castelnovo2022clarification, nilforoshan2022causal}.

\subsection*{No unresolved discrimination}
No unresolved discrimination is satisfied if a causal graph has no discriminatory paths from our sensitive attribute, $G$, to the classification of disease \citep{verma2018fairness}. A path through a resolved attribute would thus be acceptable since it is influenced in a non-discriminatory manner \citep{castelnovo2022clarification}. In Figure \ref{figure:dag}, we have a path from race to number of doctor visits to our classification where number of doctor visits is not a resolving attribute; thus, we have unresolved discrimination. 

\subsection*{No proxy discrimination}
A causal graph exhibits no proxy discrimination if there is no path from the sensitive attribute, $G$, to our classification of disease $D$, through a proxy \citep{verma2018fairness}. Using the same example of the path from race to the number of doctor visits to our classification of disease, since we have number of doctor visits as a proxy, Figure \ref{figure:dag} is not free of proxy discrimination. 

\subsection*{Fair inference}
Fair inference considers different paths to determine if they are acceptable or not \citep{verma2018fairness}. This requires domain knowledge of the impacts attributes have on one another and on the decision variable. For example, the number of doctor visits could be an indicator of repeated health problems. Although a proxy for race, it could be useful in determining if a patient has a disease and thus could be considered legitimate. After considering each path and classifying it as legitimate or not, a causal graph satisfies fair inference if there are no illegitimate paths from $G$ to $D$. In our case, we do not have any illegitimate paths so Figure \ref{figure:dag} satisfies fair inference. \\

Causal definitions of fairness allow us to take into consideration influences based on the sensitive attribute that may unjustly affect the final decision. This can then be used prior to building the model to determine methods to reduce these disparities so decision-making is more fair.



%
%
%
%
%
\def\bzero{{\bf 0}}
\def\bone{{\bf 1}}
%
%
%
%
\def\ba{{\mbox{\boldmath$a$}}}
\def\bb{{\bf b}}
\def\bc{{\bf c}}
\def\bd{{\bf d}}
\def\be{{\bf e}}
\def\bdf{{\bf f}}
\def\bg{{\mbox{\boldmath$g$}}}
\def\bh{{\bf h}}
\def\bi{{\bf i}}
\def\bj{{\bf j}}
\def\bk{{\bf k}}
\def\bl{{\bf l}}
\def\bm{{\bf m}}
\def\bn{{\bf n}}
\def\bo{{\bf o}}
\def\bp{{\bf p}}
\def\bq{{\bf q}}
\def\br{{\bf r}}
\def\bs{{\bf s}}
\def\bt{{\bf t}}
\def\bu{{\bf u}}
\def\bv{{\bf v}}
\def\bw{{\bf w}}
\def\bx{{\bf x}}
\def\by{{\bf y}}
\def\bz{{\bf z}}
\def\bA{{\bf A}}
\def\bB{{\bf B}}
\def\bC{{\bf C}}
\def\bD{{\bf D}}
\def\bE{{\bf E}}
\def\bF{{\bf F}}
\def\bG{{\bf G}}
\def\bH{{\bf H}}
\def\bI{{\bf I}}
\def\bJ{{\bf J}}
\def\bK{{\bf K}}
\def\bL{{\bf L}}
\def\bM{{\bf M}}
\def\bN{{\bf N}}
\def\bO{{\bf O}}
\def\bP{{\bf P}}
\def\bQ{{\bf Q}}
\def\bR{{\bf R}}
\def\bS{{\bf S}}
\def\bT{{\bf T}}
\def\bU{{\bf U}}
\def\bV{{\bf V}}
\def\bW{{\bf W}}
\def\bX{{\bf X}}
\def\bY{{\bf Y}}
\def\bZ{{\bf Z}}
\def\smbZ{\scriptstyle{\bf Z}}
\def\smM{\scriptstyle{M}}
\def\smN{\scriptstyle{N}}
\def\smbT{\scriptstyle{\bf T}}
%
%
%
%
\def\thick#1{\hbox{\rlap{$#1$}\kern0.25pt\rlap{$#1$}\kern0.25pt$#1$}}
\def\balpha{\boldsymbol{\alpha}}
\def\bbeta{\boldsymbol{\beta}}
\def\bgamma{\boldsymbol{\gamma}}
\def\bdelta{\boldsymbol{\delta}}
\def\bepsilon{\boldsymbol{\epsilon}}
\def\bvarepsilon{\boldsymbol{\varepsilon}}
\def\bzeta{\boldsymbol{\zeta}}
\def\bdeta{\boldsymbol{\eta}}
\def\btheta{\boldsymbol{\theta}}
\def\biota{\boldsymbol{\iota}}
\def\bkappa{\boldsymbol{\kappa}}
\def\blambda{\boldsymbol{\lambda}}
\def\bmu{\boldsymbol{\mu}}
\def\bnu{\boldsymbol{\nu}}
\def\bxi{\boldsymbol{\xi}}
\def\bomicron{\boldsymbol{\omicron}}
\def\bpi{\boldsymbol{\pi}}
\def\brho{\boldsymbol{\rho}}
\def\bsigma{\boldsymbol{\sigma}}
\def\btau{\boldsymbol{\tau}}
\def\bupsilon{\boldsymbol{\upsilon}}
\def\bphi{\boldsymbol{\phi}}
\def\bchi{\boldsymbol{\chi}}
\def\bpsi{\boldsymbol{\psi}}
\def\bomega{\boldsymbol{\omega}}
\def\bAlpha{\boldsymbol{\Alpha}}
\def\bBeta{\boldsymbol{\Beta}}
\def\bGamma{\boldsymbol{\Gamma}}
\def\bDelta{\boldsymbol{\Delta}}
\def\bEpsilon{\boldsymbol{\Epsilon}}
\def\bZeta{\boldsymbol{\Zeta}}
\def\bEta{\boldsymbol{\Eta}}
\def\bTheta{\boldsymbol{\Theta}}
\def\bIota{\boldsymbol{\Iota}}
\def\bKappa{\boldsymbol{\Kappa}}
\def\bLambda{{\boldsymbol{\Lambda}}}
\def\bMu{\boldsymbol{\Mu}}
\def\bNu{\boldsymbol{\Nu}}
\def\bXi{\boldsymbol{\Xi}}
\def\bOmicron{\boldsymbol{\Omicron}}
\def\bPi{\boldsymbol{\Pi}}
\def\bRho{\boldsymbol{\Rho}}
\def\bSigma{\boldsymbol{\Sigma}}
\def\bTau{\boldsymbol{\Tau}}
\def\bUpsilon{\boldsymbol{\Upsilon}}
\def\bPhi{\boldsymbol{\Phi}}
\def\bChi{\boldsymbol{\Chi}}
\def\bPsi{\boldsymbol{\Psi}}
\def\bOmega{\boldsymbol{\Omega}}
%
%
%
\def\smalpha{{{\scriptstyle{\alpha}}}}
\def\smbeta{{{\scriptstyle{\beta}}}}
\def\smgamma{{{\scriptstyle{\gamma}}}}
\def\smdelta{{{\scriptstyle{\delta}}}}
\def\smepsilon{{{\scriptstyle{\epsilon}}}}
\def\smvarepsilon{{{\scriptstyle{\varepsilon}}}}
\def\smzeta{{{\scriptstyle{\zeta}}}}
\def\smdeta{{{\scriptstyle{\eta}}}}
\def\smtheta{{{\scriptstyle{\theta}}}}
\def\smiota{{{\scriptstyle{\iota}}}}
\def\smkappa{{{\scriptstyle{\kappa}}}}
\def\smlambda{{{\scriptstyle{\lambda}}}}
\def\smmu{{{\scriptstyle{\mu}}}}
\def\smnu{{{\scriptstyle{\nu}}}}
\def\smxi{{{\scriptstyle{\xi}}}}
\def\smomicron{{{\scriptstyle{\omicron}}}}
\def\smpi{{{\scriptstyle{\pi}}}}
\def\smrho{{{\scriptstyle{\rho}}}}
\def\smsigma{{{\scriptstyle{\sigma}}}}
\def\smtau{{{\scriptstyle{\tau}}}}
\def\smupsilon{{{\scriptstyle{\upsilon}}}}
\def\smphi{{{\scriptstyle{\phi}}}}
\def\smchi{{{\scriptstyle{\chi}}}}
\def\smpsi{{{\scriptstyle{\psi}}}}
\def\smomega{{{\scriptstyle{\omega}}}}
\def\smAlpha{{{\scriptstyle{\Alpha}}}}
\def\smBeta{{{\scriptstyle{\Beta}}}}
\def\smGamma{{{\scriptstyle{\Gamma}}}}
\def\smDelta{{{\scriptstyle{\Delta}}}}
\def\smEpsilon{{{\scriptstyle{\Epsilon}}}}
\def\smZeta{{{\scriptstyle{\Zeta}}}}
\def\smEta{{{\scriptstyle{\Eta}}}}
\def\smTheta{{{\scriptstyle{\Theta}}}}
\def\smIota{{{\scriptstyle{\Iota}}}}
\def\smKappa{{{\scriptstyle{\Kappa}}}}
\def\smLambda{{{\scriptstyle{\Lambda}}}}
\def\smMu{{{\scriptstyle{\Mu}}}}
\def\smNu{{{\scriptstyle{\Nu}}}}
\def\smXi{{{\scriptstyle{\Xi}}}}
\def\smOmicron{{{\scriptstyle{\Omicron}}}}
\def\smPi{{{\scriptstyle{\Pi}}}}
\def\smRho{{{\scriptstyle{\Rho}}}}
\def\smSigma{{{\scriptstyle{\Sigma}}}}
\def\smTau{{{\scriptstyle{\Tau}}}}
\def\smUpsilon{{{\scriptstyle{\Upsilon}}}}
\def\smPhi{{{\scriptstyle{\Phi}}}}
\def\smChi{{{\scriptstyle{\Chi}}}}
\def\smPsi{{{\scriptstyle{\Psi}}}}
\def\smOmega{{{\scriptstyle{\Omega}}}}
%
%

%
\def\smbalpha{\boldsymbol{{\scriptstyle{\alpha}}}}
\def\smbbeta{\boldsymbol{{\scriptstyle{\beta}}}}
\def\smbgamma{\boldsymbol{{\scriptstyle{\gamma}}}}
\def\smbdelta{\boldsymbol{{\scriptstyle{\delta}}}}
\def\smbepsilon{\boldsymbol{{\scriptstyle{\epsilon}}}}
\def\smbvarepsilon{\boldsymbol{{\scriptstyle{\varepsilon}}}}
\def\smbzeta{\boldsymbol{{\scriptstyle{\zeta}}}}
\def\smbdeta{\boldsymbol{{\scriptstyle{\eta}}}}
\def\smbtheta{\boldsymbol{{\scriptstyle{\theta}}}}
\def\smbiota{\boldsymbol{{\scriptstyle{\iota}}}}
\def\smbkappa{\boldsymbol{{\scriptstyle{\kappa}}}}
\def\smblambda{\boldsymbol{{\scriptstyle{\lambda}}}}
\def\smbmu{\boldsymbol{{\scriptstyle{\mu}}}}
\def\smbnu{\boldsymbol{{\scriptstyle{\nu}}}}
\def\smbxi{\boldsymbol{{\scriptstyle{\xi}}}}
\def\smbomicron{\boldsymbol{{\scriptstyle{\omicron}}}}
\def\smbpi{\boldsymbol{{\scriptstyle{\pi}}}}
\def\smbrho{\boldsymbol{{\scriptstyle{\rho}}}}
\def\smbsigma{\boldsymbol{{\scriptstyle{\sigma}}}}
\def\smbtau{\boldsymbol{{\scriptstyle{\tau}}}}
\def\smbupsilon{\boldsymbol{{\scriptstyle{\upsilon}}}}
\def\smbphi{\boldsymbol{{\scriptstyle{\phi}}}}
\def\smbchi{\boldsymbol{{\scriptstyle{\chi}}}}
\def\smbpsi{\boldsymbol{{\scriptstyle{\psi}}}}
\def\smbomega{\boldsymbol{{\scriptstyle{\omega}}}}
\def\smbAlpha{\boldsymbol{{\scriptstyle{\Alpha}}}}
\def\smbBeta{\boldsymbol{{\scriptstyle{\Beta}}}}
\def\smbGamma{\boldsymbol{{\scriptstyle{\Gamma}}}}
\def\smbDelta{\boldsymbol{{\scriptstyle{\Delta}}}}
\def\smbEpsilon{\boldsymbol{{\scriptstyle{\Epsilon}}}}
\def\smbZeta{\boldsymbol{{\scriptstyle{\Zeta}}}}
\def\smbEta{\boldsymbol{{\scriptstyle{\Eta}}}}
\def\smbTheta{\boldsymbol{{\scriptstyle{\Theta}}}}
\def\smbIota{\boldsymbol{{\scriptstyle{\Iota}}}}
\def\smbKappa{\boldsymbol{{\scriptstyle{\Kappa}}}}
\def\smbLambda{\boldsymbol{{\scriptstyle{\Lambda}}}}
\def\smbMu{\boldsymbol{{\scriptstyle{\Mu}}}}
\def\smbNu{\boldsymbol{{\scriptstyle{\Nu}}}}
\def\smbXi{\boldsymbol{{\scriptstyle{\Xi}}}}
\def\smbOmicron{\boldsymbol{{\scriptstyle{\Omicron}}}}
\def\smbPi{\boldsymbol{{\scriptstyle{\Pi}}}}
\def\smbRho{\boldsymbol{{\scriptstyle{\Rho}}}}
\def\smbSigma{\boldsymbol{{\scriptstyle{\Sigma}}}}
\def\smbTau{\boldsymbol{{\scriptstyle{\Tau}}}}
\def\smbUpsilon{\boldsymbol{{\scriptstyle{\Upsilon}}}}
\def\smbPhi{\boldsymbol{{\scriptstyle{\Phi}}}}
\def\smbChi{\boldsymbol{{\scriptstyle{\Chi}}}}
\def\smbPsi{\boldsymbol{{\scriptstyle{\Psi}}}}
\def\smbOmega{\boldsymbol{{\scriptstyle{\Omega}}}}
%
%
%
%
\def\ahat{{\widehat a}}
\def\bhat{{\widehat b}}
\def\chat{{\widehat c}}
\def\dhat{{\widehat d}}
\def\ehat{{\widehat e}}
\def\fhat{{\widehat f}}
\def\ghat{{\widehat g}}
\def\hhat{{\widehat h}}
\def\ihat{{\widehat i}}
\def\jhat{{\widehat j}}
\def\khat{{\widehat k}}
\def\lhat{{\widehat l}}
\def\mhat{{\widehat m}}
\def\nhat{{\widehat n}}
\def\ohat{{\widehat o}}
\def\phat{{\widehat p}}
\def\qhat{{\widehat q}}
\def\rhat{{\widehat r}}
\def\shat{{\widehat s}}
\def\that{{\widehat t}}
\def\uhat{{\widehat u}}
\def\vhat{{\widehat v}}
\def\what{{\widehat w}}
\def\xhat{{\widehat x}}
\def\yhat{{\widehat y}}
\def\zhat{{\widehat z}}
\def\Ahat{{\widehat A}}
\def\Bhat{{\widehat B}}
\def\Chat{{\widehat C}}
\def\Dhat{{\widehat D}}
\def\Ehat{{\widehat E}}
\def\Fhat{{\widehat F}}
\def\Ghat{{\widehat G}}
\def\Hhat{{\widehat H}}
\def\Ihat{{\widehat I}}
\def\Jhat{{\widehat J}}
\def\Khat{{\widehat K}}
\def\Lhat{{\widehat L}}
\def\Mhat{{\widehat M}}
\def\Nhat{{\widehat N}}
\def\Ohat{{\widehat O}}
\def\Phat{{\widehat P}}
\def\Qhat{{\widehat Q}}
\def\Rhat{{\widehat R}}
\def\Shat{{\widehat S}}
\def\That{{\widehat T}}
\def\Uhat{{\widehat U}}
\def\Vhat{{\widehat V}}
\def\What{{\widehat W}}
\def\Xhat{{\widehat X}}
\def\Yhat{{\widehat Y}}
\def\Zhat{{\widehat Z}}
%
%
%
\def\atilde{{\widetilde a}}
\def\btilde{{\widetilde b}}
\def\ctilde{{\widetilde c}}
\def\dtilde{{\widetilde d}}
\def\etilde{{\widetilde e}}
\def\ftilde{{\widetilde f}}
\def\gtilde{{\widetilde g}}
\def\htilde{{\widetilde h}}
\def\itilde{{\widetilde i}}
\def\jtilde{{\widetilde j}}
\def\ktilde{{\widetilde k}}
\def\ltilde{{\widetilde l}}
\def\mtilde{{\widetilde m}}
\def\ntilde{{\widetilde n}}
\def\otilde{{\widetilde o}}
\def\ptilde{{\widetilde p}}
\def\qtilde{{\widetilde q}}
\def\rtilde{{\widetilde r}}
\def\stilde{{\widetilde s}}
\def\ttilde{{\widetilde t}}
\def\utilde{{\widetilde u}}
\def\vtilde{{\widetilde v}}
\def\wtilde{{\widetilde w}}
\def\xtilde{{\widetilde x}}
\def\ytilde{{\widetilde y}}
\def\ztilde{{\widetilde z}}
\def\Atilde{{\widetilde A}}
\def\Btilde{{\widetilde B}}
\def\Ctilde{{\widetilde C}}
\def\Dtilde{{\widetilde D}}
\def\Etilde{{\widetilde E}}
\def\Ftilde{{\widetilde F}}
\def\Gtilde{{\widetilde G}}
\def\Htilde{{\widetilde H}}
\def\Itilde{{\widetilde I}}
\def\Jtilde{{\widetilde J}}
\def\Ktilde{{\widetilde K}}
\def\Ltilde{{\widetilde L}}
\def\Mtilde{{\widetilde M}}
\def\Ntilde{{\widetilde N}}
\def\Otilde{{\widetilde O}}
\def\Ptilde{{\widetilde P}}
\def\Qtilde{{\widetilde Q}}
\def\Rtilde{{\widetilde R}}
\def\Stilde{{\widetilde S}}
\def\Ttilde{{\widetilde T}}
\def\Utilde{{\widetilde U}}
\def\Vtilde{{\widetilde V}}
\def\Wtilde{{\widetilde W}}
\def\Xtilde{{\widetilde X}}
\def\Ytilde{{\widetilde Y}}
\def\Ztilde{{\widetilde Z}}
%
%
%
%
\def\bahat{{\widehat \ba}}
\def\bbhat{{\widehat \bb}}
\def\bchat{{\widehat \bc}}
\def\bdhat{{\widehat \bd}}
\def\behat{{\widehat \be}}
\def\bfhat{{\widehat \bf}}
\def\bghat{{\widehat \bg}}
\def\bhhat{{\widehat \bh}}
\def\bihat{{\widehat \bi}}
\def\bjhat{{\widehat \bj}}
\def\bkhat{{\widehat \bk}}
\def\blhat{{\widehat \bl}}
\def\bmhat{{\widehat \bm}}
\def\bnhat{{\widehat \bn}}
\def\bohat{{\widehat \bo}}
\def\bphat{{\widehat \bp}}
\def\bqhat{{\widehat \bq}}
\def\brhat{{\widehat \br}}
\def\bshat{{\widehat \bs}}
\def\bthat{{\widehat \bt}}
\def\buhat{{\widehat \bu}}
\def\bvhat{{\widehat \bv}}
\def\bwhat{{\widehat \bw}}
\def\bxhat{{\widehat \bx}}
\def\byhat{{\widehat \by}}
\def\bzhat{{\widehat \bz}}
\def\bAhat{{\widehat \bA}}
\def\bBhat{{\widehat \bB}}
\def\bChat{{\widehat \bC}}
\def\bDhat{{\widehat \bD}}
\def\bEhat{{\widehat \bE}}
\def\bFhat{{\widehat \bF}}
\def\bGhat{{\widehat \bG}}
\def\bHhat{{\widehat \bH}}
\def\bIhat{{\widehat \bI}}
\def\bJhat{{\widehat \bJ}}
\def\bKhat{{\widehat \bK}}
\def\bLhat{{\widehat \bL}}
\def\bMhat{{\widehat \bM}}
\def\bNhat{{\widehat \bN}}
\def\bOhat{{\widehat \bO}}
\def\bPhat{{\widehat \bP}}
\def\bQhat{{\widehat \bQ}}
\def\bRhat{{\widehat \bR}}
\def\bShat{{\widehat \bS}}
\def\bThat{{\widehat \bT}}
\def\bUhat{{\widehat \bU}}
\def\bVhat{{\widehat \bV}}
\def\bWhat{{\widehat \bW}}
\def\bXhat{{\widehat \bX}}
\def\bYhat{{\widehat \bY}}
\def\bZhat{{\widehat \bZ}}
%
%
%
%
%
\def\batilde{{\widetilde \ba}}
\def\bbtilde{{\widetilde \bb}}
\def\bctilde{{\widetilde \bc}}
\def\bdtilde{{\widetilde \bd}}
\def\betilde{{\widetilde \be}}
\def\bftilde{{\widetilde \bf}}
\def\bgtilde{{\widetilde \bg}}
\def\bhtilde{{\widetilde \bh}}
\def\bitilde{{\widetilde \bi}}
\def\bjtilde{{\widetilde \bj}}
\def\bktilde{{\widetilde \bk}}
\def\bltilde{{\widetilde \bl}}
\def\bmtilde{{\widetilde \bm}}
\def\bntilde{{\widetilde \bn}}
\def\botilde{{\widetilde \bo}}
\def\bptilde{{\widetilde \bp}}
\def\bqtilde{{\widetilde \bq}}
\def\brtilde{{\widetilde \br}}
\def\bstilde{{\widetilde \bs}}
\def\bttilde{{\widetilde \bt}}
\def\butilde{{\widetilde \bu}}
\def\bvtilde{{\widetilde \bv}}
\def\bwtilde{{\widetilde \bw}}
\def\bxtilde{{\widetilde \bx}}
\def\bytilde{{\widetilde \by}}
\def\bztilde{{\widetilde \bz}}
\def\bAtilde{{\widetilde \bA}}
\def\bBtilde{{\widetilde \bB}}
\def\bCtilde{{\widetilde \bC}}
\def\bDtilde{{\widetilde \bD}}
\def\bEtilde{{\widetilde \bE}}
\def\bFtilde{{\widetilde \bF}}
\def\bGtilde{{\widetilde \bG}}
\def\bHtilde{{\widetilde \bH}}
\def\bItilde{{\widetilde \bI}}
\def\bJtilde{{\widetilde \bJ}}
\def\bKtilde{{\widetilde \bK}}
\def\bLtilde{{\widetilde \bL}}
\def\bMtilde{{\widetilde \bM}}
\def\bNtilde{{\widetilde \bN}}
\def\bOtilde{{\widetilde \bO}}
\def\bPtilde{{\widetilde \bP}}
\def\bQtilde{{\widetilde \bQ}}
\def\bRtilde{{\widetilde \bR}}
\def\bStilde{{\widetilde \bS}}
\def\bTtilde{{\widetilde \bT}}
\def\bUtilde{{\widetilde \bU}}
\def\bVtilde{{\widetilde \bV}}
\def\bWtilde{{\widetilde \bW}}
\def\bXtilde{{\widetilde \bX}}
\def\bYtilde{{\widetilde \bY}}
\def\bZtilde{{\widetilde \bZ}}
%
%
%
%
%
%
\def\alphahat{{\widehat\alpha}}
\def\betahat{{\widehat\beta}}
\def\gammahat{{\widehat\gamma}}
\def\deltahat{{\widehat\delta}}
\def\epsilonhat{{\widehat\epsilon}}
\def\varepsilonhat{{\widehat\varepsilon}}
\def\zetahat{{\widehat\zeta}}
\def\etahat{{\widehat\eta}}
\def\thetahat{{\widehat\theta}}
\def\iotahat{{\widehat\iota}}
\def\kappahat{{\widehat\kappa}}
\def\lambdahat{{\widehat\lambda}}
\def\muhat{{\widehat\mu}}
\def\nuhat{{\widehat\nu}}
\def\xihat{{\widehat\xi}}
\def\omicronhat{{\widehat\omicron}}
\def\pihat{{\widehat\pi}}
\def\rhohat{{\widehat\rho}}
\def\sigmahat{{\widehat\sigma}}
\def\tauhat{{\widehat\tau}}
\def\upsilonhat{{\widehat\upsilon}}
\def\phihat{{\widehat\phi}}
\def\chihat{{\widehat\chi}}
\def\psihat{{\widehat\psi}}
\def\omegahat{{\widehat\omega}}
\def\Alphahat{{\widehat\Alpha}}
\def\Betahat{{\widehat\Beta}}
\def\Gammahat{{\widehat\Gamma}}
\def\Deltahat{{\widehat\Delta}}
\def\Epsilonhat{{\widehat\Epsilon}}
\def\Zetahat{{\widehat\Zeta}}
\def\Etahat{{\widehat\Eta}}
\def\Thetahat{{\widehat\Theta}}
\def\Iotahat{{\widehat\Iota}}
\def\Kappahat{{\widehat\Kappa}}
\def\Lambdahat{{\widehat\Lambda}}
\def\Muhat{{\widehat\Mu}}
\def\Nuhat{{\widehat\Nu}}
\def\Xihat{{\widehat\Xi}}
\def\Omicronhat{{\widehat\Omicron}}
\def\Pihat{{\widehat\Pi}}
\def\Rhohat{{\widehat\Rho}}
\def\Sigmahat{{\widehat\Sigma}}
\def\Tauhat{{\widehat\Tau}}
\def\Upsilonhat{{\widehat\Upsilon}}
\def\Phihat{{\widehat\Phi}}
\def\Chihat{{\widehat\Chi}}
\def\Psihat{{\widehat\Psi}}
\def\Omegahat{{\widehat\Omega}}
%
%
%
%
%
\def\alphatilde{{\widetilde\alpha}}
\def\betatilde{{\widetilde\beta}}
\def\gammatilde{{\widetilde\gamma}}
\def\deltatilde{{\widetilde\delta}}
\def\epsilontilde{{\widetilde\epsilon}}
\def\varepsilontilde{{\widetilde\varepsilon}}
\def\zetatilde{{\widetilde\zeta}}
\def\etatilde{{\widetilde\eta}}
\def\thetatilde{{\widetilde\theta}}
\def\iotatilde{{\widetilde\iota}}
\def\kappatilde{{\widetilde\kappa}}
\def\lambdatilde{{\widetilde\lambda}}
\def\mutilde{{\widetilde\mu}}
\def\nutilde{{\widetilde\nu}}
\def\xitilde{{\widetilde\xi}}
\def\omicrontilde{{\widetilde\omicron}}
\def\pitilde{{\widetilde\pi}}
\def\rhotilde{{\widetilde\rho}}
\def\sigmatilde{{\widetilde\sigma}}
\def\tautilde{{\widetilde\tau}}
\def\upsilontilde{{\widetilde\upsilon}}
\def\phitilde{{\widetilde\phi}}
\def\chitilde{{\widetilde\chi}}
\def\psitilde{{\widetilde\psi}}
\def\omegatilde{{\widetilde\omega}}
\def\Alphatilde{{\widetilde\Alpha}}
\def\Betatilde{{\widetilde\Beta}}
\def\Gammatilde{{\widetilde\Gamma}}
\def\Deltatilde{{\widetilde\Delta}}
\def\Epsilontilde{{\widetilde\Epsilon}}
\def\Zetatilde{{\widetilde\Zeta}}
\def\Etatilde{{\widetilde\Eta}}
\def\Thetatilde{{\widetilde\Theta}}
\def\Iotatilde{{\widetilde\Iota}}
\def\Kappatilde{{\widetilde\Kappa}}
\def\Lambdatilde{{\widetilde\Lambda}}
\def\Mutilde{{\widetilde\Mu}}
\def\Nutilde{{\widetilde\Nu}}
\def\Xitilde{{\widetilde\Xi}}
\def\Omicrontilde{{\widetilde\Omicron}}
\def\Pitilde{{\widetilde\Pi}}
\def\Rhotilde{{\widetilde\Rho}}
\def\Sigmatilde{{\widetilde\Sigma}}
\def\Tautilde{{\widetilde\Tau}}
\def\Upsilontilde{{\widetilde\Upsilon}}
\def\Phitilde{{\widetilde\Phi}}
\def\Chitilde{{\widetilde\Chi}}
\def\Psitilde{{\widetilde\Psi}}
\def\Omegatilde{{\widetilde\Omega}}
%
%
%
%
%
%
\def\balphahat{{\widehat\balpha}}
\def\bbetahat{{\widehat\bbeta}}
\def\bgammahat{{\widehat\bgamma}}
\def\bdeltahat{{\widehat\bdelta}}
\def\bepsilonhat{{\widehat\bepsilon}}
\def\bzetahat{{\widehat\bzeta}}
\def\bdetahat{{\widehat\bdeta}}
\def\bthetahat{{\widehat\btheta}}
\def\biotahat{{\widehat\biota}}
\def\bkappahat{{\widehat\bkappa}}
\def\blambdahat{{\widehat\blambda}}
\def\bmuhat{{\widehat\bmu}}
\def\bnuhat{{\widehat\bnu}}
\def\bxihat{{\widehat\bxi}}
\def\bomicronhat{{\widehat\bomicron}}
\def\bpihat{{\widehat\bpi}}
\def\brhohat{{\widehat\brho}}
\def\bsigmahat{{\widehat\bsigma}}
\def\btauhat{{\widehat\btau}}
\def\bupsilonhat{{\widehat\bupsilon}}
\def\bphihat{{\widehat\bphi}}
\def\bchihat{{\widehat\bchi}}
\def\bpsihat{{\widehat\bpsi}}
\def\bomegahat{{\widehat\bomega}}
\def\bAlphahat{{\widehat\bAlpha}}
\def\bBetahat{{\widehat\bBeta}}
\def\bGammahat{{\widehat\bGamma}}
\def\bDeltahat{{\widehat\bDelta}}
\def\bEpsilonhat{{\widehat\bEpsilon}}
\def\bZetahat{{\widehat\bZeta}}
\def\bEtahat{{\widehat\bEta}}
\def\bThetahat{{\widehat\bTheta}}
\def\bIotahat{{\widehat\bIota}}
\def\bKappahat{{\widehat\bKappa}}
\def\bLambdahat{{\widehat\bLambda}}
\def\bMuhat{{\widehat\bMu}}
\def\bNuhat{{\widehat\bNu}}
\def\bXihat{{\widehat\bXi}}
\def\bOmicronhat{{\widehat\bOmicron}}
\def\bPihat{{\widehat\bPi}}
\def\bRhohat{{\widehat\bRho}}
\def\bSigmahat{{\widehat\bSigma}}
\def\bTauhat{{\widehat\bTau}}
\def\bUpsilonhat{{\widehat\bUpsilon}}
\def\bPhihat{{\widehat\bPhi}}
\def\bChihat{{\widehat\bChi}}
\def\bPsihat{{\widehat\bPsi}}
\def\bOmegahat{{\widehat\bOmega}}%
\def\balphahattrans{{\balphahat^{_{\transpose}}}}
\def\bbetahattrans{{\bbetahat^{_{\transpose}}}}
\def\bgammahattrans{{\bgammahat^{_{\transpose}}}}
\def\bdeltahattrans{{\bdeltahat^{_{\transpose}}}}
\def\bepsilonhattrans{{\bepsilonhat^{_{\transpose}}}}
\def\bzetahattrans{{\bzetahat^{_{\transpose}}}}
\def\bdetahattrans{{\bdetahat^{_{\transpose}}}}
\def\bthetahattrans{{\bthetahat^{_{\transpose}}}}
\def\biotahattrans{{\biotahat^{_{\transpose}}}}
\def\bkappahattrans{{\bkappahat^{_{\transpose}}}}
\def\blambdahattrans{{\blambdahat^{_{\transpose}}}}
\def\bmuhattrans{{\bmuhat^{_{\transpose}}}}
\def\bnuhattrans{{\bnuhat^{_{\transpose}}}}
\def\bxihattrans{{\bxihat^{_{\transpose}}}}
\def\bomicronhattrans{{\bomicronhat^{_{\transpose}}}}
\def\bpihattrans{{\bpihat^{_{\transpose}}}}
\def\brhohattrans{{\brhohat^{_{\transpose}}}}
\def\bsigmahattrans{{\bsigmahat^{_{\transpose}}}}
\def\btauhattrans{{\btauhat^{_{\transpose}}}}
\def\bupsilonhattrans{{\bupsilonhat^{_{\transpose}}}}
\def\bphihattrans{{\bphihat^{_{\transpose}}}}
\def\bchihattrans{{\bchihat^{_{\transpose}}}}
\def\bpsihattrans{{\bpsihat^{_{\transpose}}}}
\def\bomegahattrans{{\bomegahat^{_{\transpose}}}}
\def\bAlphahattrans{{\bAlphahat^{_{\transpose}}}}
\def\bBetahattrans{{\bBetahat^{_{\transpose}}}}
\def\bGammahattrans{{\bGammahat^{_{\transpose}}}}
\def\bDeltahattrans{{\bDeltahat^{_{\transpose}}}}
\def\bEpsilonhattrans{{\bEpsilonhat^{_{\transpose}}}}
\def\bZetahattrans{{\bZetahat^{_{\transpose}}}}
\def\bEtahattrans{{\bEtahat^{_{\transpose}}}}
\def\bThetahattrans{{\bThetahat^{_{\transpose}}}}
\def\bIotahattrans{{\bIotahat^{_{\transpose}}}}
\def\bKappahattrans{{\bKappahat^{_{\transpose}}}}
\def\bLambdahattrans{{\bLambdahat^{_{\transpose}}}}
\def\bMuhattrans{{\bMuhat^{_{\transpose}}}}
\def\bNuhattrans{{\bNuhat^{_{\transpose}}}}
\def\bXihattrans{{\bXihat^{_{\transpose}}}}
\def\bOmicronhattrans{{\bOmicronhat^{_{\transpose}}}}
\def\bPihattrans{{\bPihat^{_{\transpose}}}}
\def\bRhohattrans{{\bRhohat^{_{\transpose}}}}
\def\bSigmahattrans{{\bSigmahat^{_{\transpose}}}}
\def\bTauhattrans{{\bTauhat^{_{\transpose}}}}
\def\bUpsilonhattrans{{\bUpsilonhat^{_{\transpose}}}}
\def\bPhihattrans{{\bPhihat^{_{\transpose}}}}
\def\bChihattrans{{\bChihat^{_{\transpose}}}}
\def\bPsihattrans{{\bPsihat^{_{\transpose}}}}
\def\bOmegahattrans{{\bOmegahat^{_{\transpose}}}}%
%
\def\smbalpha{\widehat{\smbalpha}}
\def\smbbetahat{\widehat{\smbbeta}}
\def\smbgammahat{\widehat{\smbgamma}}
\def\smbdeltahat{\widehat{\smbdelta}}
\def\smbepsilonhat{\widehat{\smbepsilon}}
\def\smbvarepsilonhat{\widehat{\smbvarepsilon}}
\def\smbzetahat{\widehat{\smbzeta}}
\def\smbdetahat{\widehat{\smbeta}}
\def\smbthetahat{\widehat{\smbtheta}}
\def\smbiotahat{\widehat{\smbiota}}
\def\smbkappahat{\widehat{\smbkappa}}
\def\smblambdahat{\widehat{\smblambda}}
\def\smbmuhat{\widehat{\smbmu}}
\def\smbnuhat{\widehat{\smbnu}}
\def\smbxihat{\widehat{\smbxi}}
\def\smbomicronhat{\widehat{\smbomicron}}
\def\smbpihat{\widehat{\smbpi}}
\def\smbrhohat{\widehat{\smbrho}}
\def\smbsigmahat{\widehat{\smbsigma}}
\def\smbtauhat{\widehat{\smbtau}}
\def\smbupsilonhat{\widehat{\smbupsilon}}
\def\smbphihat{\widehat{\smbphi}}
\def\smbchihat{\widehat{\smbchi}}
\def\smbpsihat{\widehat{\smbpsi}}
\def\smbomegahat{\widehat{\smbomega}}
\def\smbAlphahat{\widehat{\smbAlpha}}
\def\smbBetahat{\widehat{\smbBeta}}
\def\smbGammahat{\widehat{\smbGamma}}
\def\smbDeltahat{\widehat{\smbDelta}}
\def\smbEpsilonhat{\widehat{\smbEpsilon}}
\def\smbZetahat{\widehat{\smbZeta}}
\def\smbEtahat{\widehat{\smbEta}}
\def\smbThetahat{\widehat{\smbTheta}}
\def\smbIotahat{\widehat{\smbIota}}
\def\smbKappahat{\widehat{\smbKappa}}
\def\smbLambdahat{\widehat{\smbLambda}}
\def\smbMuhat{\widehat{\smbMu}}
\def\smbNuhat{\widehat{\smbNu}}
\def\smbXihat{\widehat{\smbXi}}
\def\smbOmicronhat{\widehat{\smbOmicron}}
\def\smbPihat{\widehat{\smbPi}}
\def\smbRhohat{\widehat{\smbRho}}
\def\smbSigmahat{\widehat{\smbSigma}}
\def\smbTauhat{\widehat{\smbTau}}
\def\smbUpsilonhat{\widehat{\smbUpsilon}}
\def\smbPhihat{\widehat{\smbPhi}}
\def\smbChihat{\widehat{\smbChi}}
\def\smbPsihat{\widehat{\smbPsi}}
\def\smbOmegahat{\widehat{\smbOmega}}
%
%
%
%
%
\def\balphatilde{{\widetilde\balpha}}
\def\bbetatilde{{\widetilde\bbeta}}
\def\bgammatilde{{\widetilde\bgamma}}
\def\bdeltatilde{{\widetilde\bdelta}}
\def\bepsilontilde{{\widetilde\bepsilon}}
\def\bzetatilde{{\widetilde\bzeta}}
\def\bdetatilde{{\widetilde\bdeta}}
\def\bthetatilde{{\widetilde\btheta}}
\def\biotatilde{{\widetilde\biota}}
\def\bkappatilde{{\widetilde\bkappa}}
\def\blambdatilde{{\widetilde\blambda}}
\def\bmutilde{{\widetilde\bmu}}
\def\bnutilde{{\widetilde\bnu}}
\def\bxitilde{{\widetilde\bxi}}
\def\bomicrontilde{{\widetilde\bomicron}}
\def\bpitilde{{\widetilde\bpi}}
\def\brhotilde{{\widetilde\brho}}
\def\bsigmatilde{{\widetilde\bsigma}}
\def\btautilde{{\widetilde\btau}}
\def\bupsilontilde{{\widetilde\bupsilon}}
\def\bphitilde{{\widetilde\bphi}}
\def\bchitilde{{\widetilde\bchi}}
\def\bpsitilde{{\widetilde\bpsi}}
\def\bomegatilde{{\widetilde\bomega}}
\def\bAlphatilde{{\widetilde\bAlpha}}
\def\bBetatilde{{\widetilde\bBeta}}
\def\bGammatilde{{\widetilde\bGamma}}
\def\bDeltatilde{{\widetilde\bDelta}}
\def\bEpsilontilde{{\widetilde\bEpsilon}}
\def\bZetatilde{{\widetilde\bZeta}}
\def\bEtatilde{{\widetilde\bEta}}
\def\bThetatilde{{\widetilde\bTheta}}
\def\bIotatilde{{\widetilde\bIota}}
\def\bKappatilde{{\widetilde\bKappa}}
\def\bLambdatilde{{\widetilde\bLambda}}
\def\bMutilde{{\widetilde\bMu}}
\def\bNutilde{{\widetilde\bNu}}
\def\bXitilde{{\widetilde\bXi}}
\def\bOmicrontilde{{\widetilde\bOmicron}}
\def\bPitilde{{\widetilde\bPi}}
\def\bRhotilde{{\widetilde\bRho}}
\def\bSigmatilde{{\widetilde\bSigma}}
\def\bTautilde{{\widetilde\bTau}}
\def\bUpsilontilde{{\widetilde\bUpsilon}}
\def\bPhitilde{{\widetilde\bPhi}}
\def\bChitilde{{\widetilde\bChi}}
\def\bPsitilde{{\widetilde\bPsi}}
\def\bOmegatilde{{\widetilde\bOmega}}
%
%
%
%
%
\def\abar{\bar{ a}}
\def\bbar{\bar{ b}}
\def\cbar{\bar{ c}}
\def\dbar{\bar{ d}}
\def\ebar{\bar{ e}}
\def\fbar{\bar{ f}}
\def\gbar{\bar{ g}}
\def\hbar{\bar{ h}}
\def\ibar{\bar{ i}}
\def\jbar{\bar{ j}}
\def\kbar{\bar{ k}}
\def\lbar{\bar{ l}}
\def\mbar{\bar{ m}}
\def\nbar{\bar{ n}}
\def\obar{\bar{ o}}
\def\pbar{\bar{ p}}
\def\qbar{\bar{ q}}
\def\rbar{\bar{ r}}
\def\sbar{\bar{ s}}
\def\tbar{\bar{ t}}
\def\ubar{\bar{ u}}
\def\vbar{\bar{ v}}
\def\wbar{\bar{ w}}
\def\xbar{\bar{ x}}
\def\ybar{\bar{ y}}
\def\zbar{\bar{ z}}
\def\Abar{\bar{ A}}
\def\Bbar{\bar{ B}}
\def\Cbar{\bar{ C}}
\def\Dbar{\bar{ D}}
\def\Ebar{\bar{ E}}
\def\Fbar{\bar{ F}}
\def\Gbar{\bar{ G}}
\def\Hbar{\bar{ H}}
\def\Ibar{\bar{ I}}
\def\Jbar{\bar{ J}}
\def\Kbar{\bar{ K}}
\def\Lbar{\bar{ L}}
\def\Mbar{\bar{ M}}
\def\Nbar{\bar{ N}}
\def\Obar{\bar{ O}}
\def\Pbar{\bar{ P}}
\def\Qbar{\bar{ Q}}
\def\Rbar{\bar{ R}}
\def\Sbar{\bar{ S}}
\def\Tbar{\bar{ T}}
\def\Ubar{\bar{ U}}
\def\Vbar{\bar{ V}}
\def\Wbar{\bar{ W}}
\def\Xbar{\bar{ X}}
\def\Ybar{\bar{ Y}}
\def\Zbar{\bar{ Z}}
%
%
%
%
%
\def\babar{\overline{ \ba}}
\def\bbbar{\overline{ \bb}}
\def\bcbar{\overline{ \bc}}
\def\bdbar{\overline{ \bd}}
\def\bebar{\overline{ \be}}
\def\bfbar{\overline{ \bf}}
\def\bgbar{\overline{ \bg}}
\def\bhbar{\overline{ \bh}}
\def\bibar{\overline{ \bi}}
\def\bjbar{\overline{ \bj}}
\def\bkbar{\overline{ \bk}}
\def\blbar{\overline{ \bl}}
\def\bmbar{\overline{ \bm}}
\def\bnbar{\overline{ \bn}}
\def\bobar{\overline{ \bo}}
\def\bpbar{\overline{ \bp}}
\def\bqbar{\overline{ \bq}}
\def\brbar{\overline{ \br}}
\def\bsbar{\overline{ \bs}}
\def\btbar{\overline{ \bt}}
\def\bubar{\overline{ \bu}}
\def\bvbar{\overline{ \bv}}
\def\bwbar{\overline{ \bw}}
\def\bxbar{\overline{ \bx}}
\def\bybar{\overline{ \by}}
\def\bzbar{\overline{ \bz}}
\def\bAbar{\overline{ \bA}}
\def\bBbar{\overline{ \bB}}
\def\bCbar{\overline{ \bC}}
\def\bDbar{\overline{ \bD}}
\def\bEbar{\overline{ \bE}}
\def\bFbar{\overline{ \bF}}
\def\bGbar{\overline{ \bG}}
\def\bHbar{\overline{ \bH}}
\def\bIbar{\overline{ \bI}}
\def\bJbar{\overline{ \bJ}}
\def\bKbar{\overline{ \bK}}
\def\bLbar{\overline{ \bL}}
\def\bMbar{\overline{ \bM}}
\def\bNbar{\overline{ \bN}}
\def\bObar{\overline{ \bO}}
\def\bPbar{\overline{ \bP}}
\def\bQbar{\overline{ \bQ}}
\def\bRbar{\overline{ \bR}}
\def\bSbar{\overline{ \bS}}
\def\bTbar{\overline{ \bT}}
\def\bUbar{\overline{ \bU}}
\def\bVbar{\overline{ \bV}}
\def\bWbar{\overline{ \bW}}
\def\bXbar{\overline{ \bX}}
\def\bYbar{\overline{ \bY}}
\def\bZbar{\overline{ \bZ}}
%
%

%
%
%
\def\asc{{\cal a}}
\def\bsc{{\cal b}}
\def\csc{{\cal c}}
\def\dsc{{\cal d}}
\def\esc{{\cal e}}
\def\dsc{{\cal f}}
\def\gsc{{\cal g}}
\def\hsc{{\cal h}}
\def\isc{{\cal i}}
\def\jsc{{\cal j}}
\def\ksc{{\cal k}}
\def\lsc{{\cal l}}
\def\msc{{\cal m}}
\def\nsc{{\cal n}}
\def\osc{{\cal o}}
\def\psc{{\cal p}}
\def\qsc{{\cal q}}
\def\rsc{{\cal r}}
\def\ssc{{\cal s}}
\def\tsc{{\cal t}}
\def\usc{{\cal u}}
\def\vsc{{\cal v}}
\def\wsc{{\cal w}}
\def\xsc{{\cal x}}
\def\ysc{{\cal y}}
\def\zsc{{\cal z}}
\def\Asc{{\cal A}}
\def\Bsc{{\cal B}}
\def\Csc{{\cal C}}
\def\Dsc{{\cal D}}
\def\Esc{{\cal E}}
\def\Fsc{{\cal F}}
\def\Gsc{{\cal G}}
\def\Hsc{{\cal H}}
\def\Isc{{\cal I}}
\def\Jsc{{\cal J}}
\def\Ksc{{\cal K}}
\def\Lsc{{\cal L}}
\def\Msc{{\cal M}}
\def\Nsc{{\cal N}}
\def\Osc{{\cal O}}
\def\Psc{{\cal P}}
\def\Qsc{{\cal Q}}
\def\Rsc{{\cal R}}
\def\Ssc{{\cal S}}
\def\Tsc{{\cal T}}
\def\Usc{{\cal U}}
\def\Vsc{{\cal V}}
\def\Wsc{{\cal W}}
\def\Xsc{{\cal X}}
\def\Ysc{{\cal Y}}
\def\Zsc{{\cal Z}}
\def\Aschat{\widehat{{\cal A}}}
\def\Bschat{\widehat{{\cal B}}}
\def\Cschat{\widehat{{\cal C}}}
\def\Dschat{\widehat{{\cal D}}}
\def\Eschat{\widehat{{\cal E}}}
\def\Fschat{\widehat{{\cal F}}}
\def\Gschat{\widehat{{\cal G}}}
\def\Hschat{\widehat{{\cal H}}}
\def\Ischat{\widehat{{\cal I}}}
\def\Jschat{\widehat{{\cal J}}}
\def\Kschat{\widehat{{\cal K}}}
\def\Lschat{\widehat{{\cal L}}}
\def\Mschat{\widehat{{\cal M}}}
\def\Nschat{\widehat{{\cal N}}}
\def\Oschat{\widehat{{\cal O}}}
\def\Pschat{\widehat{{\cal P}}}
\def\Qschat{\widehat{{\cal Q}}}
\def\Rschat{\widehat{{\cal R}}}
\def\Sschat{\widehat{{\cal S}}}
\def\Tschat{\widehat{{\cal T}}}
\def\Uschat{\widehat{{\cal U}}}
\def\Vschat{\widehat{{\cal V}}}
\def\Wschat{\widehat{{\cal W}}}
\def\Xschat{\widehat{{\cal X}}}
\def\Yschat{\widehat{{\cal Y}}}
\def\Zschat{\widehat{{\cal Z}}}
\def\Asctilde{\widetilde{{\cal A}}}
\def\Bsctilde{\widetilde{{\cal B}}}
\def\Csctilde{\widetilde{{\cal C}}}
\def\Dsctilde{\widetilde{{\cal D}}}
\def\Esctilde{\widetilde{{\cal E}}}
\def\Fsctilde{\widetilde{{\cal F}}}
\def\Gsctilde{\widetilde{{\cal G}}}
\def\Hsctilde{\widetilde{{\cal H}}}
\def\Isctilde{\widetilde{{\cal I}}}
\def\Jsctilde{\widetilde{{\cal J}}}
\def\Ksctilde{\widetilde{{\cal K}}}
\def\Lsctilde{\widetilde{{\cal L}}}
\def\Msctilde{\widetilde{{\cal M}}}
\def\Nsctilde{\widetilde{{\cal N}}}
\def\Osctilde{\widetilde{{\cal O}}}
\def\Psctilde{\widetilde{{\cal P}}}
\def\Qsctilde{\widetilde{{\cal Q}}}
\def\Rsctilde{\widetilde{{\cal R}}}
\def\Ssctilde{\widetilde{{\cal S}}}
\def\Tsctilde{\widetilde{{\cal T}}}
\def\Usctilde{\widetilde{{\cal U}}}
\def\Vsctilde{\widetilde{{\cal V}}}
\def\Wsctilde{\widetilde{{\cal W}}}
\def\Xsctilde{\widetilde{{\cal X}}}
\def\Ysctilde{\widetilde{{\cal Y}}}
\def\Zsctilde{\widetilde{{\cal Z}}}
\def\bAsc{\mathbf{\cal A}}
\def\bBsc{\mathbf{\cal B}}
\def\bCsc{\mathbf{\cal C}}
\def\bDsc{\mathbf{\cal D}}
\def\bEsc{\mathbf{\cal E}}
\def\bFsc{\mathbf{\cal F}}
\def\bGsc{\mathbf{\cal G}}
\def\bHsc{\mathbf{\cal H}}
\def\bIsc{\mathbf{\cal I}}
\def\bJsc{\mathbf{\cal J}}
\def\bKsc{\mathbf{\cal K}}
\def\bLsc{\mathbf{\cal L}}
\def\bMsc{\mathbf{\cal M}}
\def\bNsc{\mathbf{\cal N}}
\def\bOsc{\mathbf{\cal O}}
\def\bPsc{\mathbf{\cal P}}
\def\bQsc{\mathbf{\cal Q}}
\def\bRsc{\mathbf{\cal R}}
\def\bSsc{\mathbf{\cal S}}
\def\bTsc{\mathbf{\cal T}}
\def\bUsc{\mathbf{\cal U}}
\def\bVsc{\mathbf{\cal V}}
\def\bWsc{\mathbf{\cal W}}
\def\bXsc{\mathbf{\cal X}}
\def\bYsc{\mathbf{\cal Y}}
\def\bZsc{\mathbf{\cal Z}}
\def\bAschat{\widehat{\mathbf{\cal A}}}
\def\bBschat{\widehat{\mathbf{\cal B}}}
\def\bCschat{\widehat{\mathbf{\cal C}}}
\def\bDschat{\widehat{\mathbf{\cal D}}}
\def\bEschat{\widehat{\mathbf{\cal E}}}
\def\bFschat{\widehat{\mathbf{\cal F}}}
\def\bGschat{\widehat{\mathbf{\cal G}}}
\def\bHschat{\widehat{\mathbf{\cal H}}}
\def\bIschat{\widehat{\mathbf{\cal I}}}
\def\bJschat{\widehat{\mathbf{\cal J}}}
\def\bKschat{\widehat{\mathbf{\cal K}}}
\def\bLschat{\widehat{\mathbf{\cal L}}}
\def\bMschat{\widehat{\mathbf{\cal M}}}
\def\bNschat{\widehat{\mathbf{\cal N}}}
\def\bOschat{\widehat{\mathbf{\cal O}}}
\def\bPschat{\widehat{\mathbf{\cal P}}}
\def\bQschat{\widehat{\mathbf{\cal Q}}}
\def\bRschat{\widehat{\mathbf{\cal R}}}
\def\bSschat{\widehat{\mathbf{\cal S}}}
\def\bTschat{\widehat{\mathbf{\cal T}}}
\def\bUschat{\widehat{\mathbf{\cal U}}}
\def\bVschat{\widehat{\mathbf{\cal V}}}
\def\bWschat{\widehat{\mathbf{\cal W}}}
\def\bXschat{\widehat{\mathbf{\cal X}}}
\def\bYschat{\widehat{\mathbf{\cal Y}}}
\def\bZschat{\widehat{\mathbf{\cal Z}}}
\def\afrak{\mathfrak{a}}
\def\bfrak{\mathfrak{b}}
\def\cfrak{\mathfrak{c}}
\def\dfrak{\mathfrak{d}}
\def\efrak{\mathfrak{e}}
\def\ffrak{\mathfrak{f}}
\def\gfrak{\mathfrak{g}}
\def\hfrak{\mathfrak{h}}
\def\ifrak{\mathfrak{i}}
\def\jfrak{\mathfrak{j}}
\def\kfrak{\mathfrak{k}}
\def\lfrak{\mathfrak{l}}
\def\mfrak{\mathfrak{m}}
\def\nfrak{\mathfrak{n}}
\def\ofrak{\mathfrak{o}}
\def\pfrak{\mathfrak{p}}
\def\qfrak{\mathfrak{q}}
\def\rfrak{\mathfrak{r}}
\def\sfrak{\mathfrak{s}}
\def\tfrak{\mathfrak{t}}
\def\ufrak{\mathfrak{u}}
\def\vfrak{\mathfrak{v}}
\def\wfrak{\mathfrak{w}}
\def\xfrak{\mathfrak{x}}
\def\yfrak{\mathfrak{y}}
\def\zfrak{\mathfrak{z}}
\def\Afrak{\mathfrak{ A}}
\def\Bfrak{\mathfrak{ B}}
\def\Cfrak{\mathfrak{ C}}
\def\Dfrak{\mathfrak{ D}}
\def\Efrak{\mathfrak{ E}}
\def\Ffrak{\mathfrak{ F}}
\def\Gfrak{\mathfrak{ G}}
\def\Hfrak{\mathfrak{ H}}
\def\Ifrak{\mathfrak{ I}}
\def\Jfrak{\mathfrak{ J}}
\def\Kfrak{\mathfrak{ K}}
\def\Lfrak{\mathfrak{ L}}
\def\Mfrak{\mathfrak{ M}}
\def\Nfrak{\mathfrak{ N}}
\def\Ofrak{\mathfrak{ O}}
\def\Pfrak{\mathfrak{ P}}
\def\Qfrak{\mathfrak{ Q}}
\def\Rfrak{\mathfrak{ R}}
\def\Sfrak{\mathfrak{ S}}
\def\Tfrak{\mathfrak{ T}}
\def\Ufrak{\mathfrak{ U}}
\def\Vfrak{\mathfrak{ V}}
\def\Wfrak{\mathfrak{ W}}
\def\Xfrak{\mathfrak{ X}}
\def\Yfrak{\mathfrak{ Y}}
\def\Zfrak{\mathfrak{ Z}}

\def\bAfrak{\mathbf{\mathfrak{A}}}
\def\bBfrak{\mathbf{\mathfrak{B}}}
\def\bCfrak{\mathbf{\mathfrak{C}}}
\def\bDfrak{\mathbf{\mathfrak{D}}}
\def\bEfrak{\mathbf{\mathfrak{E}}}
\def\bFfrak{\mathbf{\mathfrak{F}}}
\def\bGfrak{\mathbf{\mathfrak{G}}}
\def\bHfrak{\mathbf{\mathfrak{H}}}
\def\bIfrak{\mathbf{\mathfrak{I}}}
\def\bJfrak{\mathbf{\mathfrak{J}}}
\def\bKfrak{\mathbf{\mathfrak{K}}}
\def\bLfrak{\mathbf{\mathfrak{L}}}
\def\bMfrak{\mathbf{\mathfrak{M}}}
\def\bNfrak{\mathbf{\mathfrak{N}}}
\def\bOfrak{\mathbf{\mathfrak{O}}}
\def\bPfrak{\mathbf{\mathfrak{P}}}
\def\bQfrak{\mathbf{\mathfrak{Q}}}
\def\bRfrak{\mathbf{\mathfrak{R}}}
\def\bSfrak{\mathbf{\mathfrak{S}}}
\def\bTfrak{\mathbf{\mathfrak{T}}}
\def\bUfrak{\mathbf{\mathfrak{U}}}
\def\bVfrak{\mathbf{\mathfrak{V}}}
\def\bWfrak{\mathbf{\mathfrak{W}}}
\def\bXfrak{\mathbf{\mathfrak{X}}}
\def\bYfrak{\mathbf{\mathfrak{Y}}}
\def\bZfrak{\mathbf{\mathfrak{Z}}}

\def\bAfrakhat{\mathbf{\widehat{\mathfrak{A}}}}
\def\bBfrakhat{\mathbf{\widehat{\mathfrak{B}}}}
\def\bCfrakhat{\mathbf{\widehat{\mathfrak{C}}}}
\def\bDfrakhat{\mathbf{\widehat{\mathfrak{D}}}}
\def\bEfrakhat{\mathbf{\widehat{\mathfrak{E}}}}
\def\bFfrakhat{\mathbf{\widehat{\mathfrak{F}}}}
\def\bGfrakhat{\mathbf{\widehat{\mathfrak{G}}}}
\def\bHfrakhat{\mathbf{\widehat{\mathfrak{H}}}}
\def\bIfrakhat{\mathbf{\widehat{\mathfrak{I}}}}
\def\bJfrakhat{\mathbf{\widehat{\mathfrak{J}}}}
\def\bKfrakhat{\mathbf{\widehat{\mathfrak{K}}}}
\def\bLfrakhat{\mathbf{\widehat{\mathfrak{L}}}}
\def\bMfrakhat{\mathbf{\widehat{\mathfrak{M}}}}
\def\bNfrakhat{\mathbf{\widehat{\mathfrak{N}}}}
\def\bOfrakhat{\mathbf{\widehat{\mathfrak{O}}}}
\def\bPfrakhat{\mathbf{\widehat{\mathfrak{P}}}}
\def\bQfrakhat{\mathbf{\widehat{\mathfrak{Q}}}}
\def\bRfrakhat{\mathbf{\widehat{\mathfrak{R}}}}
\def\bSfrakhat{\mathbf{\widehat{\mathfrak{S}}}}
\def\bTfrakhat{\mathbf{\widehat{\mathfrak{T}}}}
\def\bUfrakhat{\mathbf{\widehat{\mathfrak{U}}}}
\def\bVfrakhat{\mathbf{\widehat{\mathfrak{V}}}}
\def\bWfrakhat{\mathbf{\widehat{\mathfrak{W}}}}
\def\bXfrakhat{\mathbf{\widehat{\mathfrak{X}}}}
\def\bYfrakhat{\mathbf{\widehat{\mathfrak{Y}}}}
\def\bZfrakhat{\mathbf{\widehat{\mathfrak{Z}}}}
%
%
%
%
\def\etal{{\em et al.}}
%
%
%
%
%
\def\cumsum{\mbox{cumsum}}
\def\real{{\mathbb R}}
\def\intinfinf{\int_{-\infty}^{\infty}}
\def\intzinf{\int_{0}^{\infty}}
\def\intzt{\int_0^t}
\def\transpose{{\sf \scriptscriptstyle{T}}}
\def\smhalf{{\textstyle{1\over2}}}
\def\third{{\textstyle{1\over3}}}
\def\twothirds{{\textstyle{2\over3}}}
\def\bell{\bmath{\ell}}
\def\half{\frac{1}{2}}
\def\ninv{n^{-1}}
\def\nhalf{n^{\half}}
\def\mhalf{m^{\half}}
\def\nnhalf{n^{-\half}}
\def\mnhalf{m^{-\half}}
\def\MN{\mbox{MN}}
\def\N{\mbox{N}}
\def\E{\mbox{E}}
\def\pr{P}
\def\var{\mbox{var}}
\def\limn{\lim_{n\to \infty} }
\def\intt{\int_{\tau_a}^{\tau_b}}
\def\sumin{\sum_{i=1}^n}
\def\sumjn{\sum_{j=1}^n}
\def\SUMin{{\displaystyle \sum_{i=1}^n}}
\def\SUMjn{{\displaystyle \sum_{j=1}^n}}
\def\myendthm{\begin{flushright} $\diamond $ \end{flushright}}
\def\convd{\overset{\Dsc}{\longrightarrow}}
\def\convp{\overset{\Psc}{\longrightarrow}}
\def\convas{\overset{a.s.}{\longrightarrow}}
\def\hn{\mbox{H}_0}
\def\ha{\mbox{H}_1}

%
%
%
%
%
\def\trans{^{\transpose}}
\def\inv{^{-1}}
\def\twobyone#1#2{\left[
\begin{array}
{c}
#1\\
#2\\
\end{array}
\right]}
%
%
%
%
%
\def\argmindum{\mathop{\mbox{argmin}}}
\def\argmin#1{\argmindum_{#1}}
\def\argmaxdum{\mathop{\mbox{argmax}}}
\def\argmax#1{\argmaxdum_{#1}}
\def\blockdiag{\mbox{blockdiag}}
\def\corr{\mbox{corr}}
\def\cov{\mbox{cov}}
\def\diag{\mbox{diag}}
\def\dffit{df_{{\rm fit}}}
\def\dfres{df_{{\rm res}}}
\def\dfyhat{df_{\yhat}}
\def\diag{\mbox{diag}}
\def\diagonal{\mbox{diagonal}}
\def\logit{\mbox{logit}}
\def\stdev{\mbox{st.\,dev.}}
\def\stdevhat{{\widehat{\mbox{st.dev}}}}
\def\tr{\mbox{tr}}
\def\trigamma{\mbox{trigamma}}
\def\var{\mbox{var}}
\def\vecof{\mbox{vec}}
\def\AIC{\mbox{AIC}}
\def\AMISE{\mbox{AMISE}}
\def\Corr{\mbox{Corr}}
\def\Cov{\mbox{Cov}}
\def\CV{\mbox{CV}}
\def\GCV{\mbox{GCV}}
\def\LR{\mbox{LR}}
\def\MISE{\mbox{MISE}}
\def\MSSE{\mbox{MSSE}}
\def\ML{\mbox{ML}}
\def\REML{\mbox{REML}}
\def\RMSE{{\rm RMSE}}
\def\RSS{\mbox{RSS}}
\def\Var{\mbox{Var}}
%
%
%
%
\def\bib{\vskip12pt\par\noindent\hangindent=1 true cm\hangafter=1}
\def\jump{\vskip3mm\noindent}
\def\mybox#1{\vskip1mm \begin{center}
        \hspace{.0\textwidth}\vbox{\hrule\hbox{\vrule\kern6pt
\parbox{.9\textwidth}{\kern6pt#1\vskip6pt}\kern6pt\vrule}\hrule}
        \end{center} \vskip-5mm}
\def\lboxit#1{\vbox{\hrule\hbox{\vrule\kern6pt
      \vbox{\kern6pt#1\vskip6pt}\kern6pt\vrule}\hrule}}
\def\boxit#1{\begin{center}\fbox{#1}\end{center}}
\def\thickboxit#1{\vbox{{\hrule height 1mm}\hbox{{\vrule width 1mm}\kern6pt
          \vbox{\kern6pt#1\kern6pt}\kern6pt{\vrule width 1mm}}
               {\hrule height 1mm}}}
\def\instep{\vskip12pt\par\hangindent=30 true mm\hangafter=1}
\def\uWand{\underline{Wand}}
\def\remtask#1#2{\mmnote{\thickboxit
                 {\bf #1\ \theremtask}}\refstepcounter{remtask}}
%
%
%

%
%
\def\aism{{\it Ann. Inst. Statist. Math.}\ }
\def\ajs{{\it Austral. J. Statist.}\ }
\def\ANNSTAT{{\it The Annals of Statistics}\ }
\def\annmath{{\it Ann. Math. Statist.}\ }
\def\applstat{{\it Appl. Statist.}\ }
\def\BIOMETRICS{{\it Biometrics}\ }
\def\cjs{{\it Canad. J. Statist.}\ }
\def\csda{{\it Comp. Statist. Data Anal.}\ }
\def\cstm{{\it Comm. Statist. Theory Meth.}\ }
\def\ieeetit{{\it IEEE Trans. Inf. Theory}\ }
\def\isr{{\it Internat. Statist. Rev.}\ }
\def\JASA{{\it Journal of the American Statistical Association}\ }
\def\JCGS{{\it Journal of Computational and Graphical Statistics}\ }
\def\jscs{{\it J. Statist. Comput. Simulation}\ }
\def\jma{{\it J. Multivariate Anal.}\ }
\def\jns{{\it J. Nonparametric Statist.}\ }
\def\JRSSA{{\it Journal of the Royal Statistics Society, Series A}\ }
\def\JRSSB{{\it Journal of the Royal Statistics Society, Series B}\ }
\def\JRSSC{{\it Journal of the Royal Statistics Society, Series C}\ }
\def\jspi{{\it J. Statist. Planning Inference}\ }
\def\ptrf{{\it Probab. Theory Rel. Fields}\ }
\def\sankhyaa{{\it Sankhy$\bar{{\it a}}$} Ser. A\ }
\def\sjs{{\it Scand. J. Statist.}\ }
\def\spl{{\it Statist. Probab. Lett.}\ }
\def\statsci{{\it Statist. Sci.}\ }
\def\techno{{\it Technometrics}\ }
\def\tpa{{\it Theory Probab. Appl.}\ }
\def\zw{{\it Z. Wahr. ver. Geb.}\ }
%
%
%
%
\def\Brent{{\bf BRENT:}\ }
\def\David{{\bf DAVID:}\ }
\def\Erin{{\bf ERIN:}}
\def\Gerda{{\bf GERDA:}\ }
\def\Joel{{\bf JOEL:}\ }
\def\Marc{{\bf MARC:}\ }
\def\Matt{{\bf MATT:}\ }
\def\Tianxi{{\bf TIANXI:}\ }
%
%
%
%
\def\bZE{\bZ_{\scriptscriptstyle E}}
\def\bZT{\bZ_{\scriptscriptstyle T}}
\def\bbE{\bb_{\scriptscriptstyle E}}
\def\bbT{\bb_{\scriptscriptstyle T}}
\def\bbhatT{\bbhat_{\scriptscriptstyle T}}
\def\fX{f_{\scriptscriptstyle X}}
\def\sigeps{\sigma_{\varepsilon}}
\def\bVtheta{\bV_{\smbtheta}}
\def\bVthetainv{\bVtheta^{-1}}
\def\bKsc{\boldsymbol{\Ksc}}
\def\bxbar{\bar{\bx}}
\def\bPL{b^{\scriptscriptstyle{\rm PL}}}
\def\bVA{b^{\scriptscriptstyle{\rm VA}}}
\def\zPL{z^{\scriptscriptstyle{\rm PL}}}
\def\zVA{z^{\scriptscriptstyle{\rm VA}}}
\def\bYmis{\bY_{\scriptscriptstyle{\rm mis}}}
\def\bYmishat{{\widehat{\bYmis}}}
\def\bYmisone{\bY_{\scriptscriptstyle{\rm mis,1}}}
\def\bYmistwo{\bY_{\scriptscriptstyle{\rm mis,2}}}
\def\bYobs{\bY_{\scriptscriptstyle{\rm obs}}}
\def\bdobs{\bd_{\scriptscriptstyle{\rm obs}}}
\def\bdmis{\bd_{\scriptscriptstyle{\rm mis}}}
%
%
%
%
\def\bfDelta{{\mbox{\boldmath$\Delta$}}}
\def\bfkappa{{\mbox{\boldmath$\kappa$}}}
\def\bfgamma{{\mbox{\boldmath$\gamma$}}}
\def\bftheta{{\mbox{\boldmath$\theta$}}}
\def\bfmu{{\mbox{\boldmath$\mu$}}}
\def\bfdelta{{\mbox{\boldmath$\delta$}}}
\def\bfeps{{\mbox{\boldmath$\varepsilon$}}}
\def\bfnu{{\mbox{\boldmath$\nu$}}}
\def\bfzeta{{\mbox{\boldmath$\zeta$}}}
\def\bfchi{{\mbox{\boldmath$\chi$}}}
\def\bbX{\mathbb{X}}
\def\bbV{\mathbb{V}} 
\def\bbA{\mathbb{A}}
\def\bbB{\mathbb{B}}
\def\bbK{\mathbb{K}}
\def\bbP{\mathbb{P}}
\def\bbD{\mathbb{D}}

\def\Abb{\mathbb{A}}
\def\Bbb{\mathbb{B}}
\def\Cbb{\mathbb{C}}
\def\Dbb{\mathbb{D}}
\def\Ebb{\mathbb{E}}
\def\Fbb{\mathbb{F}}
\def\Gbb{\mathbb{G}}
\def\Hbb{\mathbb{H}}
\def\Ibb{\mathbb{I}}
\def\Jbb{\mathbb{J}}
\def\Kbb{\mathbb{K}}
\def\Lbb{\mathbb{L}}
\def\Mbb{\mathbb{M}}
\def\Nbb{\mathbb{N}}
\def\Mbb{\mathbb{M}}
\def\Nbb{\mathbb{N}}
\def\Obb{\mathbb{O}}
\def\Pbb{\mathbb{P}}
\def\Qbb{\mathbb{Q}}
\def\Rbb{\mathbb{R}}
\def\Sbb{\mathbb{S}}
\def\Tbb{\mathbb{T}}
\def\Ubb{\mathbb{U}}
\def\Vbb{\mathbb{V}}
\def\Wbb{\mathbb{W}}
\def\Xbb{\mathbb{X}}
\def\Ybb{\mathbb{Y}}
\def\Zbb{\mathbb{Z}}

\def\Abbtilde{\widetilde{\mathbb{A}}}
\def\Bbbtilde{\widetilde{\mathbb{B}}}
\def\Cbbtilde{\widetilde{\mathbb{C}}}
\def\Dbbtilde{\widetilde{\mathbb{D}}}
\def\Ebbtilde{\widetilde{\mathbb{E}}}
\def\Fbbtilde{\widetilde{\mathbb{F}}}
\def\Gbbtilde{\widetilde{\mathbb{G}}}
\def\Hbbtilde{\widetilde{\mathbb{H}}}
\def\Ibbtilde{\widetilde{\mathbb{I}}}
\def\Jbbtilde{\widetilde{\mathbb{J}}}
\def\Kbbtilde{\widetilde{\mathbb{K}}}
\def\Lbbtilde{\widetilde{\mathbb{L}}}
\def\Mbbtilde{\widetilde{\mathbb{M}}}
\def\Nbbtilde{\widetilde{\mathbb{N}}}
\def\Mbbtilde{\widetilde{\mathbb{M}}}
\def\Nbbtilde{\widetilde{\mathbb{N}}}
\def\Obbtilde{\widetilde{\mathbb{O}}}
\def\Pbbtilde{\widetilde{\mathbb{P}}}
\def\Qbbtilde{\widetilde{\mathbb{Q}}}
\def\Rbbtilde{\widetilde{\mathbb{R}}}
\def\Sbbtilde{\widetilde{\mathbb{S}}}
\def\Tbbtilde{\widetilde{\mathbb{T}}}
\def\Ubbtilde{\widetilde{\mathbb{U}}}
\def\Vbbtilde{\widetilde{\mathbb{V}}}
\def\Wbbtilde{\widetilde{\mathbb{W}}}
\def\Xbbtilde{\widetilde{\mathbb{X}}}
\def\Ybbtilde{\widetilde{\mathbb{Y}}}
\def\Zbbtilde{\widetilde{\mathbb{Z}}}

%
%
%
%
\def\miss{\mbox{{\tiny miss}}}
\def\obs{\scriptsize{\mbox{obs}}}

%
%
%
%
\def\bmath#1{\mbox{\boldmath$#1$}}
\def\fat#1{\hbox{\rlap{$#1$}\kern0.25pt\rlap{$#1$}\kern0.25pt$#1$}}
\def\wh{\widehat}
\def\flambda{\fat{\lambda}}
\def\beps{\bmath{\varepsilon}}
\def\bSlambda{\bS_{\lambda}}
\def\ErrorSS{\mbox{RSS}}
\def\bsqbar{\bar{{b^2}}}
\def\bcubar{\bar{{b^3}}}
\def\plargest{p_{\rm \,largest}}
\def\summheading#1{\subsection*{#1}\hskip3mm}
\def\summbreak{\vskip3mm\par}
\def\df{df}
\def\adf{adf}
\def\dffit{df_{{\rm fit}}}
\def\dfres{df_{{\rm res}}}
\def\dfyhat{df_{\yhat}}
\def\sigb{\sigma_b}
\def\sigu{\sigma_u}
\def\sigepshat{{\widehat\sigma}_{\varepsilon}}
\def\siguhat{{\widehat\sigma}_u}
\def\sigepshat{{\widehat\sigma}_{\varepsilon}}
\def\sigbhat{{\widehat\sigma}_b}
\def\sighat{{\widehat\sigma}}
\def\sigsqb{\sigma^2_b}
\def\sigsqeps{\sigma^2_{\varepsilon}}
\def\sigsqepszerohat{{\widehat\sigma}^2_{\varepsilon,0}}
\def\sigsqepshat{{\widehat\sigma}^2_{\varepsilon}}
\def\sigsqbhat{{\widehat\sigma}^2_b}
\def\dfnumer{{\rm df(II}|{\rm I)}}
\def\mhatlam{{\widehat m}_{\lambda}}
\def\calD{\Dsc}
\def\Aeps{A_{\epsilon}}
\def\Beps{B_{\epsilon}}
\def\Ab{A_b}
\def\Bb{B_b}
\def\bXtmain{\tilde{\bX}_r}
\def\main{\mbox{\tt main}}
\def\argminbetab{\argmin{\bbeta,\bb}}
\def\calB{\Bsc}
\def\respvar{\mbox{\tt log(amt)}}

\def\Abb{\mathbb{A}}
\def\Zbb{\mathbb{Z}}
\def\Wbb{\mathbb{W}}
\def\Wbbhat{\widehat{\mathbb{W}}}
\def\Kbbtilde{\widetilde{\mathbb{K}}}
\def\Pbbtilde{\widetilde{\mathbb{P}}}
\def\Dbbtilde{\widetilde{\mathbb{D}}}
\def\Bbbtilde{\widetilde{\mathbb{B}}}

\def\Abbhat{\widehat{\mathbb{A}}}

\def\ellhat{\widehat{\ell}}
\def\pn{\phantom{-}}
\def\pp{\phantom{1}}

\def\PP{\stackrel{P}{\rightarrow}}
\def\DD{\Rightarrow}
%
%

{
\let\newpage\relax
\maketitle
}

\tableofcontents

\newpage

\section{Comparison with Existing Literature and Software}

\begin{table}[htbp!]
\resizebox{\textwidth}{!}{%
    \begin{tabular}{|c|c|c|c|c|c|c|}
    \hline
       \textbf{Reference} & \textbf{\begin{tabular}[c]{@{}c@{}}Specific \\ to Healthcare\end{tabular}} & \textbf{\begin{tabular}[c]{@{}c@{}}Fairness \\ Categories\end{tabular}}  & \textbf{\begin{tabular}[c]{@{}c@{}}Code \\ Available\end{tabular}} & \textbf{\begin{tabular}[c]{@{}c@{}}Illustrated \\ with Real Data\end{tabular}}  & \textbf{\begin{tabular}[c]{@{}c@{}}Inference\end{tabular}} & \textbf{Format}  \\ 
       \hline
       This Tutorial  & \textcolor{blue}{\checkmark} & \textcolor{blue}{G, I, S, C} & \textcolor{blue}{\texttt{R}} & \textcolor{blue}{\checkmark} & \textcolor{blue}{\checkmark} & Article \\
       \hline
       \multicolumn{7}{|l|}{\textit{Reviews and Surveys}} \\
       \hline
       \cite{mehrabi_survey_2021} & \textcolor{red}{\texttimes} & G, I, S & \textcolor{red}{\texttimes} & \textcolor{red}{\texttimes} & \textcolor{red}{\texttimes} & Article\\
       \hline
       \cite{plecko2022causal} & \textcolor{red}{\texttimes} & G, I, C & \textcolor{red}{\texttimes} & \textcolor{red}{\texttimes} & \textcolor{blue}{\checkmark} & Article \\ \hline
       \cite{verma_fairness_2018, mitchell_prediction-based_2021, makhlouf_applicability_2021, wang_brief_2022, makhlouf2022survey, alves_survey_2023, caton_fairness_2023} & \textcolor{red}{\texttimes} & G, I, C & \textcolor{red}{\texttimes} & \textcolor{red}{\texttimes} & \textcolor{red}{\texttimes} & Article \\ \hline
       \cite{nilforoshan_causal_2022} & \textcolor{red}{\texttimes} & G, C & \textcolor{blue}{\texttt{Python}} & \textcolor{blue}{\checkmark} & \textcolor{red}{\texttimes} & Article \\ \hline
       \cite{rajkomar_ensuring_2018, wang_bias_2022, yfantidou_accuracy_2023, chen_algorithmic_2023, chen2024unmasking} & \textcolor{blue}{\checkmark} & G & \textcolor{red}{\texttimes} & \textcolor{red}{\texttimes} & \textcolor{red}{\texttimes} & Article\\
       \hline
       \multicolumn{7}{|l|}{\textit{Tutorials}}\\
       \hline
       \cite{ahmad_fairness_2020} & \textcolor{blue}{\checkmark} & G, I, S & \textcolor{blue}{\texttt{Python}} & \textcolor{blue}{\checkmark} & \textcolor{red}{\texttimes} & Slides\\
       \hline
       \cite{ghani_addressing_2023} & \textcolor{red}{\texttimes} & G & \textcolor{blue}{\texttt{Python}} & \textcolor{blue}{\checkmark} & \textcolor{red}{\texttimes} & Slides, Video \\
       \hline
       \cite{chang_bias_2019, bird_fairness-aware_2019} & \textcolor{red}{\texttimes} & G & \textcolor{red}{\texttimes} &\textcolor{blue}{\checkmark} &\textcolor{red}{\texttimes} & Slides, Video \\
       \hline
       \cite{farnadi_algorithmic_2022} & \textcolor{red}{\texttimes} & G, C & \textcolor{red}{\texttimes} & \textcolor{blue}{\checkmark} & \textcolor{red}{\texttimes} & Slides, Video\\
       \hline
       \cite{arvind_narayanan_tutorial_2018} & \textcolor{red}{\texttimes} & G, I & \textcolor{red}{\texttimes} & \textcolor{red}{\texttimes} & \textcolor{red}{\texttimes} & Video \\
       \hline
       \end{tabular}}
    \caption{\textbf{Comparison of current tutorial with existing literature on algorithmic fairness.} G: Group  fairness, I: Individual fairness, S: Subgroup fairness, C: Causality-based fairness}
    \label{tab:lit-review}
\end{table}

\begin{table}[htbp!]
\centering
\resizebox{\textwidth}{!}{%
    \begin{tabular}{|c|c|c|c|}
    \hline
    \textbf{Toolkit} & \textbf{Supported Languages} & \textbf{Point Estimates Type} & \textbf{Confidence Interval Availability} \\
    \hline
    Our Package & \texttt{R} & Difference, Ratio & \textcolor{blue}{\checkmark} \\
    \hline
    Fairness Package \cite{nikita_fairness_2021} & \texttt{R} & Ratio & \textcolor{red}{\texttimes} \\
    \hline
    Fairlearn \cite{weerts_fairlearn_2023} & \texttt{Python} & Difference, Ratio & \textcolor{red}{\texttimes} \\
    \hline
    Aequitas \cite{saleiro_aequitas_2019}& \texttt{Python} & Ratio & \textcolor{red}{\texttimes} \\
    \hline
    AI Fairness 360 \cite{bellamy_ai_2018}& \texttt{Python, R} & - & \textcolor{red}{\texttimes} \\
    \hline
    TensorFlow Fairness Indicators \cite{tensorflow_developers_tensorflow_2024}& \texttt{Python} & - & \textcolor{red}{\texttimes} \\
    \hline
    \end{tabular}
}
\caption{\textbf{Comparison of our R package with existing fairness toolkits.} The last two toolkits are based on direct visualizations of model performance metrics. The metrics included in each toolkits are generally very similar.}
\label{tab:software-comp}
\end{table}
\newpage

\newpage

\section{Group Fairness Criteria}

\subsection{Statistical Metrics for Model Evaluation}

Model performance in classification is typically evaluated from two perspectives: discrimination and calibration. The mathematical and intuitive definitions of various discrimination and calibration metrics are presented in Table \ref{tab:modeleval}.  Discrimination evaluates an ML model's capacity to distinguish between the positive and negative class while calibration evaluates the level of agreement between the risk score and the observed frequency of that outcome. A model is calibrated if, among the proportion of observations with a risk score of $s$, a fraction $s$ have the positive outcome.  A well-calibrated model is crucial for clinical decision-making as it ensures the risk scores derived from ML model accurately reflect the true likelihood of a positive outcome \cite{van_calster_calibration_2019}.  While numerous measures of calibration exist, we present the Brier score (BS) and mean absolute error (MAE).  A BS or MAE closer to 0 indicates a more calibrated model. It is worth noting that both BS and MAE remains valid when $Y$ is continuous. In this context, the BS is equivalent to the mean squared error (MSE). 

\begin{table}[ht!]
\centering
\centering
\resizebox{\textwidth}{!}{%
    \begin{tabular}{|l|l|l|}
  \hline
   \textbf{Metric} & \textbf{Definition} & \textbf{Explanation} \\
   \hline
   \multicolumn{3}{|l|}{\textit{Discrimination}} \\
   \hline
   True Positive (TP) & $\# \{Y =1, D = 1\}$ & Count of correctly classified positives. \\
   \hline
   False Positive (FP) & $\# \{Y =0, D = 1\}$ & Negatives incorrectly classified as positives. \\
   \hline
   True Negative (TN) & $\# \{Y =0, D = 0\}$ & Count of correctly classified negatives. \\
   \hline
   False Negative (FN) & $\# \{Y =1, D = 0\}$ & Positives incorrectly classified as negatives. \\
   \hline
   True Positive Rate (TPR) & $P(D=1\mid Y=1)= \frac{TP}{TP+FN}$ & Proportion of positives correctly classified. \\
   \hline
   True Negative Rate (TNR) & $P(D = 0 \mid Y = 0) = \frac{TN}{TN+FP}$ & Proportion of negatives correctly classified. \\
   \hline
   False Positive Rate (FPR) & $P(D = 1\mid Y = 0) = \frac{FP}{FP+TN}$ & Proportion of negatives incorrectly classified. \\
   \hline
   False Negative Rate (FNR) & $P(D = 0 \mid Y = 1) = \frac{FN}{TP+FN}$ & Proportion of positives incorrectly classified. \\
   \hline
   Positive Predictive Value (PPV) & $P(Y = 1 \mid D = 1) = \frac{TP}{TP+FP}$ & Proportion of predicted positives that are true. \\
   \hline
   Negative Predictive Value (NPV) & $P(Y = 0 \mid D = 0) = \frac{TN}{TN+FN}$ & Proportion of predicted negatives that are true. \\
   \hline
   Accuracy (ACC) & $P(D = Y) = \frac{TP + TN}{TP + FP + TN + FN}$ & Overall proportion of correctly classified. \\
   \hline
   \multicolumn{3}{|l|}{\textit{Calibration}} \\
   \hline
   Brier's Score (BS)* & $\mathbb{E}[(S-Y)^2]$ & Mean squared difference between predicted probabilities and outcomes. \\
   \hline
   Mean Absolute Error (MAE)* & $\mathbb{E}|S-Y|$ & Mean absolute difference between predicted probabilities and outcomes\\
   \hline
\end{tabular}}
\caption{\textbf{Statistical metrics for evaluating classification performance.} Symbols: $\# =$ number of, $P =$ probability, $E =$ expected value. Notations: $Y:$ outcome of interest, $S:$ model output, $D:$ predicted outcome by thresholding $S$. An asterisk (*) indicates that the metrics are applicable whether $Y$ is binary or continuous. }
\label{tab:modeleval}
\end{table}

\subsection{Continuous Outcome}
Many clinical applications involve models for predicting continuous outcomes, such as healthcare costs.  In Table \ref{tab:modeleval}, we briefly touched on a few metrics that can be used to evaluate the performance of an algorithm when the outcome is continuous. The parity between these metrics can be used to evaluate fairness analogously to the setting of a binary outcome. Recently, \cite{steinberg_fairness_2020} proposed methods to measure independence, separation, and sufficiency for continuous outcomes by estimating conditional densities or using mutual information, though their methods are limited to binary protected attributes. An alternative approach involves conducting conditional statistical tests, such as testing independence through the Hirschfeld-Gebelein-Renyi maximal correlation coefficient, which accommodates both continuous outcomes and protected attributes \cite{mary_fairness-aware_2019}. This is an ongoing area of research and we encourage readers to consult \cite{steinberg_fairness_2020, kim_measuring_2023} for further discussion of the topic.

\subsection{Categorical Protected Attributes}
When the protected attributes are binary, natural measures of parity include the difference and the ratio as we introduced in the main text of tutorial. However, many protected attributes 
define more than two groups, such as race and marital status.  One approach for such protected attributes is to directly visualize the metrics in toolkit such as AI Fairness 360 \cite{bellamy_ai_2018}. However, this method has notable limitations, including subjective judgment and challenges when dealing with numerous subgroups \cite{lum_-biasing_2022}. To address these issues, various meta-metrics have been proposed and applied to quantify group-wise disparities into a single metric. \cite{lum_-biasing_2022} have summarized common meta-metrics from the literature, which we present in Table \ref{tab:MetaMetrics} below. These metrics primarily focus on two key aspects: extremum and variability of model performances among groups.

\begin{table}[htbp!]
\centering
    \begin{tabular}{|l|l|l|}
    \hline
        \textbf{Meta-Metrics} & \textbf{Formula} & \textbf{Type} \\
        \hline 
         max-min difference & $\max_{j} \mathcal{M}_j - \min_j \mathcal{M}_j$ & Extremum \\
         \hline
         max-min ratio & $ \frac{\max_j \mathcal{M}_j}{\min_j \mathcal{M}_j}$ & Extremum \\
         \hline
         max absolute difference & $\max_j \left|\mathcal{M}_j - \frac{1}{K}\sum_{i=1}^{K}\mathcal{M}_i\right|$ & Extremum \\
         \hline
         mean absolute deviation & $\frac{1}{K} \sum_{j=1}^{K} \left|\mathcal{M}_j - \frac{1}{K}\sum_{i=1}^{K}\mathcal{M}_i\right|$ & Variability \\
         \hline
         variance & $\frac{1}{K-1} \sum_{j=1}^{K} \left(\mathcal{M}_j - \frac{1}{K}\sum_{i=1}^{K}\mathcal{M}_i\right)^2$ & Variability \\
         \hline
         generalized entropy index ($\alpha \neq 0, 1$) & $\frac{1}{K\alpha(\alpha-1)}\sum_{j=1}^{K} \left[\left(\frac{\mathcal{M}_j}{\frac{1}{K}\sum_{i=1}^{K}\mathcal{M}_i}\right)^\alpha - 1 \right]$ & Variability \\
         \hline
    \end{tabular}
    \caption{\textbf{Meta-metrics for evaluating fairness across $K$ protected groups.}  $\mathcal{M}_j$ denotes a model performance metric (e.g., the true positive rate) for $j = 1, \dots, K$.
    These meta-metrics summarize a model's performance metric, $\mathcal{M}_j$, across $j = 1, ..., K$ groups. }
    \label{tab:MetaMetrics}
\end{table}

\subsection{Continuous Protected Attributes}
When dealing with continuous protected attributes, such as age, applying thresholds or bins may introduce threshold effects \cite{mary_fairness-aware_2019}. Only very recently, \cite{mary_fairness-aware_2019} proposed using the Hirschfeld-Gebelein-Rényi (HGR) maximal correlation coefficient as a metric to measure fairness for continuous protected attributes. This is an ongoing area of research and we encourage readers to consult \cite{mary_fairness-aware_2019, giuliani_generalized_2023} and references therein.

\subsection{Subgroup Fairness}
When multiple attributes are of interest (e.g., sex = \{male, female\} and race = \{white, black\}), a model can be fair in both groups, but show disparities in an intersectional subgroup (e.g., \{black male\}), a phenomenon known as ``Fairness Gerrymandering" \cite{kearns_preventing_2018}. Subgroup fairness or intersectional fairness selects a specific fairness constraint, such as statistical parity or equalized odds, and then evaluates it across a large collection of subgroups. For instance, a recent study evaluated a chest X-ray algorithm across the 8 possible intersectional subgroup of income = \{high, low\}, insurance = \{high, low\}, race =\{white, non-white\} by evaluating the differences between the maximum and minimum false positive rates \cite{moukheiber_looking_2024}.

\section{Group Fairness Incompatibilities}
\subsection{Independence versus Sufficiency}
If the outcome and protected groups are dependent ($Y \not\indep A$), then it is not possible to achieve both independence and sufficiency. Specifically, when $Y$ is binary and prevalence differs across protected groups, an algorithm cannot simultaneously satisfy these two criteria.

\subsection{Independence versus Separation}
Besides the dependency between the outcome and protected group ($Y \not\indep A$), if prediction also depends on the outcome ($D \not\indep Y$), independence and separation are incompatible. It is important to note that this statement only holds for binary outcomes. If the prediction is not entirely uninformative for predicting the labels, then there should be correlation between the prediction and the label (i.e. $D \not\indep Y$). However, with unequal prevalence across protected groups, satisfying both criteria simultaneously would yield predictions that are essentially uninformative. 

\subsection{Sufficiency versus Separation}
Besides the dependency between the outcome and protected group ($Y \not\indep A$), if the density of the joint distribution of protected attributes, prediction, and outcome ($(A, D, Y)$) is strictly positive, then sufficiency and separation are incompatible. For binary outcomes, the second condition means that there is at least one false prediction. In other words, the prediction cannot completely determine the outcome. Consequently, with unequal prevalence across protected groups, achieving both sufficiency and separation becomes unattainable unless the algorithm gives perfect predictions. 


\section{Case Study}

\subsection{Dataset}
The dataset used in Case Study is derived from the MIMIC-II critical care database. It contains clinical data and outcomes used to ``investigate the effectiveness of indwelling arterial catheters in hemodynamically stable patients with respiratory failure regarding mortality outcomes" \cite{raffa2016clinical}. The dataset includes 46 variables, including demographics, clinical characteristics such as white blood cell count and heart rate during ICU stays, and a 28-day mortality indicator on 1,776 patients. It is publicly available at \url{https://physionet.org/content/mimic2-iaccd/1.0/}.

\subsection{Data Preprocessing}
We removed three variables that had more than 10\% missing values: body mass index (26.2\%), first partial pressure of oxygen (10.5\%), and first partial pressure of carbon dioxide (10.5\%). For the remaining variables with missing values, we imputed the missing data with their respective medians. Additionally, we removed the variable `sepsis\_flg' as it was absent in all records. Lastly, four variables — death in hospital, death in ICU, day post ICU admission of censoring or death, and censored or death—were removed because they cannot be used as predictors for mortality as they are inherently dependent on the outcome being predicted.

\subsection{Model Building}
We used the first 700 individuals as the training set to fit a random forest (RF) model and the remaining 1,076 individuals served as the testing set. The hyperparameters for the RF model were set to use 1000 trees and a random sampling of 6 variables at each split, determined by the square root of the number of predictors. The area under the receiver operating characteristic curve (AUC) for our model on the test set is 0.90. Additionally, the accuracy of this random forest classification model on the test set is 0.88. We selected a cut-off value of 0.41 for fairness evaluation, as it achieves a false positive rate (FPR) of approximately 5\%.



\newpage

\section{Construction of Confidence Intervals}
We illustrate the procedure for constructing confidence intervals (CIs) using false positive rate (FPR). Let $\widehat{\textrm{FPR}}_a$ and $\textrm{FPR}_a$ represent the estimated and true FPRs in group $a \in \mathcal{A}$. We construct Wald-type CIs based on the following result: $\sqrt{n} \left(\widehat{\textrm{FPR}}_a - \textrm{FPR}_a \right)$ is asymptotic normal with mean 0 for all $a \in \mathcal{A}$ \cite[e.g.][]{gronsbell2018semisupervised}. Let $\widehat{\Delta}_{\textrm{FPR}} = \widehat{\textrm{FPR}}_{a_1} - \widehat{\textrm{FPR}}_{a_0}$ represent the estimated difference between group $a_1$ and $a_0$. It follows that $\sqrt{n}\left(\widehat{\Delta}_{\textrm{FPR}} - \Delta_{\textrm{FPR}}\right)$ is asymptotic normal with mean 0. To estimate the standard error of $\widehat{\Delta}_{\textrm{FPR}}$, we a non-parametric bootstrap \cite{efron1986bootstrap}. For each bootstrap iteration $b$ in \{1,..., B\}, we sample the data with replacement within each group and then calculate $\widehat{\Delta}_{\textrm{FPR}}^{(b)}$. The estimated standard error is the empirical standard error of the difference:
\begin{equation*}
    \widehat{\textrm{se}}\left[\widehat{\Delta}_{\textrm{FPR}}\right] = \sqrt{\frac{1}{B-1}\sum_{b=1}^B \left(\widehat{\Delta}_{\textrm{FPR}}^{(b)} - \frac{1}{B} \sum_{b=1}^B \widehat{\Delta}_{\textrm{FPR}}^{(b)}\right)^2}.
\end{equation*}
The corresponding Wald-based $100(1-\alpha)\%$ CI is given by $\widehat{\Delta}_{\textrm{FPR}} \pm z_{1-\alpha/2} * \widehat{\textrm{se}}\left[\widehat{\Delta}_{\textrm{FPR}}\right] $, where $z_{1-\alpha/2}$ is the $100(1-\alpha/2)$ percentile of the standard normal distribution. \\

To construct a $100(1-\alpha)\%$ Wald-type CI for the ratio $\widehat{\rho}_{\textrm{FPR}} = \frac{\widehat{\textrm{FPR}}_{a_1}}{\widehat{\textrm{FPR}}_{a_0}}$, we note that $\sqrt{n}\left[\log\left(\widehat{\rho}_{\textrm{FPR}}\right)- \log\left(\rho_{\textrm{FPR}}\right)\right]$ is asymptotic normal with mean 0. For each bootstrap iteration $b$ in \{1,..., B\}, we sample the data with replacement within each group and then calculate $\log\left[\widehat{\rho}_{\textrm{FPR}}^{(b)}\right]$. The estimated standard error of the log ratio is:
\begin{equation*}
\widehat{\textrm{se}}\left[\log\left(\widehat{\rho}_{\textrm{FPR}}\right)\right] = \sqrt{\frac{1}{B-1}\sum_{b=1}^B \left\{\log\left[\widehat{\rho}_{\textrm{FPR}}^{(b)}\right] - \frac{1}{B} \sum_{b=1}^B \log\left[\widehat{\rho}_{\textrm{FPR}}^{(b)}\right]\right\}^2}.
\end{equation*}
The corresponding $100(1-\alpha)\%$ CI is given by $\exp \left\{\log\left[\widehat{\rho}_{\textrm{FPR}}\right] \pm z_{1-\alpha/2} * \widehat{\textrm{se}}\left[\log\left(\widehat{\rho}_{\textrm{FPR}}\right)\right]\right\} $, where $z_{1-\alpha/2}$ is the $100(1-\alpha/2)$ percentile of the standard normal distribution.

\section{Causal Fairness Criteria}

\begin{table}[htbp!]
    \centering
    \begin{tabular}{|p{0.2\textwidth}|p{0.75\textwidth}|}
    \hline
    \textbf{Metric} & \textbf{Definition}  \\
    \hline
    Counterfactual Equalized Odds & $P(D = 1 \mid Y^1 = 0, A = a_0) = P(D =1 \mid Y^1 = 0, A = a_1)$ and
    $P(D = 0 \mid Y^1 = 1, A = a_0) = P(D = 0 \mid Y^1 = 1, A= a_1), \forall a_0, a_1 \in \Asc$ \\
    \hline
    Counterfactual Predictive Parity & $P(Y^1 = 1 \mid D = 1, A = a_0) = P(Y^1 = 1 \mid D = 1, A = a_0), \forall a_0, a_1 \in \Asc $\\
    \hline
    Principal Fairness & $P(D = 1 \mid R, A = a_0) = P(D = 1 \mid R, A = a_1)$ for each $R$ and $\forall a_0, a_1 \in \Asc$ \\
    \hline
    Conditional Principal Fairness & $P(D = 1 \mid R, \bZ,  A = a_0) = P(D = 1 \mid \bZ, R, A = a_1)$ for each $R$ and $\forall a_0, a_1 \in \Asc$ \\
     \hline
    Counterfactual parity & $E(D(a)) = E(D(a') )$\\ 
    \hline
    Conditional counterfactual fairness & $E(D(a)  \mid \bZ) = E(D(a')  \mid \bZ) $\\ 
    \hline
    \end{tabular}
    \caption{\textbf{Some common notions of causal fairness}. Symbols: $P =$ probability, $E$ = expectation. Notations: $S:$ risk score, $D:$ final prediction based on thresholding $S$,  $A:$ protected attribute that takes value in the set $\mathcal{A}$, $Y^0:$ potential outcome if the decision had been $D = 0$, $Y^1:$ potential outcome if the decision had been $D = 1$, $R$:  $(Y^0, Y^1)$, $\bZ$: additional set of features, $D(a)$: the prediction when one's protected attribute is $a\in \Asc$}
    \label{tab:causal}
\end{table}

\clearpage

\begin{singlespace}
\printbibliography
\end{singlespace}